\pdfoutput=1

\PassOptionsToPackage{table}{xcolor}

\documentclass{article} 

\usepackage[final]{colm2025_conference}
\usepackage{fancyhdr}       

\usepackage[top=0.8in, margin=1.4in]{geometry}

\usepackage{longtable} 
\usepackage{microtype}
\usepackage{hyperref}
\usepackage{url}
\usepackage{amsmath}
\usepackage{booktabs}
\usepackage{tcolorbox}
\usepackage{tabularx}
\usepackage{adjustbox}
\usepackage{listings}
\usepackage{xcolor}
\usepackage[table]{xcolor} 
\usepackage{float}
\usepackage{footmisc}  

\usepackage{mdframed}
\usepackage{courier}
\usepackage{verbatim}
\tcbuselibrary{skins, breakable}
\usepackage{pdfpages}
\usepackage{enumitem}
\definecolor{lightgray}{gray}{0.95}
\definecolor{bordergray}{gray}{0.6}
\lstdefinelanguage{json}{
    basicstyle=\ttfamily\small,
    numbers=none,
    numberstyle=\tiny\color{gray},
    stepnumber=1,
    numbersep=5pt,
    showstringspaces=false,
    breaklines=true,
    frame=single,
    backgroundcolor=\color{lightgray},
    literate=
     *{0}{{{\color{blue}0}}}{1}
      {1}{{{\color{blue}1}}}{1}
      {2}{{{\color{blue}2}}}{1}
      {3}{{{\color{blue}3}}}{1}
      {4}{{{\color{blue}4}}}{1}
      {5}{{{\color{blue}5}}}{1}
      {6}{{{\color{blue}6}}}{1}
      {7}{{{\color{blue}7}}}{1}
      {8}{{{\color{blue}8}}}{1}
      {9}{{{\color{blue}9}}}{1}
      {:}{{{\color{red}:}}}{1}
      {,}{{{\color{red},}}}{1}
      {"}{{{\color{brown}"}}}{1},
}

\tcbuselibrary{listingsutf8}

\definecolor{lightgray}{gray}{0.97}
\definecolor{bordergray}{gray}{0.6}

\newtcolorbox{fancybox}[1][]{
  colback=lightgray,
  colframe=bordergray,
  coltitle=black,
  fonttitle=\bfseries,
  title=#1,
  boxrule=0.5pt,
  arc=4pt,
  outer arc=2pt,
  boxsep=5pt,
  left=5pt,
  right=5pt,
  top=5pt,
  bottom=5pt,
  enhanced,
  sharp corners=south,
  breakable
}

\usepackage{lineno}

\definecolor{darkblue}{rgb}{0, 0, 0.5}
\hypersetup{colorlinks=true, citecolor=darkblue, linkcolor=darkblue, urlcolor=darkblue}

\usepackage{titling}
\pretitle{\begin{center}\Large\bfseries} 
\posttitle{\par\end{center}\vskip 1.5em}
\preauthor{\begin{center}\large}   
\postauthor{\par\end{center}\vskip 1em}
\predate{}\postdate{}

\AtBeginDocument{\fontsize{10.5pt}{12.6pt}\selectfont}

\title{Agent Context Protocols Enhance Collective Inference}

\date{}

\author{%
\makebox[\textwidth][c]{\parbox{1.5\textwidth}{\centering\bfseries
Devansh Bhardwaj\textsuperscript{1}\thanks{Equal contribution}\quad
Arjun Beniwal\textsuperscript{2}\footnotemark[1]\quad
Shreyas Chaudhari\textsuperscript{3}\quad
Ashwin Kalyan\textsuperscript{4}\quad
Tanmay Rajpurohit\textsuperscript{5}}}\\[0.3em]
\makebox[\textwidth][c]{\parbox{1.5\textwidth}{\centering\bfseries
Karthik R.~Narasimhan\textsuperscript{6}\quad
Ameet Deshpande\textsuperscript{6}\quad
Vishvak Murahari\textsuperscript{6}}}\\[0.6em]
\makebox[\textwidth][c]{\parbox{1.5\textwidth}{\centering\large\normalfont
\textsuperscript{1}Indian Institute of Technology Roorkee\quad
\textsuperscript{2}New York University\quad
\textsuperscript{3}University of Massachusetts Amherst}}\\
\makebox[\textwidth][c]{\parbox{1.5\textwidth}{\centering\large\normalfont
\textsuperscript{4}Independent Researcher\quad
\textsuperscript{5}Georgia Tech\quad
\textsuperscript{6}Princeton University}}
}

\newcommand{\ExecGraph}{Execution Blueprint}
\newcommand{\acp}{\textsc{ACP}}

\newcommand{\blfootnote}[1]{%
  \begingroup
    \renewcommand\thefootnote{}
    \footnotetext{#1}
    \addtocounter{footnote}{-1}
  \endgroup
}

\begin{document}

\maketitle

\thispagestyle{empty}   

\begin{abstract}

    AI agents have become increasingly adept at complex tasks such as coding, reasoning, and multimodal understanding.
    However, building generalist systems requires moving beyond individual agents to collective inference---a paradigm where multi-agent systems with diverse, task-specialized agents complement each other through communication and collaboration.
    Traditionally, communication and coordination in multi-agent systems are enabled with imprecise and ad-hoc unstructured natural language, which greatly limits complex inter-agent interaction and constrains interoperability with domain-specific agents.
    To this end, we introduce \textit{agent context protocols} (ACPs): a domain and agent-agnostic set of structured protocols for agent-agent communication, coordination, and error handling.
    ACPs leverage persistent execution blueprints---structured dependency graphs that encode task dependencies and store agent outputs---along with standardized communication schemas, enabling robust and fault-tolerant multi-agent collective inference.
    Our ACP-powered generalist multi-agent systems achieve state-of-the-art (SOTA) results: 28.3\% accuracy on AssistantBench for long-horizon web assistance and best-in-class multimodal technical reports, outperforming commercial AI systems as evaluated by humans.
    ACPs are highly modular and extensible, and enables practitioners to rapidly develop best-in-class generalist systems.
\end{abstract}

\blfootnote{Code available at \url{https://github.com/agent-context-protocol}}
\section{Introduction}

AI agents have demonstrated human-like ability in coding, language generation, reasoning, and multimodal understanding, and have pervaded a wide range of applications and domains, such as---OpenAI Codex \citep{Chen2021Codex}, Medical QA – Med-PaLM \citep{singhal2023large}, \citep{singhal2025toward2}, ManusAI \citep{manusAI}.

While different agents demonstrate domain and task-specific mastery, powerful generalist systems can be constructed through \textit{collective inference} (\cite{dafoe2020open, hong2023metagpt, nomura2024towards, boiko2023emergentautonomousscientificresearch})---where multiple specialized agents seamlessly collaborate and communicate to complement each other. Multi-agent systems are tackling increasingly sophisticated problems---planning and executing an ever-growing set of interdependent heterogeneous agents, synthesizing a wide range of multimodal data, handling a large set of domains, and acting in complex environments over a long horizon.

While multi-agent systems are mushrooming in usage, they lack standardization and interoperability in agentic collaboration, communication, and coordination.
Structured, standardized, and fault-tolerant protocols unlock powerful capabilities.
For instance, such protocols for single-agent like \textit{model context protocol (MCP)} \citep{mcp} have enabled context-aware reasoning at scale through seamless communication between AI agents and data sources.

However, modern inter-agent communication is still enabled through ad-hoc and imprecise unstructured natural language (\cite{autogen2023}, \cite{hong2023metagpt}, \cite{li2023camel}, \cite{fourney2024magenticone}, \cite{chen2024internetagentsweavingweb}).
This limits complex inter-agent interaction, particularly with rigid tools, and hinders self-diagnosing capabilities -- causing simple errors to cascade across complex interdependent agents.

\begin{figure}[t]
    \centering
    \includegraphics[width=1.0\linewidth]{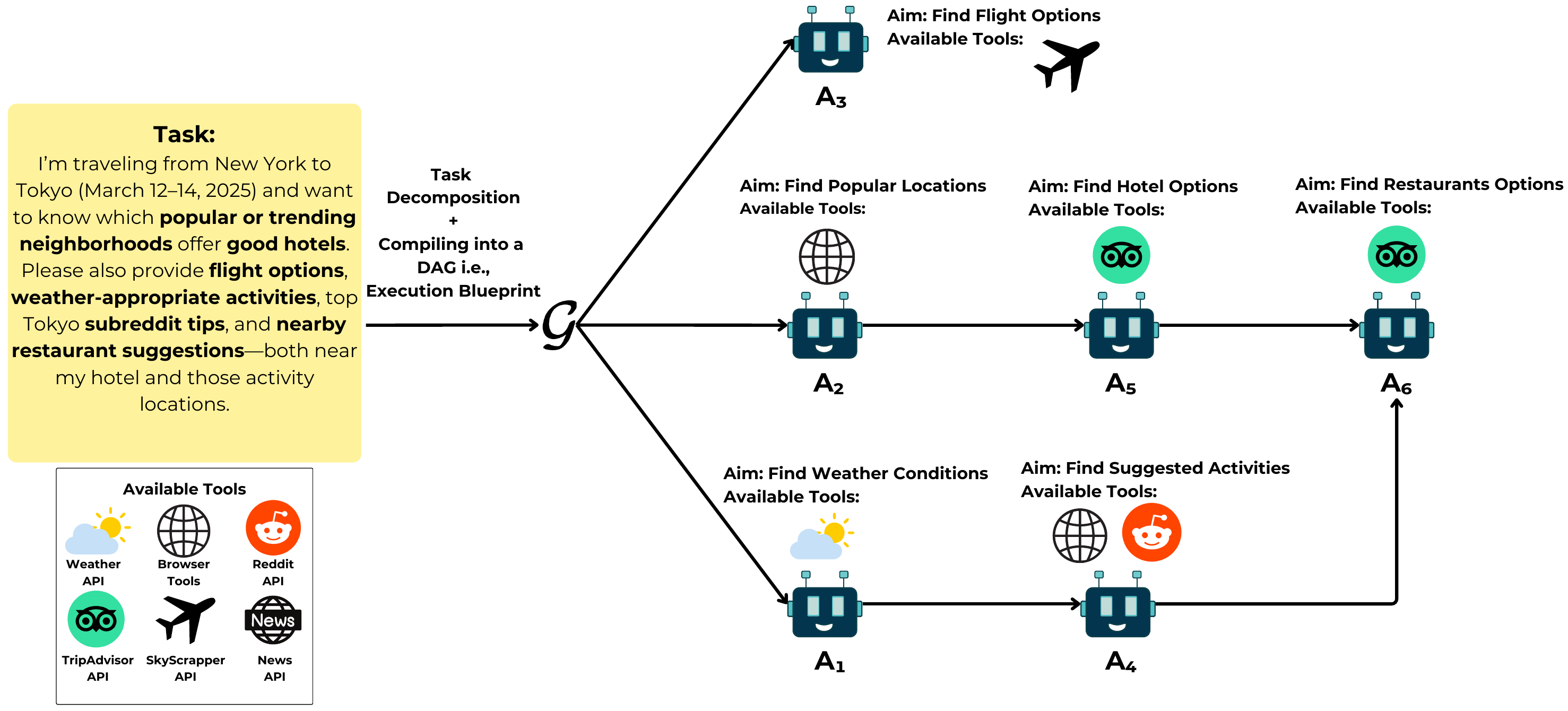}
    \caption{An illustrative overview of our ACP-based system. A complex task $T$ is decomposed into sub-tasks (each handled by an agent $A_i$, having capabilities $o_i$---i.e., specialized tools) and compiled into a DAG-based Execution Blueprint. ACPs then coordinate these specialized agents via structured communication and robust error handling, ensuring each sub-task faithfully adheres to the Execution Blueprint.}
    \label{fig:main_figure}
\vspace{-14pt}
\end{figure}

In this work, we introduce \textit{Agent Context Protocols} (ACPs), a domain and agent-agnostic set of structured protocols for agent-agent communication, coordination, and error handling.
ACPs enable fault-tolerant and long-horizon multi-step collective inference across a diverse set of domain and task-specialized agents.
ACPs are incredibly modular and extensible -- practitioners can leverage ACP to rapidly prototype generalist systems with different capabilities and experiment with a variety of domain and task-specialized agents.

ACPs model collective inference of agents as a persistent directed acyclic graph (DAG) of agent actions/outputs and ensure faithful execution of arbitrarily complex DAGs.
ACPs enable this by leveraging structured protocols and standardized schemas for inter-agent communication (agent responses and requests, assistance requests, error protocols, action execution, and status updates) to enable robust multi-step collective inference.

Importantly, ACPs facilitate fault-tolerant execution and error resolution across very long horizons.
Complex multi-agent systems consist of stochastic agents and tools, which could deviate from the structured schemas and cause errors.
ACPs introduce standardized descriptive status codes (akin to HTTP \citep{http})  and structured context-rich error messages that work with enhanced reasoner LLMs to re-plan and recover.
These status codes and error messages signal the execution status of an agent to the broader system and enable effective localization, categorization, and diagnosis of errors.

We demonstrate two state-of-the-art (SOTA) generalist multi-agent systems powered by ACP, focusing on two highly complex generalist objectives: long-horizon web assistance and synthesizing detailed multimodal reports. ACPs achieve SOTA accuracy (28.3\%) on challenging long-horizon web assistance tasks within AssistantBench~\citep{yoran2024assistantbench}.
In addition, ACPs generate best-in-class multimodal technical reports, significantly beating SOTA approaches (\texttt{Perplexity Deep Research} 
and \texttt{Gemini Deep Research}), across $\approx$ 85\% of the dimensions evaluated.
 
Additional ablation studies on the challenging task of creating information-rich dashboards, which requires significant parallelism and sequential complexity, demonstrate the critical role of the coordination and fault tolerance capabilities provided by ACPs.

ACPs provide a crucial, standardized foundation for robust multi-agent communication,  enhancing collective inference and paving the way for rapid creation of generalist systems.


\section{Formulation and Methodology}

We now formulate multi-agent collaboration to solve complex tasks.
Denote a team of $k$ LLM-based agents by $\mathcal{A} = \{A_1, \dots, A_k\}$. Their goal is to solve a complex problem, denoted by $T$. Examples include tasks that demand intermixing abilities such as multimodal question answering or long-horizon web searches for information retrieval.

\paragraph{Agents and Capabilities:}
Each agent $A_i$ is characterized by its \textit{actions} or \textit{capabilities}, denoted by $\mathcal{O}_i = \{o^i_1, o^i_2, \dots, o^i_m\}$. These capabilities commonly abstract more granular operations. As an example, the web-searching capability encapsulates a series of low-level operations. 
The union of action sets of all agents $\mathcal{O} = \bigcup_{i=1}^k \mathcal{O}_i$ represents the overall set of skills available to address $T$.

\paragraph{Solving a Complex Problem with Agent Capabilities:}
A complex problem $T$ is split into sub-tasks $\{\tau_1, \tau_2, \dots, \tau_n\}$. For instance, one sub-task might use \texttt{WeatherAPI} to retrieve climate data, while another might rely on a \texttt{BrowserTool} to scrape an online datasource. These sub-tasks have data dependencies forming a \emph{directed acyclic graph} (DAG). Concretely, if sub-task $\tau_j$ requires the output of $\tau_i$, the dependency is denoted by a directed edge $\tau_i \to \tau_j$. Sub-tasks are then executed in a topological order, ensuring that prerequisites complete before downstream sub-tasks begin.
Each sub-task $\tau_i$ is assigned to an agent $A_j$ with the appropriate capabilities. 

\paragraph{\ExecGraph:}
To handle tasks at the level of individual tool invocations, each sub-task $\tau_i$ is mapped to a sequence of agent actions $\tau_i \;\longmapsto\; \bigl(o_{1}^{(j)}, o_{2}^{(j)}, \dots, o_{m}^{(j)}\bigr)$,
where each $o_{k}^{(j)}$ corresponds to a specific tool call (e.g., \texttt{WeatherAPI}, \texttt{BrowserTool}). Collecting all these fine-grained steps across sub-tasks yields a \textit{global DAG}, referred to as the \textbf{\ExecGraph}. It is denoted by $\mathcal{G} = (\mathcal{O}, E)$, where each node $o\in \mathcal{O}$ is a single tool invocation, and edges $(o_i \to o_k) \in E$ capture data dependencies. Since no node can depend on itself or on a future node, $\mathcal{G}$ is acyclic.
In addition to encoding data flow, the \ExecGraph{} also serves as a repository of intermediate outputs from agent actions, which downstream actions can subsequently utilize based on their dependencies.

\paragraph{Agent Context Protocols (\acp):}
\acp\ are structured mechanisms for coordination, communication and error-handling among a large set of heterogenous agents. They ensure that:
\begin{enumerate}[leftmargin=*,topsep=0pt, noitemsep]
    \item Agents can coordinate tool invocations in line with the \ExecGraph, storing and retrieving outputs for subsequent sub-tasks.
    \item Standardized message schemas (\texttt{AGENT\_REQUEST}, \texttt{AGENT\_RESPONSE}, \texttt{ASSISTANCE\_REQUEST}) govern information exchange between agents and tools.
    \item Fault tolerance is maintained via standardized error codes, so that sub-task failures or exceptions can be localized and addressed without collapsing the entire workflow.
\end{enumerate}

Overall, \acp s provide a robust foundation for execution of multi-step, multi-agent execution blueprints: each node (tool call) is invoked with a well-defined \texttt{AGENT\_REQUEST}, returns a structured \texttt{AGENT\_RESPONSE}, and flags failures via \texttt{ASSISTANCE\_REQUEST} messages when needed as described next.

\subsection{Implementational Details}
 
We leverage an agent to decompose the overall problem $T$ into sub-tasks and compile them into a global DAG, the \ExecGraph~ $\mathcal{G}$. 
Once execution starts, $\mathcal{G}$ remains the shared reference monitoring the overarching goal, intermediate inputs, outputs, and sub-task interdependencies.
The successful execution of $\mathcal{G}$, i.e., the completion of task $T$, is reliant on reliable execution of each agent action $o_i$. This reliability is facilitated by structured interaction protocols governing agent interactions with external tools, detailed next.

\paragraph{Execution Runtime.}
Each sub-task is assigned to an agent with the required capabilities. Execution proceeds in three main phases:

\begin{enumerate}[leftmargin=*, topsep=0pt, noitemsep]
    \item \textbf{AGENT\_REQUEST (Input Preparation).}
    The agent constructs a structured request specifying the \textit{method} and \textit{endpoint} (or function name), \textit{headers} as key--value pairs, and a \textit{body} with parameter names and values. These parameters may be:
    (a) \emph{LLM-generated}: produced by the agent’s internal reasoning (e.g., user queries or dynamically generated prompts), or (b) \emph{Tool-derived}: outputs from a previous node in $\mathcal{G}$.
    If any required input is missing or invalid, the agent raises an error (e.g., \texttt{601 MISSING\_REQUIRED\_PARAMETERS}), prompting an immediate \texttt{ASSISTANCE\_REQUEST}.

    \item \textbf{TOOL\_CALL (Execution).}
    The request is dispatched to the appropriate external tool (commonly, an API). Failures such as timeouts or runtime exceptions trigger a corresponding error code (e.g., \texttt{604 TOOL\_CALL\_FAILURE}) and an immediate \texttt{ASSISTANCE\_REQUEST}.

    \item \textbf{AGENT\_RESPONSE (Output Validation).}
    Once a raw response is received, the agent structures it into a \texttt{TOOL\_RESPONSE} that includes a status code, any relevant output variables, and any values on which subsequent sub-tasks depend. If critical fields are missing (\texttt{605 INCOMPLETE\_INFORMATION}) or the data is incorrect (\texttt{607 WRONG\_INFORMATION}), the agent issues an \texttt{ASSISTANCE\_REQUEST}. Validated output is stored in $\mathcal{G}$, allowing downstream sub-tasks to retrieve it.
\end{enumerate}

\paragraph{Fault Tolerance.}
Whenever an error arises, an \texttt{ASSISTANCE\_REQUEST} is posted, containing:
(1) a concise \texttt{STATUS\_UPDATE} on what has been done so far, (2) the specific standardized error code and a general description, and (3) a recommended resolution, such as retrying the call, switching to an alternative tool, or abandoning the sub-task.
A specialized fault-tolerance agent then updates $\mathcal{G}$ accordingly. If a viable workaround exists (e.g., a different tool), the sub-task is re-routed. Otherwise, the sub-task is marked as failed, ensuring unaffected parts of $\mathcal{G}$ can continue unobstructed.
Descriptions of all aforementioned error codes are provided in Table \ref{tab:error_codes} (Appendix \ref{app:status_codes}).

\paragraph{Final Coordination Layer.}
Depending on the broader objective, a final layer may aggregate \ExecGraph{} outputs into user-facing deliverables (e.g., structured reports, visualizations, or textual summaries). This layer compiles all validated data or transforms it further, according to $T$’s requirements.

\paragraph{Data Flow and Output Persistence.}
Each node’s outputs are stored in $\mathcal{G}$, making them available to downstream sub-tasks. This localizes failures: if one sub-task fails and posts an \texttt{ASSISTANCE\_REQUEST}, unrelated parts of the \ExecGraph\ proceed without interruption. In addition, the DAG dictates the sequence of sub-tasks, ensuring that each sub-task only starts once its prerequisites have run successfully.

\section{Evaluation and Analysis}

To effectively demonstrate and evaluate the capabilities of ACPs, we focus on highlighting three key advantages through our empirical evaluation:

\begin{enumerate}[leftmargin=*, noitemsep, topsep=0pt] 
\item \textbf{Complex \ExecGraph:} Our system supports intricate task structures, enabling the execution of complex workflows composed of numerous interdependent sub-tasks.

\item \textbf{Robust Error Handling:} The structured interaction mechanisms mitigate cascading failures and ensure sustained collective inference across long execution horizons.

\item \textbf{Modularity:} The ACP structure allows seamless integration of external domain-specialized tools, facilitating rapid extensibility to a wide range of tasks or domains without disrupting existing functionality.

\end{enumerate}

Our experimental evaluation validates and highlights these advantages across multiple tasks and domains. Each of the following studies emphasizes distinct aspects of the system:

\begin{itemize}[leftmargin=*, noitemsep, topsep=0pt] 

\item \textbf{Web Assistance Tasks (AssistantBench):} This benchmark evaluates the ability to perform long-horizon web navigation tasks, state-of-the-art performance on this empirically validates modularity, extensibility, and robustness.

\item \textbf{Multi-Modal Report Generation:} This experiment demonstrates the system’s capability for effective inter-agent coordination, focusing on collaborative workflows that produce structured, informative, multi-modal content. It particularly highlights how agent communication scales with increased content complexity.

\item \textbf{Information-Rich Dashboard Creation:} This case study includes ablations of the setup examining long-horizon coordination through the collaborative construction of interactive visual analytics dashboards.

\end{itemize}

\section{Web Assistance Tasks: AssistantBench}
\label{bench_apis}

\paragraph{Benchmark and Setup:}
AssistantBench \citep{yoran2024assistantbench} is a benchmark designed to evaluate how well AI agents can perform realistic, web-based tasks that require browsing, planning, and aggregating information. It includes a diverse set of 214 complex tasks such as travel planning, product comparison, and decision-making that require multiple steps and reasoning chains, serving as a strong benchmark to validate effectiveness.

The final coordination layer, specific for AssistantBench, is used to effectively manage and synthesize the outputs in a format expected by AssistantBench. This layer acts as a decision-maker that gathers all the intermediate results produced by the agents and returns the output in the desired format.

\begin{table}[htbp]
    \centering
    \label{tab:model_comparison}
    \rowcolors{3}{white}{white}  
    \resizebox{0.8\textwidth}{!}{
    \begin{tabular}{@{}l@{\hspace{0.5em}}c@{\hspace{0.5em}}c@{\hspace{0.5em}}c@{\hspace{0.5em}}c@{\hspace{0.5em}}c@{\hspace{0.5em}}c@{}}  
        \toprule
        \textbf{Model Name} & \textbf{Accuracy} & \textbf{Precision} & \textbf{EM} & \multicolumn{3}{c}{\textbf{Acc.}} \\
        \cmidrule(l){5-7}
        & & & & \textbf{(Easy)} & \textbf{(Medium)} & \textbf{(Hard)} \\
        \midrule
        \rowcolor{gray!15}
        \textbf{Ours: \acp{}} + Domain Agents (GPT-4o) & \textbf{28.30} & 30.0 & 11.0 & 67.8 & \underline{48.5} & \underline{15.5} \\
        \rowcolor{gray!15}
        \textbf{Ours: \acp{}} (GPT-4o) & 24.80 & 26.5 & 9.4 & \underline{81.6} & 38.9 & 13.5 \\
        Magentic-One (GPT-4o) & 25.30 & 25.3 & 11.0 & 69.9 & 35.6 & \textbf{16.9} \\
        Magentic-One (o1, GPT-4o) & \underline{27.70} & 29.0 & \underline{13.3} & 73.4 & 47.1 & 14.8 \\
        SPA-CB (Claude) & 26.40 & \textbf{32.2} & \textbf{13.8} & 81 & 44.6 & 13.3 \\
        SPA-CB (GPT-4T) & 25.20 & 27.5 & 9.9 & 80.7 & 42.7 & 12.4 \\
        Infogent (GPT-4o) & 14.50 & 20.4 & 5.5 & 63.9 & 19.3 & 8.4 \\
        CB-INST (GPT-4T) & 16.5 & 30.7 & 6.1 & 51.2 & 40.2 & 2.3 \\
        CB-1S (GPT-4T) & 22.2 & 24.8 & 8.3 & 67.8 & \textbf{49.7} & 4.2 \\
        RALM-INST (GPT-4T) & 11.8 & 19.5 & 5.5 & 50.2 & 17 & 6.2 \\
        RALM-1S (GPT-4T) & 10.7 & 22.4 & 3.9 & 80 & 10.5 & 5.5 \\
        SEEACT (GPT-4T) & 4.1 & 26.3 & 2.2 & 28.9 & 2.5 & 2.9 \\
        SPA (GPT-4T) & 11.1 & \underline{30.9} & 5.5 & 29.5 & 12.3 & 9.1 \\
        RALM-INST→CB (GPT-4T) & 18.7 & 19.9 & 6.6 & 57.8 & 34.7 & 8 \\
        RALM-1S→CB (GPT-4T) & 19.5 & 21.0 & 6.1 & 81.3 & 35 & 7.3 \\
        SEEACT→CB (GPT-4T) & 23.4 & 26.1 & 9.4 & \textbf{82} & 47.7 & 7.1 \\
        \bottomrule
    \end{tabular}
    }
\caption{
\textbf{Performance on AssistantBench.} Our framework achieves \textbf{SOTA accuracy} (28.30\%) with domain-specific \texttt{tools}. It demonstrates robust performance across difficulty levels—especially on medium (48.5\%) and hard (15.5\%) tasks—and shows a clear improvement over the base setup (24.80\%), validating the \textbf{Modularity} enabled by ACPs. Bolded values are the best scores, and underlined values are the next best.
}
\end{table}

\paragraph{Experiments:} {To empirically validate the modularity of our framework, we conducted two experiments on the AssistantBench benchmark. The first experiment restricted the agent capabilities to a minimal set, comprising of \texttt{MultiModalWebSearch} and \texttt{Calculator} \texttt{Tools} (see Table~\ref{bench_apis}). The second experiment extended this agent capabilities by integrating a set of domain-specific \texttt{Tools} selected for their ability to provide structured, high-precision information in areas such as location-based services and content retrieval.}

\paragraph{Result 1: \acp{}s are robust, modular, and extensible.} {By simply incorporating a handful of domain-specific \texttt{Tools}, we achieve a \textbf{state-of-the-art (SOTA) accuracy of 28.3\%}. This surpasses both domain-specific and generalist baselines, including those leveraging stronger underlying models. This underscores a core strength of our framework: \textbf{domain-specific capabilities can be added without modifying or retraining the core system.}}

\paragraph{Result 2: \acp{}s enable robust performance even with generic agents.} {Despite using only a basic toolset, the framework's communication runtime—anchored by ACPs and the Execution Blueprint—still supports coherent, complex task execution. This is evident in the base configuration’s results: an accuracy of 24.80\% on AssistantBench, which is on par with the existing SOTA's.}

Furthermore, the system demonstrates strong generalization across difficulty levels: 67.8\% on easy tasks, 48.5\% on medium tasks, and 15.5\% on hard tasks. These results emphasize not only the expressivity and scalability of the framework, but also its ability to adapt to increasingly complex workflows without sacrificing reliability.

\section{Multimodal Report Generation}
\label{sec:multi_modal_reports} 

\begin{figure}[ht]
  \centering
  \setlength{\tabcolsep}{-8pt}
  \resizebox{1\textwidth}{!}{ 
    \begin{tabular}{@{}*{5}{c}@{}}
      \includegraphics[page=1, width=0.21\textwidth]{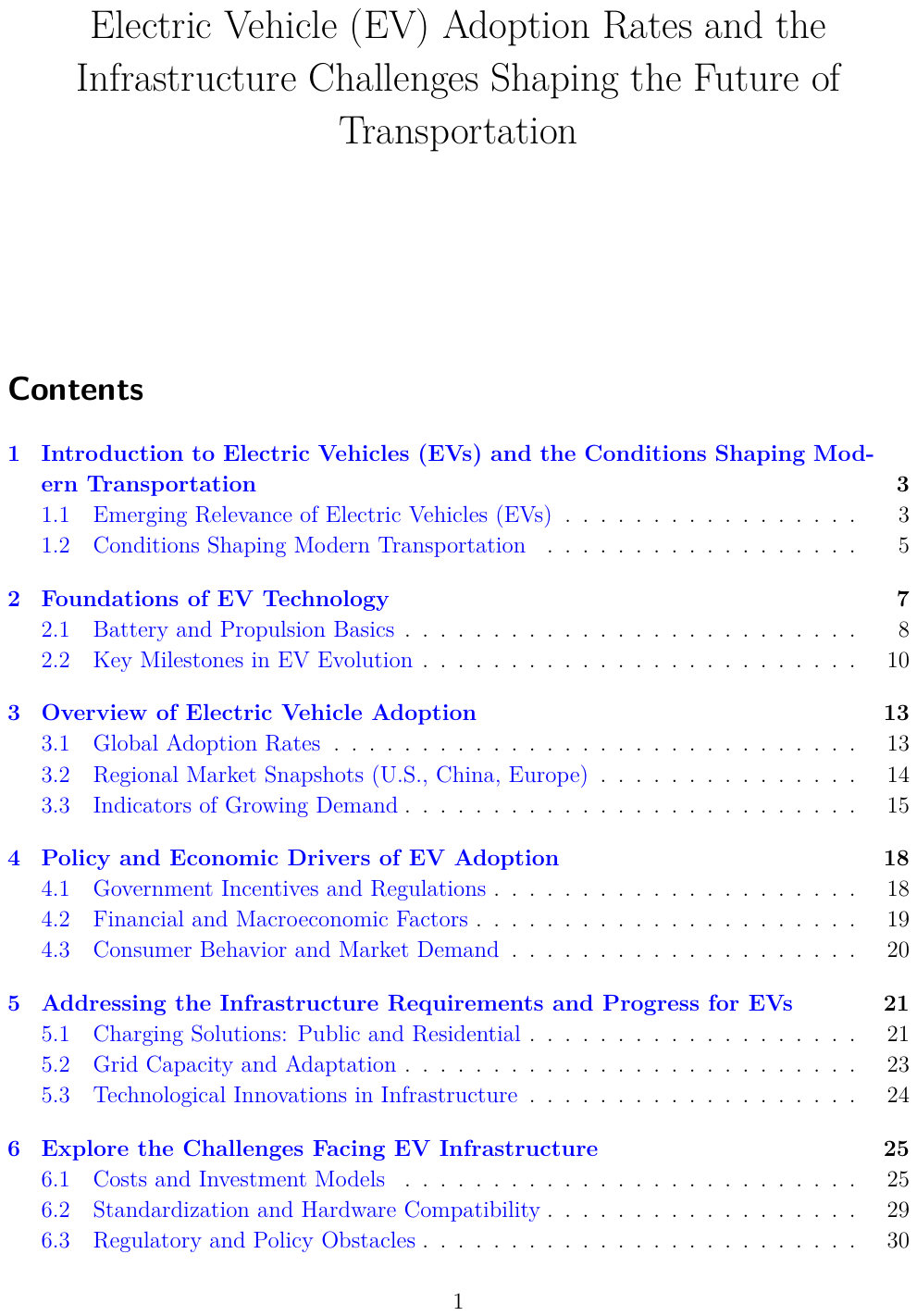} & 
      \includegraphics[page=3, width=0.21\textwidth]{figures/Report_ACP.pdf} & 
      \includegraphics[page=10, width=0.21\textwidth]{figures/Report_ACP.pdf} & 
      \includegraphics[page=13, width=0.21\textwidth]{figures/Report_ACP.pdf} & 
      \includegraphics[page=53, width=0.21\textwidth]{figures/Report_ACP.pdf}
    \end{tabular}
  }
  \caption{Sample pages from a multi-modal technical report generated by our ACP-based framework. Textual content, data visualizations, and structured references are combined into a cohesive document spanning multiple sections and \textgreater{} 30 pages. All reports can be found in Appendix \ref{appendix:report_samples}.}
\end{figure}

One of the primary advantages of our framework is its capacity to produce structured, 
informative, and multi-modal content by orchestrating the coordinated actions of multiple 
agents. This capability is particularly important for real-world scenarios where diverse 
data sources---including textual, graphical, and tabular information---must be integrated 
into a coherent narrative.

\paragraph{Setup.}
In this study, agents' capabilities comprised of two key tools: 
\texttt{BrowserTool} for retrieving up-to-date information from the web, 
and \texttt{PlotVisualizationTool} for generating plots or charts based on 
queried data. To synthesize outputs from these agents into a single 
well-organized report, we added a final coordination layer responsible 
for merging and structuring the content. This layer ensures that the final 
report captures all essential elements (textual information, data visualizations, 
and relevant citations) in a cohesive manner.

\paragraph{Evaluation Data.}
To thoroughly assess the performance of our system, we generated reports 
across five distinct domains: 
\emph{Finance}, \emph{Technology}, \emph{Healthcare}, \emph{Automobile}, 
and \emph{Real Estate} (more details regarding topics of the report in Appendix \ref{appendix:report_queries}). Specifically, we constructed one multi-modal 
report per domain, ensuring that each report leveraged relevant external 
data and visualizations to address domain-specific details and analyses. 
All results were benchmarked against two alternative report-generation 
models -- Perplexity Deep Research \citep{perplexity} and Gemini Deep Research \citep{gemini}, 
selected because they were the only freely accessible research-focused systems available for 
direct comparison. The complete reports for all methods can be found in Appendix \ref{appendix:report_samples} (and one multi-modal report generated through our (ACP) method can be found in Appendix \ref{appendix:real_report_sample}).

\paragraph{Human Evaluation.}
To evaluate the quality of the generated reports, we conducted a human study with 15 human reviewers.
For this the reviewers were asked to rate each report on six key dimensions:
(D1) \textbf{Coverage} (breadth of topics addressed), (D2) \textbf{Relevance} (pertinence to the specified domain or prompt), (D3) \textbf{Trustworthiness} (credibility and relevance of citations), (D4) \textbf{Clarity and Organization} (structure and readability), (D5) \textbf{Depth of Analysis} (level of insight or reasoning), and (D6) \textbf{Presentation Quality} (visual and stylistic appeal).
Reviewers provided a numeric score on a 0--5 scale, with 0 indicating the lowest rating 
(\emph{Severely Lacking}) and 5 the highest rating (\emph{Excellent}) for each category. This methodology allows us to 
quantitatively compare the overall performance of our framework against the baseline 
systems, and to identify key areas of strength as well as potential opportunities for 
further improvement.

\begin{figure}[!h]
    \centering
    \includegraphics[width=1.1\textwidth]{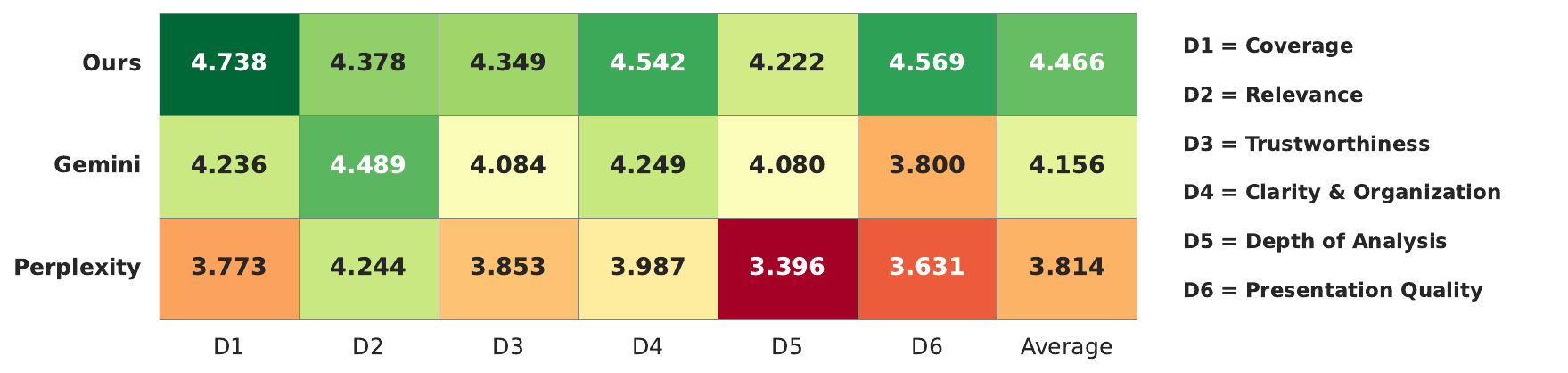}
    \caption{Heatmap of the average human ratings (0–5) across six key dimensions (D1–D6) and the overall average for multi-modal report generation. “Ours”: ACP (top row) outperforms Gemini and Perplexity, showing particularly strong gains in Coverage (D1) and Presentation Quality (D6).
    }
    \label{fig:report_heatmap}
\end{figure}

\paragraph{Results and Analysis.}
Figure~\ref{fig:report_heatmap} shows the average human ratings on six dimensions 
(D1--D6) and also includes an overall average. Our framework (top row) obtains the 
highest average score, surpassing both \texttt{Gemini Deep Research} and 
\texttt{Perplexity Deep Research} across the board. In particular, we are 
significantly higher in \emph{Coverage} (D1) and \emph{Presentation Quality} (D6), 
while remaining highly competitive in the remaining dimensions. Moreover, the final 
ACP-based reports are consistently enriched with multiple data 
visualizations, highlighting the system’s capacity for extensive coverage. This 
advantage arises from leveraging multiple agents through ACP, which effectively merges 
text, data visualizations, and citations into a cohesive report. By systematically 
coordinating tool outputs and preserving context, ACP helps reduce error cascades 
and delivers coherent multi-modal outputs even in lengthy and complex scenarios.
Moreover, achieving such comprehensive coverage and presentation quality 
required close 
to 150 communication messages among agents. This highlights ACPs robustness for long-horizon coordination mechanisms despite the elevated complexity.

\section{Case Study: Information-Rich Dashboard Creation}

\paragraph{Dataset:} We construct a synthetic dataset of user queries stratified into three complexity levels, with Level 1 (15\%) comprising queries requiring usage of 2-3 tools with moderate interdependencies, Level 2 (25\%) requiring 3-5 tools with dynamic relationships and complex reasoning, and Level 3 (60\%) containing highly complex queries. Level 3 was further subdivided into Type 1 (40\%) with deep, multi-step queries requiring extensive coordination, and Type 2 (20\%) as a combination of multiple Level 2 queries. We defer additional details regarding the creation of the dataset to Appendix \ref{appendix:dashboard_dataset}

\subsection{Ablation Study: The Importance of Coordination and Fault Tolerance}

We consider two ablations of our setup. In the first the decomposition of task $T$ into sub-tasks $\{\tau_1, \tau_2, \dots, \tau_n\}$ is provided to the team of agents (each agent is assigned a sub-task), however, the ability to ask for assistance is not provided, and neither the dependencies between the tasks; this setup is called \textit{No Assistance} baseline.
In the second, all the tools are provided to a single agent, rather than a team of agents. A singular ReAct \citep{yao2023react} must aim to resolve the user query, referred to as the \textit{Single Agent} baseline. 

The comparison between these baselines and our full framework allowed us to isolate the benefits of task decomposition (Single Agent vs. No Assistance) and structured coordination via Execution Blueprint, Structured Messages, Assistance Request, and Error Codes (No Assistance vs. ACPs).

To assess the quality of generated dashboards, we conducted a human evaluation study with 10 human evaluators. Each evaluator was presented with dashboards created by the three approaches and asked to rate them according to the criteria outlined in Table~\ref{tab:score_ratings}. To control for bias, evaluators were not informed which system produced which dashboard.

\begin{table}[ht]
\centering
\begin{tabular}{lcccc}
\toprule
\textbf{Setup} & \textbf{Overall} & \textbf{L1} & \textbf{L2} & \textbf{L3} \\
\midrule
Single Agent & 1.96 & 2.13 & 2.32 & 1.77 \\
No Assistance & 2.94 & 2.60 & 2.80 & 3.08 \\
\rowcolor{gray!20} ACP (Ours) & \textbf{3.95} & \textbf{4.00} & \textbf{4.20} & \textbf{3.83} \\
\bottomrule
\end{tabular}
\caption{Evaluation scores across system variants. Coordination and fault-tolerance via ACP yield substantial improvements over the No Assistance and Single Agent Baselines.}
\label{tab:evaluation_ratings}
\end{table}

\textbf{Task decomposition, structured coordination, and global communication collectively drive substantial performance gains.} Decomposing complex tasks into sub-tasks significantly improves system performance---raising scores from a low \textbf{1.96} with a \textit{Single Agent} baseline to \textbf{2.94} with \textit{Multi-Agent} decomposition that is present in the No Assistance Baseline. However, true gains are unlocked only when decomposition is paired with structured coordination via our \textbf{ACP}-based framework, achieving a \textbf{3.95} overall score. This leap stems from robust error-handling protocols---like standardized \texttt{ASSISTANCE\_REQUEST}s and \texttt{ERROR\_CODES}---that drastically reduce critical failures, enabling 50\% of outputs to score a perfect \textbf{5}. Moreover, the \textit{Execution Blueprint} ensures dependency-aware, coherent execution of long-horizon workflows, with particularly strong results on \textit{Level 3} tasks (\textbf{3.83}), where consistency and coordination are critical. Together, these components validate that modularity, fault tolerance, and structured communication are essential for reliable multi-agent systems tackling real-world complexity.

\subsection{Agent Execution Timeline}

\begin{figure}[b]
    \centering
    \includegraphics[width=1\linewidth]{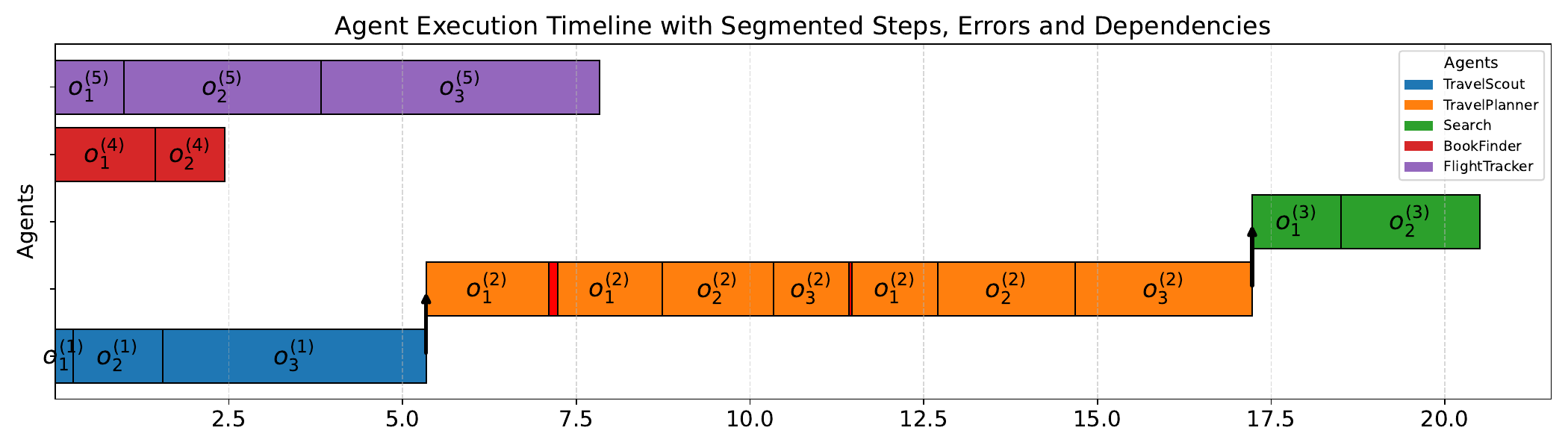}
\caption{
        \textbf{Execution timeline in response to a travel planning query.}
        This figure illustrates the execution timeline for a complex query, depicting parallel execution of independent agents and sequential execution of dependent agents.
    }
    \label{fig:agent_execution_Plot}
\end{figure}
To highlight how ACPs enable reliable execution of long-horizon workflows, we present the execution timeline of a complex travel planning query in Figure~\ref{fig:agent_execution_Plot}. The task—comprising weather checks, cultural recommendations, flight searches, and location-based suggestions—is automatically decomposed and distributed across five specialized agents. Agents operate both in parallel and in sequence. For example, weather and book recommendations run concurrently with flight searches. Once preliminary data is gathered, a final coordination layer integrates outputs to recommend themed cafés and cultural events. When an error occurs—due to missing weather attributes required for downstream reasoning—the system triggers an \texttt{ASSISTANCE\_REQUEST}, re-planning the failed step. After recovery, execution continues seamlessly with other agents, such as identifying museums or restaurants near selected events. This timeline underscores the strengths of ACPs: enabling modular execution, structured communication, and graceful recovery from errors. By supporting both breadth (parallel subtasks) and depth (long sequential dependencies), ACPs ensure robust and coherent multi-agent coordination.
\section{Related Work}
Recent efforts have explored mechanisms for coordination and collective behavior in multi-agent systems. Fang and Kress-Gazit~\citep{fang2024high} propose a task grammar using Linear Temporal Logic (LTL) to support collaboration among heterogeneous agents without predefined task assignments. RETSINA~\citep{sycara1996distributed} earlier introduced a distributed multi-agent framework where interface, task, and information agents communicate asynchronously for dynamic problem-solving. \citep{kaufmann2021active} formalize collective intelligence via the Active Inference Framework, showing how shared goals and theory-of-mind reasoning can yield emergent global behavior.

Building on this foundation, recent surveys on large language model (LLM) multi-agent systems focus on system architectures, agent roles, communication strategies, and tool use~\citep{guo2024large}, or classify collaboration mechanisms by organizational structure and interaction strategy~\citep{tran2025multi}. Several frameworks have emerged~\citep{autogen2023,marro2024scalable}, though their protocols remain comparatively simple. Previously, some works have emphasized fault-tolerant multi-agent execution~\citep{huang2024resilience}, self-reflective error handling~\citep{shinn2023reflexion}, advanced planning~\citep{erdogan2025planact}, and challenges in long-horizon tasks~\citep{chen2024travelplanner}. By contrast, our Agent Context Protocols (ACPs) introduce structured message schemas (e.g., \texttt{AGENT\_REQUEST}), standardized error handling, and a persistent global DAG (Execution Blueprint) for execution, enabling scalable, interpretable, and long-horizon workflows.

In parallel, multiple works highlight the benefits of explicitly structured communication \citep{li2023camel,shen2023hugginggptsolvingaitasks,hong2023metagpt,qian2024chatdevcommunicativeagentssoftware,nguyen2024dynasaur,fourney2024magenticone}, illustrating how conversation-based planning, role-playing paradigms, hierarchical orchestration, or on-the-fly action creation can align agents across diverse tasks. While these approaches demonstrate the value of structured coordination, they often rely on partial or domain-specific standards. Our ACP framework aims to unify these efforts with a single, extensible protocol for robust, long-horizon collaboration across heterogeneous agents.
\section{Discussion and Conclusion}

This work introduced Agent Context Protocols (ACPs), a set of structured protocols for agent-agent communication, coordination, and error handling that enhance collective inference among teams of LLM agents.
Our empirical evaluations demonstrated ACPs' robust performance across varied complex tasks, achieving state-of-the-art accuracy on a challenging long-horizon web assistance benchmark, and excelling at multimodal technical report generation—where human evaluators rated ACP-generated reports superior to outputs from commercial models in content coverage, relevance, and presentation quality.

The modular and extensible nature of ACPs facilitates effortless integration of domain-specialized tools, enabling rapid adaptation to diverse and complex real-world scenarios. Overall, ACPs establish a vital standard for effective communication and coordination in multi-agent AI systems, significantly improving reliability and fault tolerance across extensive execution workflows. This opens promising directions for future research, including investigating ACPs' scalability to larger agent populations, leveraging advanced reasoning models for execution planning, and broadening the applicability of ACP-enabled collective intelligence in dynamic, evolving environments.

\newpage
\bibliography{references}
\bibliographystyle{colm2025_conference}

\newpage
\appendix
\section{Status Codes and Their Descriptions}
\label{app:status_codes}
\begin{table}[ht]
    \centering
    \label{tab:error_codes}
    {\small
    \resizebox{0.99\textwidth}{!}{%
    \begin{tabularx}{\textwidth}{@{}lX@{}}
        \toprule
        \textbf{Error Code} & \textbf{Description} \\
        \midrule
        \textbf{Tool Call Request Stage} & \\[2pt]
        601 MISSING\_REQUIRED\_PARAMETERS 
          & Some required parameters are missing in the sub-task specification for the tool call. \\[2pt]
        602 WRONG\_STEP\_DETAILS 
          & Agent step details are incorrect or incomplete (e.g., invalid parameters or mismatched input format). \\[2pt]
        603 INVALID\_PARAMETER\_USAGE  
          & A parameter is used improperly (e.g., multiple values where only one is accepted, or an invalid format). \\
        \midrule
        \textbf{Tool Call Stage} & \\[2pt]
        604 TOOL\_CALL\_FAILURE 
          & The tool call failed during execution (e.g., network issues or unexpected runtime errors). \\
        \midrule
        \textbf{Tool Output Extraction Stage} & \\[2pt]
        605 INCOMPLETE\_INFORMATION 
          & The tool’s response lacks essential data for the current agent action step. \\[2pt]
        606 DEPENDENCY\_INCOMPLETE\_INFORMATION 
          & The tool’s response is missing critical data needed by future dependent sub-tasks. \\[2pt]
        607 WRONG\_INFORMATION 
          & The tool’s response is entirely irrelevant or erroneous for the intended task. \\
        \bottomrule
    \end{tabularx}
    
    }
    }
    \caption{
    Summary of error codes and their respective stages in the tool-invocation pipeline.}
\end{table}

\section{AssistantBench}

We evaluate our framework against a comprehensive set of baselines from the AssistantBench benchmark~\cite{yoran2024assistantbench}, which spans a variety of architectural paradigms for web-based agents. These include closed-book language models (\textbf{CB-1S} \citep{press2022measuring}, \textbf{CB-INST} \citep{yao2023react}) that rely solely on internal knowledge, retrieval-augmented models (\textbf{RALM-1S}, \textbf{RALM-INST}, \citep{trivedi2022interleaving}) which access external information to support reasoning, and hybrid pipelines that combine browsing agents like \textbf{SEEACT} \citep{zheng2024gpt} or retrieval agents with closed-book LLMs (\textbf{SEEACT$\rightarrow$CB}, \textbf{RALM-1S$\rightarrow$CB}, \textbf{RALM-INST$\rightarrow$CB}) \citep{yoran2024assistantbench}. We also compare against \textbf{SPA} (SeePlanAct) \citep{yoran2024assistantbench}, a planning-augmented version of SEEACT, and \textbf{Magentic-One} \citep{fourney2024magenticone}, a multi-agent orchestration system. 

Empirically, our ACP-based framework outperforms all baselines on overall accuracy, achieving a state-of-the-art score of 28.3\%—surpassing both generalist systems like Magentic-One and specialist planners like SPA. Notably, our system demonstrates strong performance across all difficulty levels (67.8\% on easy, 48.5\% on medium, and 15.5\% on hard), validating the benefit of structured communication, modularity, and fault-tolerance offered by ACPs.

\subsection{Assistant Bench Tools}
\renewcommand{\arraystretch}{1.5}
\begin{longtable}{p{4.5cm} p{8.5cm}}
\toprule
\textbf{Tool/API Name} & \textbf{Description} \\
\midrule
\endfirsthead

\multicolumn{2}{l}{\textit{(Continued from previous page)}} \\
\toprule
\textbf{Tool Name} & \textbf{Description} \\
\midrule
\endhead

\multicolumn{2}{r}{\textit{(Continued on next page)}} \\
\endfoot

\bottomrule
\caption{Tools used for AssistantBench along with their descriptions}
\label{tab:combined_tools}  \\
\endlastfoot
CalculatorTool & This function acted as a LLM based calculator agent to get precise answers for the calculation to be performed as described by the user. This agent uses Python coding for performing the calculations. It's input is a text based query and output is a text corresponding to the query.\\
MultiModalWebSearch & This tool is a LLM based agentic system that performs a multimodal web search by retrieving and synthesizing information from multiple sources, including text and images. It can answer general knowledge queries, retrieve structured data, and provide image-based responses when required. It's input is a text based query and output is a text corresponding to the query.\\
FileSurfer & This tool analyzes and processes file-related queries by retrieving, matching, and referencing file data. It is useful for structured file analysis, such as extracting content, summarizing documents, and identifying patterns in stored information. It's input is a text based query and output is a text corresponding to the query.\\
CodingTool & This tool provides LLM based code generation, debugging, and enhancement capabilities. It can write, analyze, and improve code snippets based on natural language queries, making it useful for developers seeking coding assistance. It's input is a text based query and output is a text corresponding to the query.\\
TripadvisorSearchLocation & This API is used to search for location details (like a city or region) based on a query string, providing results limited to locations' names, types, and geoIDs, which are used by other TripAdvisor APIs. \\
TripadvisorSearchHotels & This API retrieves available hotels in a specified location (using the Geo ID retrieved from TripadvisorSearchLocation) along with their details like price, rating, and amenities for a given check-in and check-out date. \\
TripadvisorSearchRestaurants & This API is used to search for restaurants in a specific location by providing a location's geoID (retrieved from TripadvisorSearchLocation), returning details such as restaurant name, rating, reviews, and other relevant information. \\
GoodreadsSearchBook & This API is used to search for books based on a specific keyword, returning details like the book's title, author, and ratings. \\
GoodreadsSearchQuotes & This API allows user to search for quotes based on a keyword, returning the quote text, author, and number of likes. \\
GoodreadsGetAuthorsBooks & This API retrieves a list of books written by a specific author using the author's ID. \\
SkyScrapperFlightSearch & This API retrieves available flights for a given route by providing details like origin and destination airports, travel dates, and optional filters like cabin class and carriers. The skyId from SkyScrapperSearchAirport is required for this to run. \\
SkyScrapperSearchAirport & This API searches for airports by location name, returning airport details such as the name and unique airport identifier (skyId). \\
GoogleMapsNearbySearch & Search for places within a specified area by latitude and longitude using the Google Maps Places Nearby Search API. It retrieves detailed information such as the name, location coordinates, and types of the place. \\
GoogleMapsTextSearch & Perform text-based searches for places using the Google Maps Places Text Search API. It returns details including the name, full address, and geometry of the place based on a query string. \\
GoogleMapsAutocomplete & Generate query predictions for geographic searches with the Google Maps Places Autocomplete API. It provides suggestions along with place IDs and descriptions to aid in location-based searches. \\
SocialMediaInfluencerAPI & Discover detailed insights on social media influencers across platforms like YouTube, Instagram, and TikTok using the SocialMediaInfluencerAPI. It provides metrics such as follower counts, engagement rates, content overviews, and profile images. \\
\end{longtable}

\section{Multimodal Reports}

\subsection{Detailed Domain Queries for Multi-Modal Report Generation}
\label{appendix:report_queries}

Below are the five representative queries used to evaluate our framework’s ability to generate comprehensive, multi-modal reports in distinct domains. Each query requires integrating textual data, relevant web resources, and (where applicable) data visualizations via tools such as \texttt{BrowserTool} and \texttt{PlotVisualizationTool}. Further details on how these queries were processed and benchmarked against other report-generation systems can be found in Section~\ref{sec:multi_modal_reports}.

\begin{itemize}
    \item \textbf{Finance:}\\
    ``Write a comprehensive and well-researched report on the rise of Decentralized Finance (DeFi) and how it challenges traditional banking models and regulations.''

    \item \textbf{Technology:}\\
    ``Write a comprehensive report on AI-driven automation and its influence on workforce productivity, cost savings, and job displacement.''

    \item \textbf{Healthcare:}\\
    ``Write a detailed report on building resilient healthcare supply chains, drawing lessons from the COVID-19 crisis and proposing future strategies.''

    \item \textbf{Automobile:}\\
    ``Write a comprehensive report on electric vehicle (EV) adoption rates and the infrastructure challenges shaping the future of transportation.''

    \item \textbf{Real Estate:}\\
    ``Write a comprehensive and well-researched report on the shift from urban to suburban living trends and how remote work is influencing the real estate market, property values, and urban planning.''
\end{itemize}

These prompts were chosen to cover a broad range of subject matter expertise and to highlight how our multi-agent system uses different tools to produce information-rich reports.

\subsection{Report Generation Tools}
\renewcommand{\arraystretch}{1.5}
\begin{longtable}{p{4.5cm} p{8.5cm}}
\toprule
\textbf{Tool Name} & \textbf{Description} \\
\midrule
\endfirsthead

\multicolumn{2}{l}{\textit{(Continued from previous page)}} \\
\toprule
\textbf{Tool Name} & \textbf{Description} \\
\midrule
\endhead

\multicolumn{2}{r}{\textit{(Continued on next page)}} \\
\endfoot

\bottomrule
\caption{Tools used for MultiModal Report Generation along with their descriptions}
\label{tab:two_tools_concise} \\
\endlastfoot

PlotVisualizationTool 
& 
It is an LLM based agent which generates basic visualizations (plots or images) in response to user queries. Useful for adding succinct charts or graphs that complement textual content. 
\\

\midrule

BrowserTools
&
It is a LLM based agent which performs web searches and synthesizes relevant information from multiple sources. Also supports lightweight reasoning or coding tasks to provide a cohesive answer.
\\

\end{longtable}

\subsection{Generated Multimodal Report Samples}
\label{appendix:report_samples}
All the reports generated for these domains, for all three models (Ours (ACP), Gemini, and Perplexity) can be found in:
\url{https://anonymous.4open.science/r/ACP_Reports-C701/}.

\section{Dashboard Creation}

\subsection{Human Evaluation}

\begin{table}[H]
\label{tab:score_ratings}
\centering
\renewcommand{\arraystretch}{1.5}
\resizebox{\textwidth}{!}{
\begin{tabular}{p{2.2cm} p{14cm}}
\toprule
\textbf{Score} & \textbf{Description} \\
\midrule
Score 1 & No output is retrieved. Either the response is empty or there is error in all parts of the response. \\
Score 2 & The response is incomplete, with less than half of the parts of the query answered. \\
Score 3 & The response is incomplete, with more than half of the parts of the query answered. \\
Score 4 & Comprehensive and accurate information is retrieved for all parts of the query. Clear connections are drawn between the outputs, with some effort to synthesize the results into meaningful recommendations. \\
Score 5 & The query requirements are fully met with detailed, synthesized information. Insightful connections between the outputs are demonstrated, showcasing a high level of understanding and execution. One factor to differentiate between a score of 4 and 5 could be to see whether the information across the parts of the queries is consistent.\\
\bottomrule
\end{tabular}
}
\caption{Human Evaluation Scoring Criteria. 
}
\end{table}


\subsection{Dataset Generation for Dashboard Creation}
\label{appendix:dashboard_dataset}
To ensure diversity and relevance, we first created a seed set of representative queries for each level. We then employed LLMs to generate additional queries based on these examples and the available Tools. All generated queries underwent manual review to confirm their feasibility with our available tool set.

\subsection{Example Queries by Level}
Level 1 queries included ``Find restaurants in Rome with a rating above 4.5, check if the weather is good for outdoor dining this weekend, and recommend the top 3 options'' and ``Find the top Reddit posts in r/technology this week and related news articles.''

Level 2 queries were more complex, such as ``Find the top travel destinations this winter based on weather, trending news articles, and popular Reddit discussions. Recommend hotels and restaurants for one destination of your choice.''

Level 3 (Type 1) queries required deep multi-step reasoning, exemplified by ``Find an author whose books have had a significant sales increase in the past month on Goodreads, then analyze if this correlates with any recent news events, Reddit discussions, or quoted passages going viral. If there's a correlation with news events, show me other authors who experienced similar sales patterns when comparable news events happened in the past. Compare the social media engagement patterns between these cases.''

Level 3 (Type 2) queries combined multiple complex tasks, such as ``Find the top travel destinations this winter based on weather, trending news articles, and popular Reddit discussions. Recommend hotels and restaurants for one destination of your choice. Simultaneously, identify books trending on Goodreads this month, correlate their sales increase with viral quotes or Reddit discussions, and determine if any current news events influenced this trend.''

\subsection{DashBoard Tools}
\renewcommand{\arraystretch}{1.5}
\begin{longtable}{p{4.5cm} p{8.5cm}}
\toprule
\textbf{Tool/API Name} & \textbf{Description} \\
\midrule
\endfirsthead

\multicolumn{2}{l}{\textit{(Continued from previous page)}} \\
\toprule
\textbf{Tool Name} & \textbf{Description} \\
\midrule
\endhead

\multicolumn{2}{r}{\textit{(Continued on next page)}} \\
\endfoot

\bottomrule
\caption{Tools Used for Dashboard Creation along with their Descriptions}
\label{tab:combined_tools}  \\
\endlastfoot

Perplexity & As a web search engine to retrieve and synthesize information from multiple sources into a single, concise response. \\
TripadvisorSearchLocation & This API is used to search for location details (like a city or region) based on a query string, providing results limited to locations name, type, and geoID which is used by other TripAdvisor APIs. \\
TripadvisorSearchHotels & This API retrieves available hotels in a specified location (using the Geo ID retrieved from TripadvisorSearchLocation) along with their details like price, rating, and amenities for a given check-in and check-out date. \\
TripadvisorSearchRestaurants & API is used to search for restaurants in a specific location by providing a locations geoId (geoID retrieved from TripadvisorSearchLocation), returning details such as restaurant name, rating, reviews, and other relevant information. \\
NewsAPISearch & API allows you to search for news articles on specific topics by providing a search query and optional filters like language and region. \\
RedditTopPostsBySubreddit & API allows you to retrieve the top posts from a specific subreddit for a selected time period (e.g., hour, day, month). \\
GoodreadsSearchBook & This API is used to search for books based on a specific keyword, returning details like the books title, author, and ratings. \\
GoodreadsSearchQuotes & This API allows you to search for quotes based on a keyword, returning the quote text, author, and number of likes. \\
GoodreadsGetAuthorsBooks & This API retrieves a list of books written by a specific author using the authors ID. \\
WeatherAPIRealtimeWeatherApi & This API provides real-time weather information for a specific location based on parameters like city name, postal code, or coordinates. \\
WeatherAPITimeZoneAPI & This API retrieves the time zone information for a given location using query parameters like city, zip code, or coordinates. \\
WeatherAPIForecastWeatherAPI & This API fetches weather forecasts for a specified location, including options for the number of forecast days and specific languages. \\
SkyScrapperFlightSearch & This API retrieves available flights for a given route by providing details like origin and destination airports, travel dates, and optional filters like cabin class and carriers. skyId from SkyScrapperSearchAirport is required for this to run. \\
SkyScrapperSearchAirport & This API searches for airports by location name, returning airport details like the name and unique airport identifier (skyId). \\

\end{longtable}

\section{Execution Traces with ACPs: An Example}

\begin{figure}[H]
    \centering
    \caption{\textbf{Execution Blueprint}
This diagram illustrates a structured Execution Blueprint visualization, where specialized agents—each executing specific Tool calls (e.g., Reddit, News, Tripadvisor, WeatherAPI, Goodreads)—collaborate via Agent Context Protocols (ACPs) to decompose and execute complex tasks.}
    \label{tab:method_performance}
    \includegraphics[width=1\linewidth]{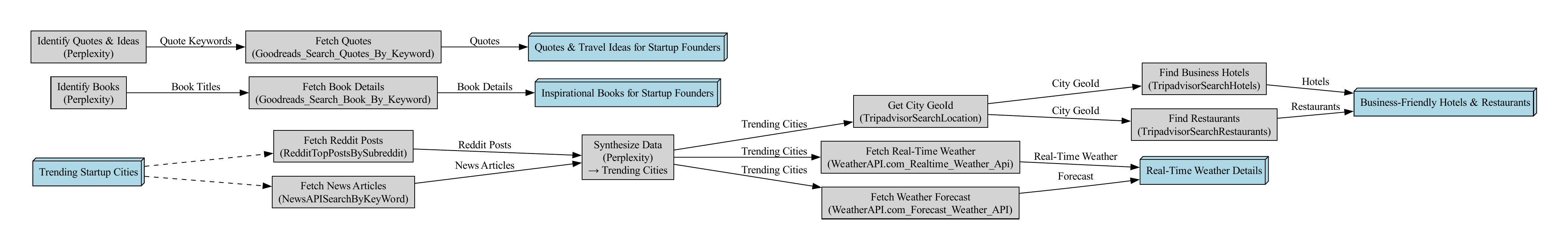}
\end{figure}

\begin{fancybox}[Example of \texttt{ASSISTANCE\_REQUEST}]
This example demonstrates how an agent encounters a missing input issue during a sub-task and raises an Assistance Request using standardized error codes and structured reasoning to recover gracefully. \\

\textbf{\$\$CHAIN\_OF\_THOUGHT\$\$}

\begin{itemize}
  \item I reviewed the workflow for Sub-Task 1, which aims to display average temperatures of vacation spots in India.
  \item In Step 1, the Perplexity API successfully provided \texttt{vacation\_spots\_list} containing the names of the top vacation spots.
  \item In Step 2, the Open-Meteo API is supposed to retrieve average temperature data.
  \item The Open-Meteo API requires latitude and longitude as input parameters.
  \item The input variables for Step 2 only include \texttt{vacation\_spots\_list} and \texttt{daily\_parameters}.
  \item The required latitude and longitude parameters are missing.
  \item Without these parameters, the Open-Meteo API cannot retrieve the weather data.
  \item We need to obtain latitude and longitude for each vacation spot, possibly by adding an intermediate step.
  \item Reminder to not use API output specifics while explaining the error in Assistance Request.
\end{itemize}

\vspace{1em}
\textbf{\$\$STATUS\_UPDATE\$\$}

\textbf{Progress:}
\begin{itemize}
  \item \textit{Previous Progress:} Successfully executed Step 1 of the workflow for Sub-Task 1.
  \item \textit{Current Progress:} Attempted to execute Step 2 using the Open-Meteo API to retrieve weather data.
\end{itemize}

\textbf{Current Step:} Sub-Task 1, Step 2

\textbf{Completed APIs:}
\begin{itemize}
  \item \texttt{Perplexity}: Retrieved a list of top vacation spots in India.
  \item \texttt{vacation\_spots\_list} is ready for use in subsequent steps.
\end{itemize}

\textbf{Encountered Issues:}
\begin{itemize}
  \item Open-Meteo API call failed due to missing latitude and longitude parameters.
\end{itemize}

\vspace{1em}
\textbf{\$\$ASSISTANCE\_REQUEST\$\$}

\textbf{Error Type:} 601 MISSING\_REQUIRED\_PARAMETERS \\
\textbf{Error Step:} Sub-Task 1, Step 2 \\
\textbf{Error API:} Open-Meteo

\textbf{Error Description:}
\begin{itemize}
  \item The Open-Meteo API requires latitude and longitude parameters, which are missing from the input variables.
\end{itemize}

\textbf{Relevant Context:}
\begin{itemize}
  \item The \texttt{vacation\_spots\_list} contains names but not coordinates.
  \item Without coordinates, we cannot fetch weather data.
\end{itemize}

\textbf{Suggested Resolution:}
\begin{itemize}
  \item Add a step to obtain latitude and longitude for each vacation spot, possibly using the Perplexity API.
\end{itemize}

\end{fancybox}

\begin{fancybox}[Example of \texttt{AGENT\_REQUEST}]
This example illustrates how the agent constructs structured requests to external tools—in this case, Perplexity AI function—to fetch missing latitude and longitude data for vacation spots. \\

\textbf{\$\$CHAIN\_OF\_THOUGHT\$\$}

\textbf{Input Variables:}
\begin{itemize}
  \item \texttt{Name:} \texttt{vacation\_spots\_list}
  \begin{itemize}
    \item \texttt{Parameter:} \texttt{query}
    \item \texttt{Type:} string
    \item \texttt{Source:} \texttt{TOOL\_Output (Sub-Task 1, Step 1)}
    \item \texttt{Description:} List of vacation spots from Sub-Task 1
    \item \texttt{Value:} \texttt{["The Dolomites, Italy", "Santorini, Greece", "Prague, Czech Republic", "Cinque Terre, Italy", "Barcelona, Spain", "London, UK", "Rome, Italy", "Amsterdam, Netherlands", "Munich, Germany", "Dublin, Ireland"]}
  \end{itemize}
\end{itemize}

\textbf{Verification of Input Variables:}
\begin{itemize}
  \item The \texttt{query} parameter is required and expects a string representing the user's question or search term.
  \item The \texttt{vacation\_spots\_list} is a list of vacation spot names.
  \item Since the \texttt{query} parameter accepts a single string, and the API recommends not to ask too much information in one search, we should execute separate API requests for each vacation spot.
\end{itemize}

\textbf{Determining Single vs. Multiple API Requests:}
\begin{itemize}
  \item The API documentation advises to break down queries and not ask for too much information in one search.
  \item Therefore, we will execute individual API requests for each vacation spot to ensure accurate and concise responses.
  \item We are keeping the number of API requests below 5 by getting latitude and longitude data of multiple cities at once.
\end{itemize}

\textbf{Method \& Function Name:}
\begin{itemize}
  \item \texttt{Method: FUNCTION}
  \item \texttt{Function Name: perplexity\_api\_response}
\end{itemize}

\textbf{Headers:}
\begin{itemize}
  \item No headers are required for this function call.
\end{itemize}

\textbf{Request Body Strategy:}
\begin{itemize}
  \item For each vacation spot, we will construct a \texttt{query} asking for the latitude and longitude of that specific place.
  \item The \texttt{preplexity\_ai\_key} is provided as \texttt{"YOUR\_API\_KEY"}.
\end{itemize}

\textbf{Final Decision:}
\begin{itemize}
  \item Create ten separate agent requests, one for each vacation spot.
  \item Each request will have its own \texttt{query} parameter formatted as \texttt{"What are the latitude and longitude of [vacation spot]?"}
\end{itemize}

\vspace{1em}
\textbf{\$\$AGENT\_REQUEST\$\$}

\texttt{Method: FUNCTION} \\
\texttt{URL: perplexity\_api\_response}

\textbf{BODY}
\begin{lstlisting}[language=json]
{
  "query": "What are the latitude and longitude for the following places: 
            \"The Dolomites, Italy\", \"Santorini, Greece\", \"Prague, Czech Republic\", 
            \"Cinque Terre, Italy\", \"Barcelona, Spain\"",
  "preplexity_ai_key": "YOUR_API_KEY"
}
\end{lstlisting}
\vspace{1em}
\textbf{\$\$AGENT\_REQUEST\$\$}

\texttt{Method: FUNCTION} \\
\texttt{URL: perplexity\_api\_response}

\textbf{BODY}
\begin{lstlisting}[language=json]
{
  "query": "What are the latitude and longitude for the following places: 
            \"London, UK\", \"Rome, Italy\", \"Amsterdam, Netherlands\", 
            \"Munich, Germany\", \"Dublin, Ireland\"",
  "preplexity_ai_key": "YOUR_API_KEY"
}
\end{lstlisting}

\end{fancybox}

\begin{fancybox}[Example of \texttt{AGENT\_RESPONSE}]
This example shows how the agent formats and validates tool outputs, converting a response into a structured output variable, which can be parsed and stored in the Execution Blueprint, such that, the downstream steps can reliably consume it in case of inter-dependencies.\\

\textbf{\$\$CHAIN\_OF\_THOUGHT\$\$}

\begin{itemize}
  \item The API response provides a comprehensive list of vacation spots in the USA, categorized by different types of destinations such as national parks, beach destinations, and urban experiences. This aligns with the expected output variable \texttt{vacation\_spots\_list\_usa} for the current step.
  \item For the current step, the Output Variable \texttt{vacation\_spots\_list\_usa} requires a list of recommended vacation spots in the USA. The response provides a list of well-known spots, and these can be extracted and formatted correctly. We will keep the description of the vacation spots as well.
  \item The extracted vacation spots include a variety of destinations like Yellowstone National Park, Grand Canyon, Hawaii, and New York City, which are popular vacation spots in the USA. These spots can be structured as a list for the output variable.
  \item For the dependent step (Sub-Task 2, Step 1), the Input Variable \texttt{vacation\_spots\_list\_usa} requires a list of vacation spots to retrieve geographic coordinates for each location. The extracted list of vacation spots from the current step meets the requirement of the dependent input and is formatted as a string array, making it compatible for use in the subsequent API call.
  \item The data is structured appropriately for both the current step's output and the dependent step’s input. The list of vacation spots is formatted as an array of strings, ready to be passed as input to retrieve latitude and longitude.
\end{itemize}

\vspace{1em}

\textbf{\$\$AGENT\_RESPONSE\$\$}

\textbf{Status\_Code:} 200 OK

\vspace{0.5em}
\textbf{Output\_Variables}
\begin{itemize}
  \item \texttt{Variable Name: vacation\_spots\_list\_usa}
  \begin{itemize}
    \item \textbf{Content:}
    
    There are numerous excellent vacation spots across the United States catering to different interests and preferences. Here are some top destinations to consider:

    \textbf{National Parks and Natural Wonders}
    \begin{itemize}
      \item Yellowstone National Park: Famous for its geothermal features like Old Faithful geyser and diverse wildlife.
      \item Grand Canyon National Park: Offers breathtaking views of one of the world's most spectacular natural formations.
      \item Glacier National Park: Known for its pristine forests, alpine meadows, and rugged mountains.
      \item Zion National Park: Features stunning red rock formations and excellent hiking trails.
    \end{itemize}

    \textbf{Beach Destinations}
    \begin{itemize}
      \item Hawaii: The islands of Maui, Oahu, and Hawaii (Big Island) offer beautiful beaches, lush landscapes, and unique cultural experiences.
      \item Myrtle Beach, South Carolina: A popular family-friendly destination with miles of sandy beaches and numerous attractions.
    \end{itemize}

    \textbf{Cities and Urban Experiences}
    \begin{itemize}
      \item New York City: A global cultural hub with iconic landmarks, world-class museums, and diverse neighborhoods.
      \item New Orleans, Louisiana: Known for its unique culture, vibrant music scene, and delicious cuisine.
      \item San Francisco, California: Features the iconic Golden Gate Bridge, diverse neighborhoods, and nearby wine country.
    \end{itemize}
  \end{itemize}
\end{itemize}

\vspace{0.5em}
\textbf{Dependent\_Input\_Variables}
\begin{itemize}
  \item \texttt{Variable Name: vacation\_spots\_list\_usa}
  \begin{itemize}
    \item \textbf{Sub-Task:} 2
    \item \textbf{Step:} 1
    \item \textbf{Type:} string
    \item \textbf{Content:} \texttt{["San Francisco", "New York City", "New Orleans", "Myrtle Beach, South Carolina", "Hawaii", "Zion National Park", "Glacier National Park", "Grand Canyon National Park", "Yellowstone National Park"]}
  \end{itemize}
\end{itemize}

\end{fancybox}

\section{Multimodal Report Sample}
\label{appendix:real_report_sample}



\section*{Supplementary material}

# 1 Introduction to Electric Vehicles (EVs) and the Conditions Shaping Modern Transportation

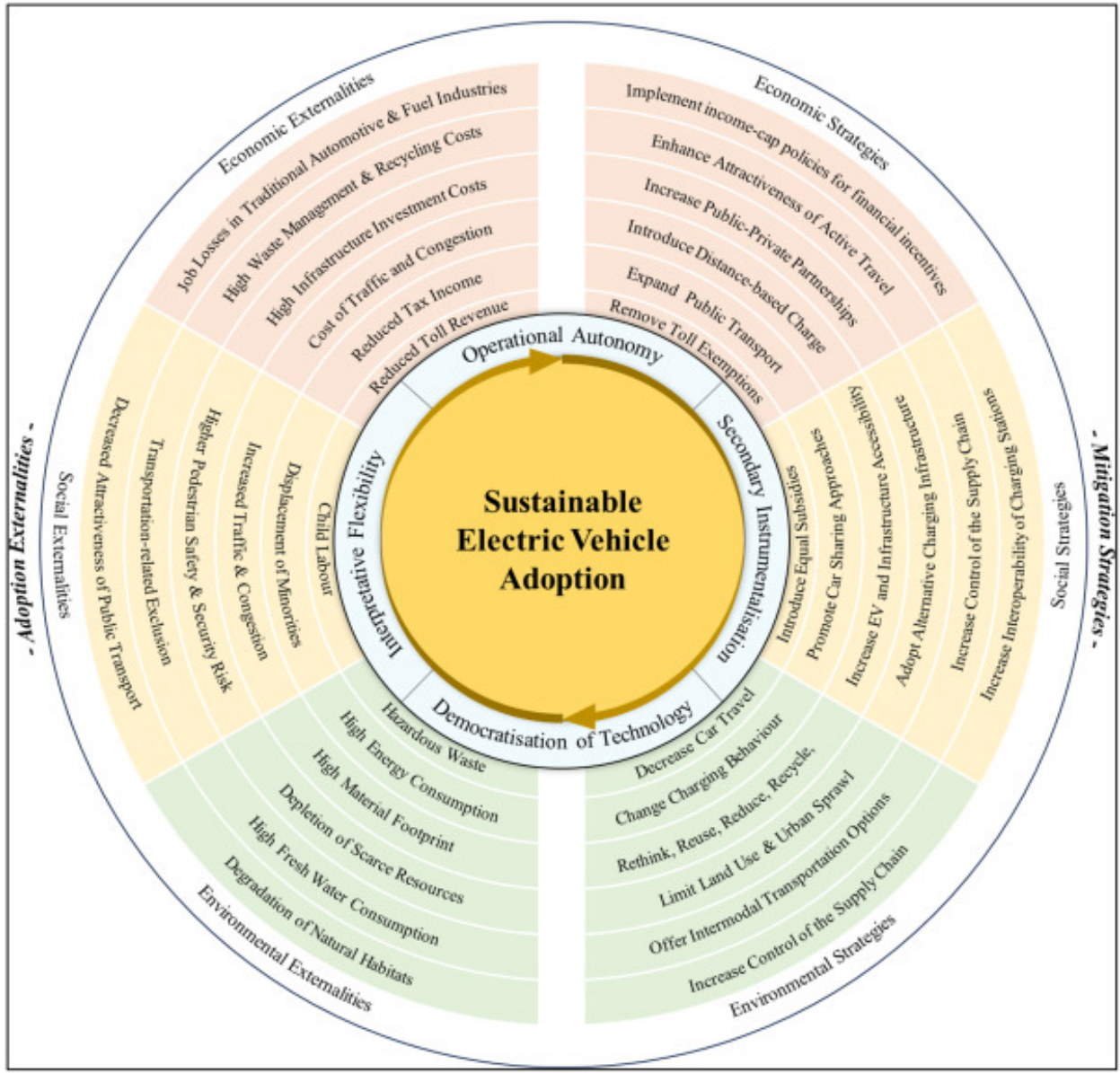

Figure 1: Electric Vehicles: A Key Innovator in Modern Mobility.

## 1.1 Emerging Relevance of Electric Vehicles (EVs)

Electric vehicles (EVs) have become a driving force in reshaping global transportation due to rising adoption rates, sustained technical innovations, and strong policy support. Their importance is evident in the following ways:

1. Market Growth and Consumer Adoption

- Recent data shows continuous market expansion, with EVs occupying an increasing share of new vehicle sales. This surge stems from a combination of declining battery costs, evolving consumer preferences for cleaner mobility, and the entrance of both established and specialized EV manufacturers.

- Many countries are setting ambitious targets and regulations that favor electric mobility, accelerating the transition from combustion engines to EVs in both passenger and commercial sectors.

2. Technological Breakthroughs

- **Batteries:** Developments such as lithium-iron-phosphate (LFP) batteries, high-capacity solid-state cells, and improved cathode chemistries are boosting energy density, lowering costs, and enhancing vehicle range. These improvements directly address earlier consumer concerns about limited driving distance and battery longevity.

- **Charging Infrastructure:** An expanding network of charging stations—now including ultra-fast DC chargers—reduces waiting times and mitigates range anxiety. Interoperability standards also streamline charging across different service providers, promoting confidence among new EV buyers.

3. Energy and Sustainability Factors

- **Emissions Reduction:** EVs run cleaner than traditional internal combustion engine vehicles, contributing to lower local air pollution and helping meet national and international climate targets.

- **Lifecycle Improvements:** Closed-loop battery recycling and carbon-neutral manufacturing initiatives aim to make electric vehicle production and disposal more sustainable over the long term.

4. Economic and Industrial Shifts

- **Affordability:** The total cost of ownership for EVs is becoming increasingly competitive due to reduced fuel and maintenance expenses. Many governments also support this growth through financial incentives, such as tax rebates or direct subsidies.

- **Job Creation and Innovation:** Rapidly scaling production facilities (including gigafactories) and advanced manufacturing methods (e.g., digital twin simulations, automation) are sparking job growth and fostering regional development in emerging EV-centered supply chains.

5. Evolving Consumer Perception

- **Mainstream Appeal:** Whereas EVs were once seen as niche options, a broader range of models now serves varied lifestyles, from compact urban vehicles to long-range SUVs.

- **Charging Confidence:** Fast charging capabilities and a more visible station network resolve earlier consumer doubts about refueling convenience, standing as a testament to how quickly the infrastructure has caught up with market demands.

6. Future Outlook

- As more automakers invest heavily in electric lineups and work with technology partners to refine batteries and charging systems, EVs will become even more affordable and practical.

- The continued interplay of regulatory measures, consumer interest, and technological advancement suggests that electric vehicles will play an increasingly central role in transportation strategies worldwide.

## 1.2 Conditions Shaping Modern Transportation

Below is a concise yet comprehensive standalone subsection exploring the environmental, economic, and social drivers that are accelerating the shift to electric vehicles (EVs) in modern transportation. It references recent findings up to March 2025.

1. Environmental Drivers **a. Emissions Reduction and Policy Support**

- EVs play a central role in reducing greenhouse gas and smog-forming emissions, which is vital for meeting both near- and long-term climate targets. Traditional vehicles account for a substantial share of transportation-related emissions; by contrast, electrification helps decarbonize mobility.

- Many governments maintain or strengthen incentives to stimulate EV adoption, including tax credits, tougher emissions standards, and zero-emission mandates. Studies suggest that if these supports are marginalized—such as by repealing EV tax credits—electric vehicle sales could drop significantly, undermining climate goals.

- Policy discontinuities across jurisdictions can generate uncertainties. Some regions have paused or weakened earlier mandates, while others are ramping up targets. Consistent policy frameworks remain key for long-term emissions mitigation.

**b. Renewable Energy Integration**

- To optimize environmental benefits, EVs are ideally charged from cleaner grids. Increasing the share of renewables—especially cost-competitive solar and wind—amplifies emissions reductions.

- Integrating large-scale renewables within power grids presents variability challenges. Utilities are therefore investing in energy storage and grid modernization to accommodate both the growing electrification of transportation and fluctuating renewable energy resources.

- Better coordination of renewable deployment with EV infrastructure (e.g., strategically placed charging stations and advanced load management) will further solidify the environmental case for widespread electric mobility.

2. Economic Drivers **a. Incentives and Cost Trends**

- Upfront cost remains a significant barrier for many consumers. Rebates, tax credits, and other financial incentives can bridge the price gap, particularly in the entry and mid-level EV segments. Notable examples include thousands of dollars in bonus cash on select plug-in hybrids and fully electric models.

- Concurrently, declining battery costs are driving down overall EV prices. Recent data indicates a modest dip in average EV transaction prices, reflecting scaled manufacturing. This cost trend enhances affordability and broadens EV accessibility.

**b. Battery Technology and Supply Chain Policies**

- Advances in battery technology, such as higher-density cells and the prospect of solid-state batteries, reduce costs and improve range, both of which bolster mainstream market appeal.

- Policymakers increasingly emphasize securing domestic supply chains for key battery materials and incentivizing local EV manufacturing. Initiatives aim to align industrial policies with EV expansion, potentially creating new jobs, stabilizing the battery supply, and reinforcing national competitiveness.

3. Social Drivers **a. Consumer Awareness and Education**

- Growing environmental awareness and the desire for clean technology have led to heightened consumer interest in EVs. Surveys reveal that even premium car buyers are shifting to EVs due to environmental convictions, operational cost benefits, and the appeal of next-generation tech features.

- Awareness campaigns organized by automakers, public agencies, and non-profit groups dispel longstanding misconceptions such as limited range or prohibitive costs. Hands-on experiences—through test drives or showcases—have proven to be especially compelling, providing tangible evidence of EV convenience and viability.

**b. Shifting Attitudes Toward Sustainability**

- Public sentiment increasingly favors sustainable lifestyles, placing EVs at the forefront of a larger cultural shift toward greener choices. University studies show that direct engagement with EVs correlates with stronger intentions to purchase or lease these vehicles.

- Localized outreach efforts—like dealerships trained in EV sales, ride-and-drive events, and coordinated campaigns on social media—further bolster public acceptance. As familiarity grows, skepticism typically recedes, helping EVs gain traction in mainstream markets.

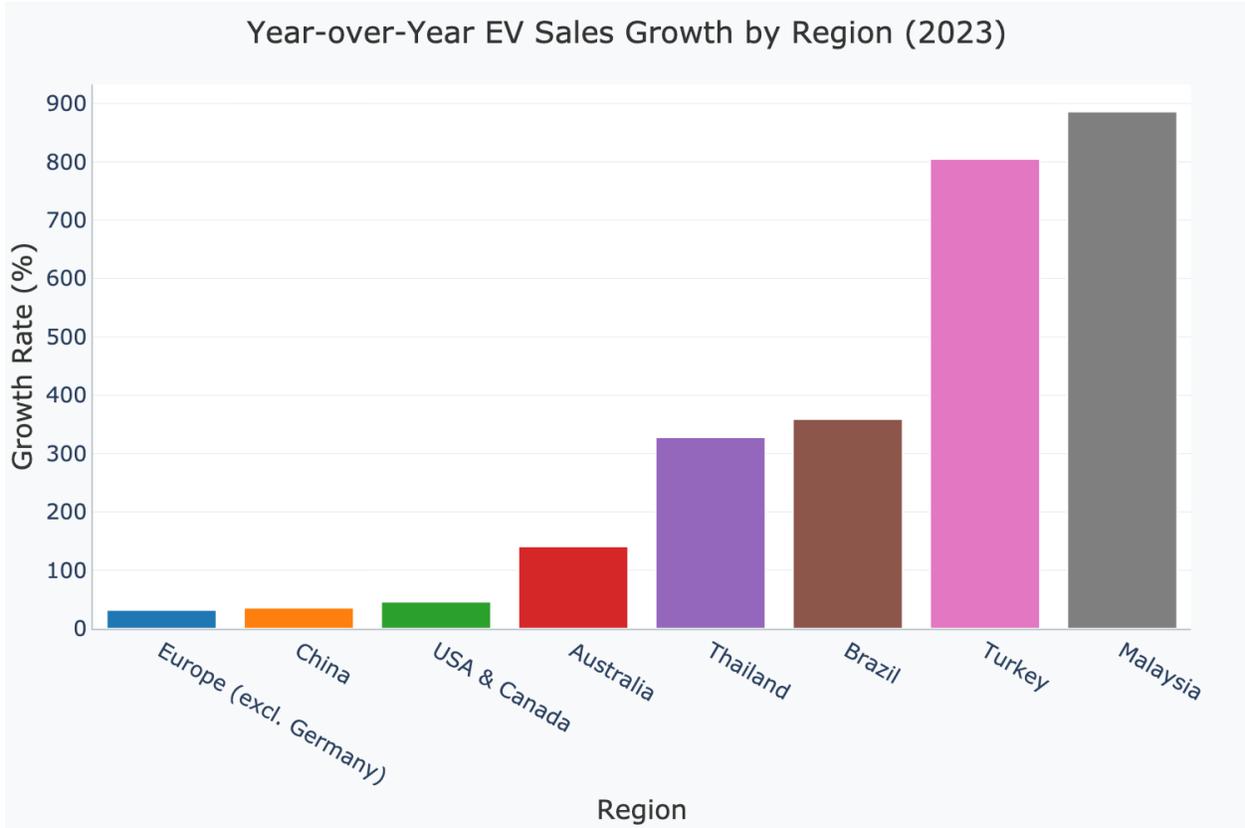

Figure 2: *Year-over-Year EV Sales Growth by Region (2023).*

As shown in Figure 2, there are significant variations in EV adoption across different markets, with some regions demonstrating exceptionally high year-over-year growth. These surging adoption rates underscore the discussions above regarding global market expansion, consumer interest, and the worldwide shift toward cleaner mobility.

# 2 Foundations of EV Technology

As electric vehicles (EVs) mature, understanding the underlying propulsion technologies provides essential context for analyzing adoption patterns and infrastructure demands. See Figure 3 for an illustration of how modern EV powertrains are reshaping automotive design.

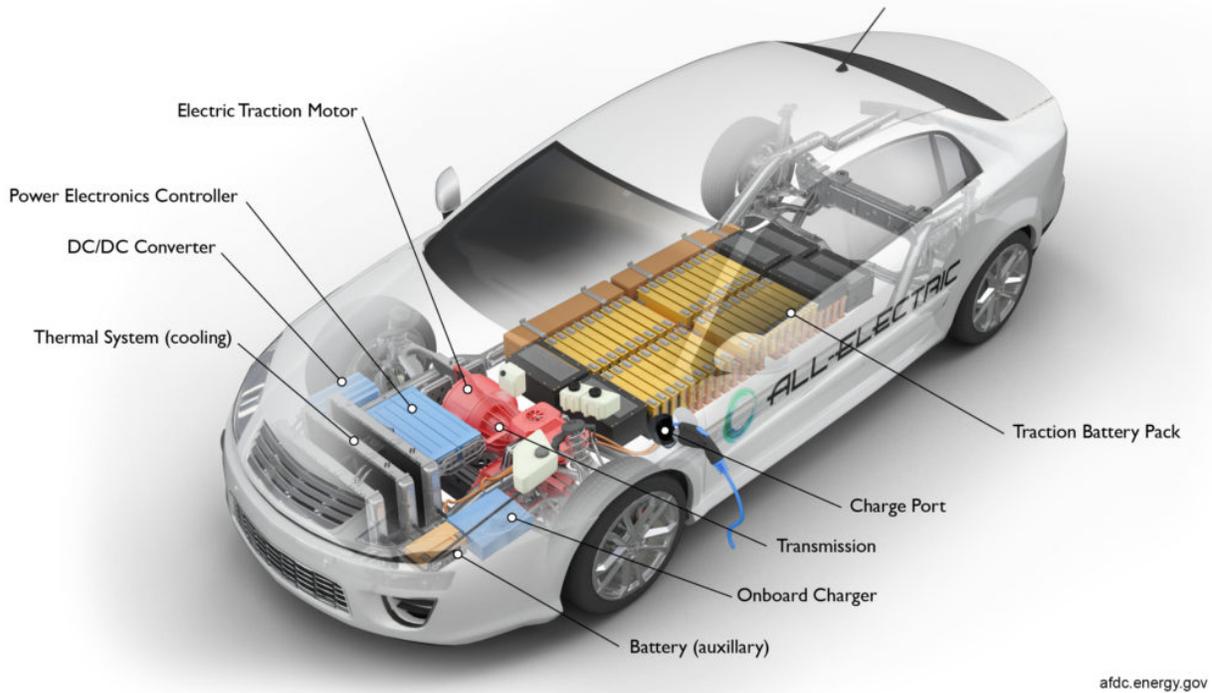

*Figure 3: A key visual representing the emergence of electric powertrains.*

## 2.1 Battery and Propulsion Basics

**1. Introduction**

Electric vehicles (EVs) and internal combustion engine (ICE) vehicles fundamentally differ in how they generate mechanical power. EVs use electric motors powered by rechargeable batteries, whereas ICE vehicles burn fuels (like gasoline, diesel, or hydrogen). This contrast underpins the broader transition toward cleaner mobility.

**2. Energy Source**

- **EVs:** Powered by battery packs—most commonly lithium-ion or solid-state cells—that are recharged from the electricity grid or renewable sources.

- **ICEs:** Depend on fossil fuels or hydrogen (in hydrogen combustion engines) for energy. Though hydrogen can reduce carbon emissions, generating NOx remains possible if not properly managed.

**3. System Efficiency**

- **EVs:** Achieve over 90% efficiency in converting stored electrical energy into motion. Regenerative braking also recovers energy. If the electricity is low-carbon or renewable, total emissions are minimized.

- **ICEs:** Inherently less efficient due to energy lost as heat. Contemporary designs, such as turbocharging or advanced combustion methods, help marginally raise efficiency, but still lag behind electric drivetrains.

### 4. Environmental and Emission Profile

- **EVs:** Zero tailpipe emissions. Their total carbon footprint depends largely on upstream electricity generation. Expanding renewable energy reduces emissions further.

- **ICEs:** Produce greenhouse gases (especially $CO_2$) and other pollutants like NOx and particulate matter. Hydrogen ICEs cut $CO_2$ substantially but still require special controls for NOx.

### 5. Maintenance Requirements

- **EVs:** Minimal mechanical complexity leads to fewer service intervals. Regenerative braking reduces brake wear. No spark plugs or oil changes typically lowers ongoing costs.

- **ICEs:** More intricate mechanical components. Regular fluid changes and upkeep of emission-control systems are necessary, often implying higher long-term maintenance.

### 6. Performance Characteristics

- **EVs:** Deliver immediate torque and brisk acceleration. Technological developments continue to extend driving range.

- **ICEs:** Power delivery is often strongest at higher RPMs, with refinement possible through forced induction. Generally do not match an electric motor's instant responsiveness.

### 7. Implications for Transportation and Future Outlook

- **Consumer Experience:** EVs are notably quieter with smoother torque delivery, appealing for both urban commutes and performance enthusiasts.

- **Cost of Ownership:** While initial EV prices may be higher, savings in fuel and maintenance often offset this over the vehicle's life.

- **Infrastructure Challenges:** Charging station availability is critical for widespread adoption. For hydrogen ICEs to see broader use, hydrogen fueling infrastructure must likewise expand.

- **Policy and Regulation:** Global emissions goals drive continued efforts to enhance EV viability, though combustion engine technologies still evolve in specialized applications.

*Table 1: Electric Vehicles (EVs) vs. Internal Combustion Engine (ICE) Vehicles: A Comprehensive Overview*

| Aspect | Description | EVs | ICEs |
|---|---|---|---|
| Market Size | Overall global demand or sales volume | Over 10 million units sold in 2022; forecast 14 million for 2023 (IEA) | 181.8 million units in 2022; projected 9.2% CAGR to 2030 (Grand View Research) |
| Market Growth Drivers | Main factors propelling adoption and expansion | Policy support, technological improvements, infrastructure investments | Demand for passenger and commercial vehicles in developed and emerging markets |
| Energy Source | Primary fuel or power mechanism | Battery packs (lithium-ion/solid-state), charged via grid/renewables | Fossil fuels (gasoline, diesel) or hydrogen (H2-ICE) |
| System Efficiency | How effectively stored energy is converted into vehicle motion | Electric motors exceed 90% efficiency; regenerative braking recaptures energy | Less efficient due to heat loss in combustion; turbocharging helps but still trails EVs |
| Environmental & Emission Profile | Greenhouse gas emissions and broader ecological impact | Zero tailpipe emissions; footprint depends on electricity source | Emit $CO_2$, NOx, particulates; hydrogen ICEs reduce carbon but still produce NOx |
| Maintenance Requirements | Frequency and complexity of upkeep | Fewer moving parts, no engine oil, typically lower costs | Requires regular fluid changes, emission control repairs, more mechanical subsystems |
| Performance | Acceleration, torque delivery, and general drivability | Instant torque, quiet operation, typically swift acceleration | Higher RPM to reach peak power; can be powerful but less immediate response |
| Fuel Economy | Typical fuel consumption or efficiency metrics | Often measured in MPGe; very efficient when charged via clean energy | Real-world usage 27.1 mpg in the U.S. (EPA), less efficient overall |
| Infrastructure | Support systems (charging/fueling) | Widespread charging network still developing; DC fast-charging crucial | Well-established global network of gas stations; limited hydrogen fueling sites |
| Alternative Fuels | Non-traditional fuels for powering vehicles | Increasing reliance on renewables (wind/solar) for charging | CNG and hydrogen (H2-ICE); remain less common than gasoline/diesel |
| Technological Advancements | R&D improvements improving usability | Battery enhancements (solid-state), refined motor designs, V2G potential | Familiar, but partial electrification, advanced combustion to reduce emissions |
| $CO_2$ Emissions | Carbon dioxide released per mile | Very low/zero at tailpipe if clean power is used | 319 g/mi average; though some incremental efficiency gains persist |
| Maintenance Costs | Typical ongoing expenses | $0.0812 per mile (AAA); minimal mechanical wear | $0.10 per mile (EPRI); more moving parts/fluid changes |
| Repair Costs | Complexity and cost for major repairs | High due to specialized parts, though fewer components overall | Generally cheaper to fix thanks to established service networks |
| Life Cycle Assessment (LCA) | Resource use across production, operation, and end-of-life | Dependent on battery sourcing and recycling; growing body of research | Maintenance phase, fuel extraction, and emissions weigh into total impact |
| Environmental Impact | Beyond greenhouse gases, e.g. air quality and resource usage | Reduced local air pollution; reliant on sustainable electricity generation | Major contributor to NOx and PM, resource-intensive fossil fuel supply chain |
| Government Regulations | Policy frameworks, mandates, or incentives | Subsidies, ZEV mandates, and tax credits accelerate EV growth | Facing stricter $CO_2$ thresholds, leading to partial hybridization |
| Future Outlook | Forecasts for next decade | Rapid sales growth, cost declines, extensive charging expansions | Retains a role in certain applications despite EV rise, but future is more limited |
| Key Growth Challenges | Potential barriers to expansion | Battery material supply, charging infrastructure, initial vehicle cost | Tightening emission standards, high fuel prices, rising EV competition |

## 2.2 Key Milestones in EV Evolution

**Early Lithium Batteries and the Shift to Higher Density** Lithium-ion battery technology, emerging strongly in the late 2000s, profoundly elevated EV range and performance. The first production vehicles to adopt these batteries demonstrated that EVs could deliver excellent acceleration and travel distances of over 200 miles, paving the way for further breakthroughs.

**Advancements in Battery Chemistry and Solid-State Development** Steady work on improving cathode materials (nickel-cobalt-aluminum, nickel-cobalt-manganese) has en-

hanced energy density and lowered costs. Emerging solid-state batteries promise even greater range and durability, using solid electrolytes that may allow faster charging and smaller battery pack footprints.

**Electric Motor Efficiency and Performance** Permanent magnet synchronous motors, advanced power electronics, and sophisticated cooling systems have boosted drivetrain efficiency and reliability. These improvements amplify EVs' hallmark of instantaneous torque—a feature widely appreciated by consumers transitioning from ICE vehicles.

**Charging Infrastructure Growth and Interoperability** To mitigate range anxiety, many governments and private companies have coordinated to build more public fast-charging stations. Interoperability initiatives, such as Tesla opening parts of its Supercharger network, represent a move toward unified standards that simplify the user experience.

**Renewable Integration and Vehicle-to-Grid (V2G)** Greater reliance on renewable energy sources, particularly wind and solar, further improves the carbon balance of EV operations. Vehicle-to-Grid (V2G) capabilities enable bidirectional power flows, allowing EVs to assist grid stabilization and potentially provide cost-offsetting revenues to owners.

**Leading Companies and Influential Players** From Tesla's global brand recognition to BYD's vertically integrated battery-vehicle manufacturing, major corporations are driving technological advances. Established automakers like Toyota, Ford, and Volkswagen are also ramping up EV investments, contributing to a diverse, rapidly expanding market.

**Market Accessibility and Outlook** Continued reductions in battery production costs and expanded charging infrastructure have placed many EVs within reach of mainstream buyers. While regional incentives and policy frameworks remain influential, industry projections show a sustained upward trajectory for electric mobility across the globe.

The bar chart in Figure 4 underscores how EV drivetrains outpace ICEs in energy conversion. Achieving 87–91% propulsion efficiency, EVs greatly outperform the roughly 20% common to ICE vehicles. These gains in efficiency are central to EVs' lower environmental footprint.

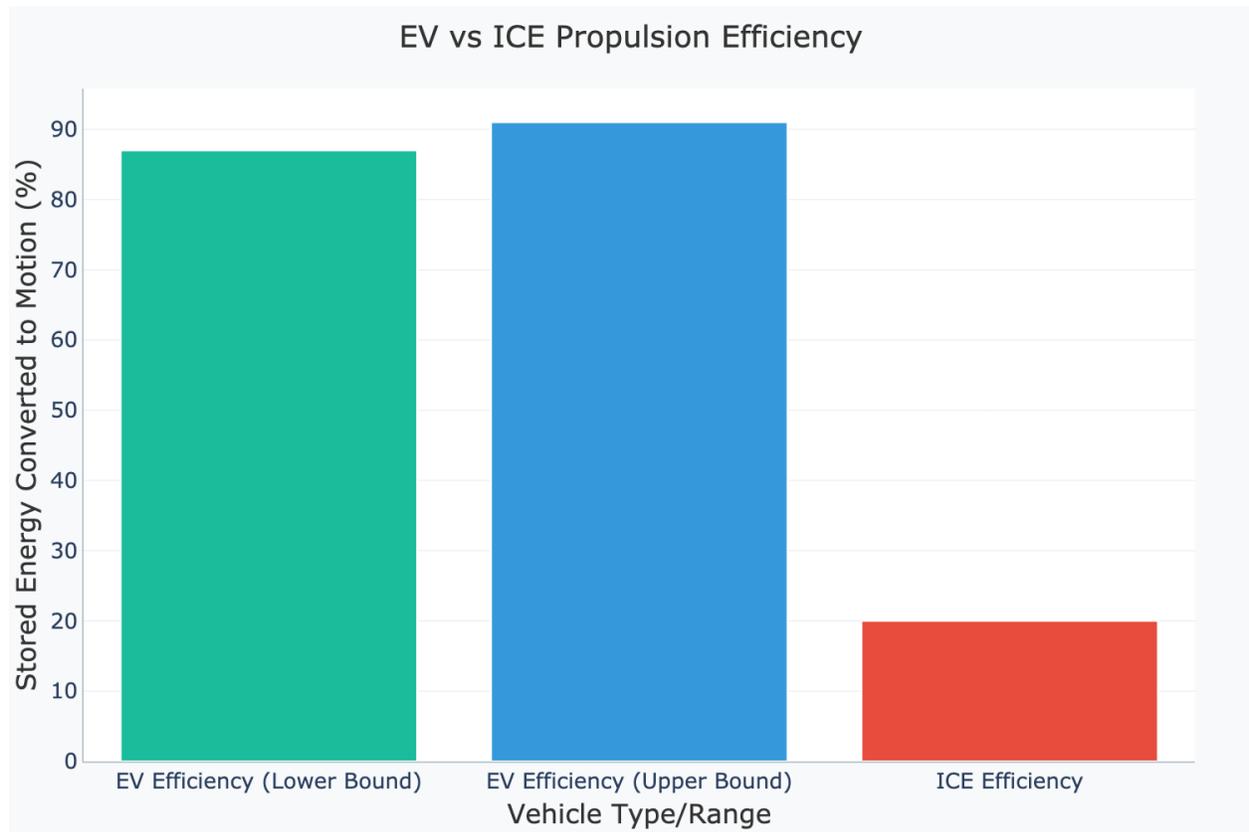

Figure 4: Comparison of propulsion efficiency in EVs and ICE vehicles.

# 3 Overview of Electric Vehicle Adoption

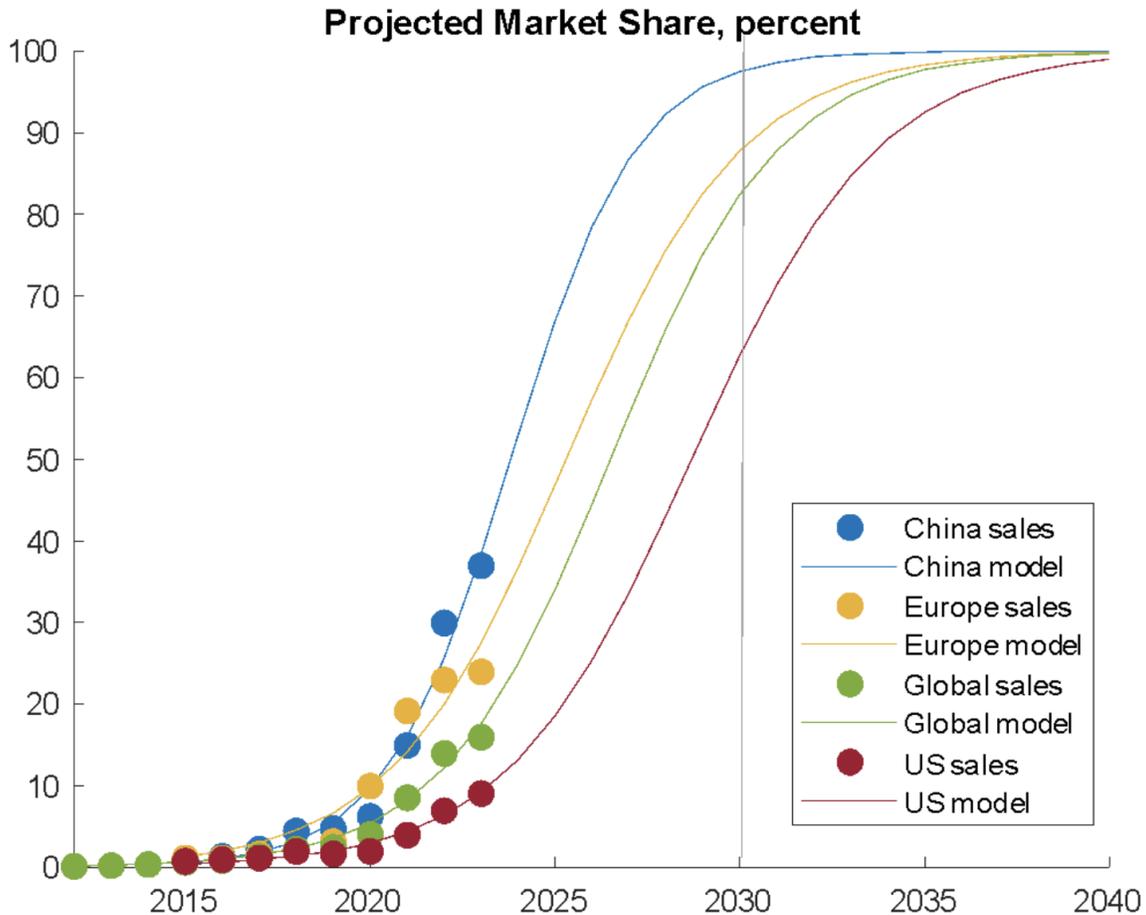

Figure 5: Depiction of global interest in electric vehicle adoption.

## 3.1 Global Adoption Rates

Recent data highlights the worldwide increase in EV market share, indicating a sustained transition away from internal combustion engines:

**Recent Worldwide Market Penetration and Sales Figures for Electric Vehicles**

*Introduction to Global EV Sales and Penetration* Electric vehicles (EVs) continue to expand their presence in the global automotive market. In 2023, EVs accounted for roughly 18% of all new car sales—equivalent to approximately 14 million units globally. This marked a substantial rise over the previous year and demonstrated how policy support, declining battery costs, and widening consumer acceptance have enabled a steady shift from internal combustion engines.

*Regional Variations and Concentration*

- **China:** Over one in three new car registrations in 2023 were electric, driven by strong government incentives and a robust local manufacturing base.

- **Europe:** EVs formed around 20% of new car registrations, aided by emissions regulations and purchase subsidies.

- **United States:** Though smaller in percentage (around 10%), the U.S. still ranks among the top three global EV markets, fueled by expanding tax credits and increased automaker commitments.

- **Other Regions:** Japan, India, and others show modest but gradually rising EV shares, largely dependent on infrastructure investments and evolving incentive programs.

*Rising Global Stock and Vehicle Types* Globally, there are over 40 million electric cars on the road. Around 70% are battery electric vehicles (BEVs), while the rest are predominantly plug-in hybrid electric vehicles (PHEVs). As technologies advance in range, performance, and affordability, the inclination toward fully electric drivetrains continues to grow.

*Growth Trajectory Through 2025* Preliminary data suggests continued momentum, particularly in China and Europe where policy targets and battery technology advances are poised to shape future gains. Other regions turning attention to charging infrastructure investments and local production may experience stronger EV market arrivals in the coming years.

*Infrastructure Considerations and Outlook* The growth in EV usage accentuates the need for robust charging infrastructure and supply chain management. Leading markets have gathered insights on grid-readiness and high-power charging networks that can inform emerging markets as they build their own EV ecosystems.

*Concluding Remarks* Overall, EVs are making notable strides worldwide, especially in regions that emphasize policy action and industrial support. While certain markets still lag, expanding model availability and progress in charging technologies are likely to accelerate uptake, supporting broader environmental and economic goals in the transportation sector.

## 3.2 Regional Market Snapshots (U.S., China, Europe)

Differences in policy focus, consumer behavior, and industrial strategies shape how these three markets lead the EV revolution.

**Policy Support**

- *United States:* Central initiatives like the Clean Vehicle Credit, ATVM Loan Program, and incentives for charging infrastructure have helped spur EV penetration, though adoption rates vary widely by state.

- *China:* National and local governments prioritize EVs through tax incentives, subsidy programs, and extensive charging network investments, driving the highest sales volume worldwide.

- *Europe:* Ambitious climate targets, emissions standards, and national-level subsidies or tax breaks create an environment where EV sales can rapidly rise in many member countries.

**Consumer Demand**

- *United States:* Around 9% of the 2023 market comprised EV sales. Consumer acceptance is buoyed by a broader range of EV models, improved highway charging, and policy incentives.

- *China:* Leading global penetration at nearly 40% of new car sales in 2023, driven by heightened environmental awareness and comprehensive government support.

- *Europe:* With EVs comprising around 23.6% of new car registrations in 2023, demand is propelled by strict $CO_2$ standards, more expansive charging networks, and various national-level purchase incentives.

**Industry Efforts**

- *United States:* Automotive firms, tech companies, and energy providers collaborate to advance EV charging and battery technologies.

- *China:* Homegrown battery and EV manufacturing capabilities, combined with government-led industry policies, spur rapid innovation and substantial production scale.

- *Europe:* A mix of joint ventures and legislative frameworks encourage regional battery production, integrated mobility solutions, and robust R&D collaboration.

*Future Outlook* Each region charts a distinct path: the U.S. focuses on infrastructural expansion, China fortifies its global leadership in battery production, and Europe relies on stringent regulations and green investment to promote EVs. Despite unique challenges, these three markets collectively steer global EV development, establishing paths and benchmarks for the rest of the world.

## 3.3 Indicators of Growing Demand

Market data points to multiple signs of quickening consumer and fleet shifts toward EVs. Major indicators include fleet-share gains, significant year-on-year sales growth, and evolving consumer sentiment.

**Fleet-Share Shifts**

- *Global:* Around 18% of new car sales were EVs by the end of 2023.

- *United States:* Hybrid, plug-in hybrid, and battery electric vehicles combined accounted for roughly 18.7% of new light-duty sales by mid-2024.

- *China:* Nearly 40% of new car sales in 2023 were EVs.
- *Europe:* Electric registrations topped 23% of total new vehicle signups in 2023.

**Year-on-Year Sales Growth**

- *Global:* EV sales expanded by approximately 35% in 2023.
- *United States:* Achieved a 46% increase.
- *China:* Registered a 36% jump.
- *Europe:* Noted a 17% uptick.

**Changes in Consumer Sentiment**

- *Key Motivations:* Consumers cite lower fuel costs and reduced environmental impact as strong incentives for going electric.
- *Persistent Barriers:* Worries remain about high upfront costs, insufficient charging coverage, battery repair expenses, and range limitations.
- *Demographic Trends:* Younger, urban, and sustainability-minded buyers are more inclined to adopt EVs.

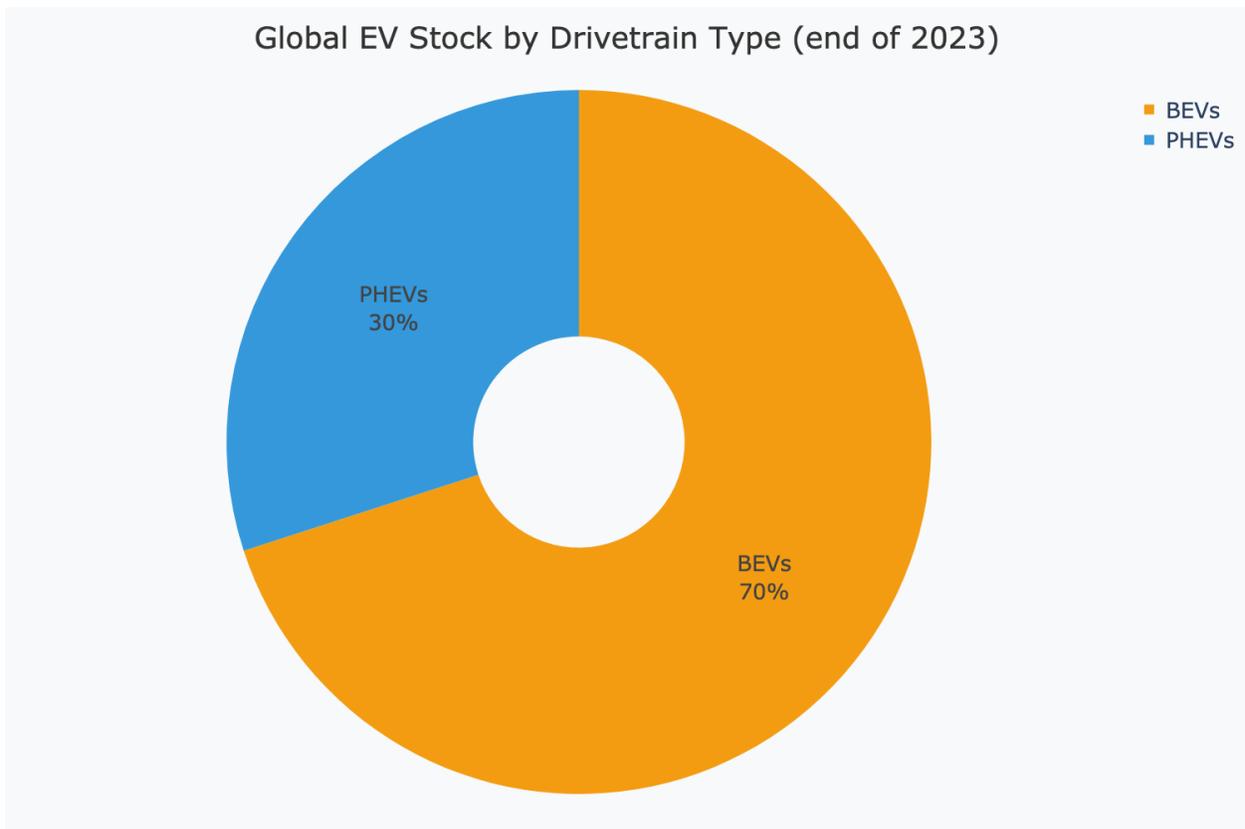

*Figure 6: 2023 EV unit sales (in millions) by region.*

16

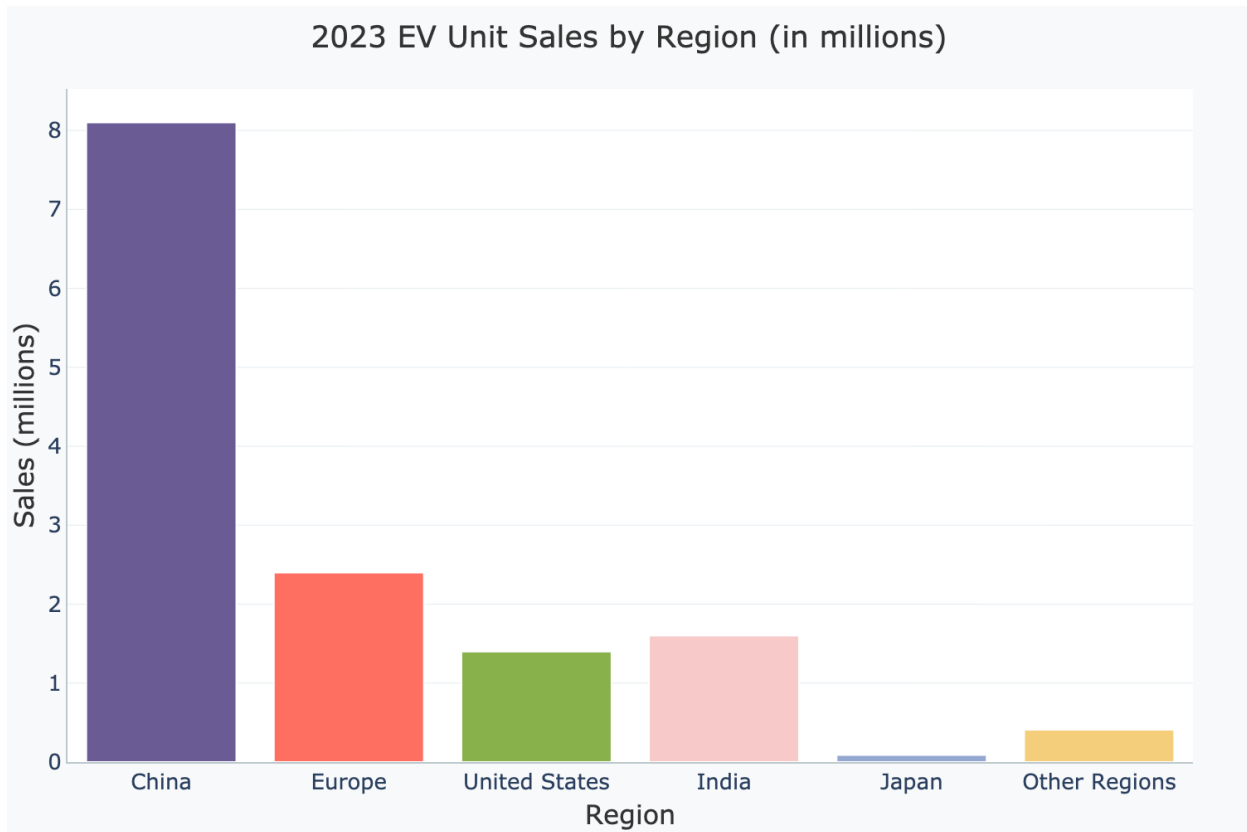

*Figure 7: Global EV stock in 2023: BEVs vs. PHEVs.*

Combining fleet-share increases, brisk sales growth, and shifting consumer sentiment underscores the rapidly rising global interest in EVs. As model diversity broadens and infrastructure improvements continue, these signals point to an enduring momentum for electrified transportation.

# 4 Policy and Economic Drivers of EV Adoption

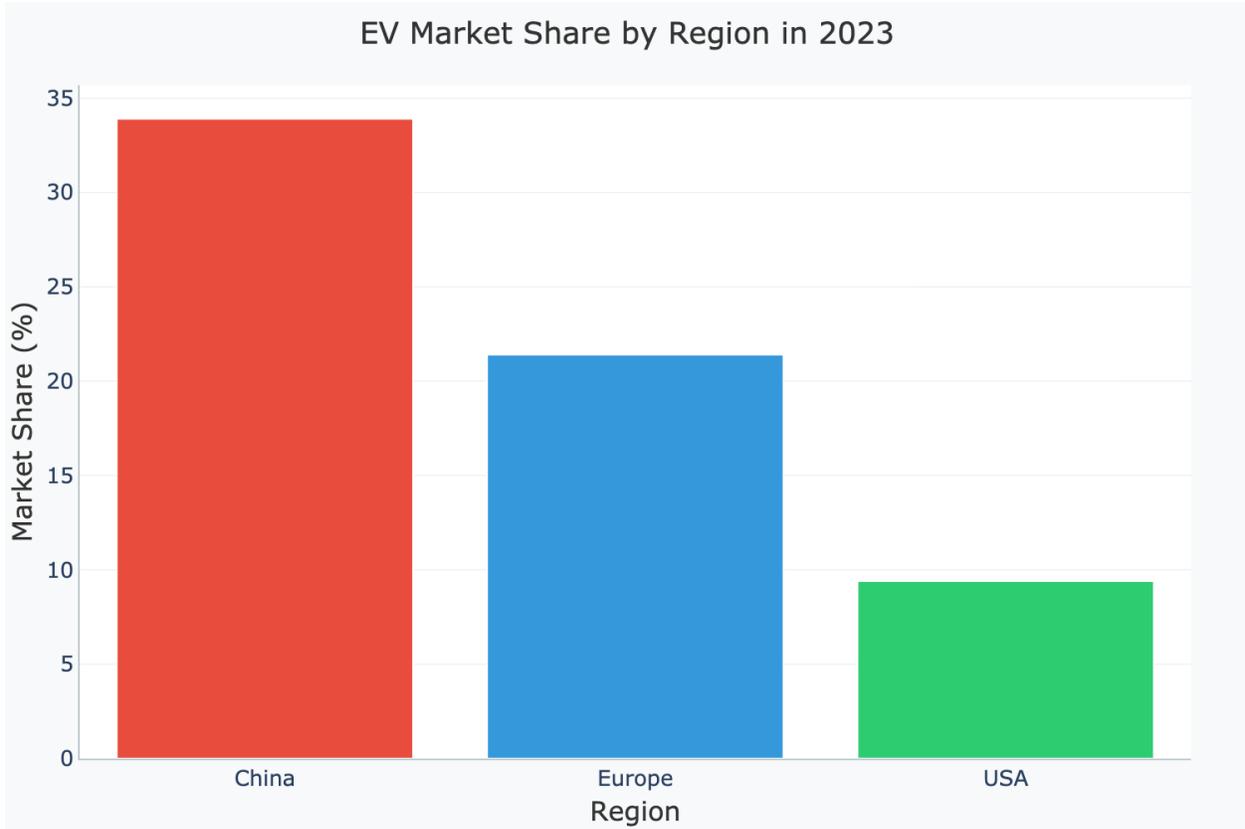

Figure 8: Comparison of 2023 electric vehicle market share across key regions (percent of new vehicle sales).

## 4.1 Government Incentives and Regulations

Tax breaks, rebates, and formal mandates serve as powerful levers for governments to propel the adoption of electric vehicles (EVs). By lowering upfront costs and reducing long-term financial risks, these policies can significantly accelerate consumer uptake and guide automakers' production strategies.

In the United States, a federal tax credit of up to US$7,500 historically helped reduce the purchase price gap between EVs and conventional vehicles. As of early 2025, proposed federal actions seek to revoke these credits under broader tax reforms. Many experts caution that removing them could temper EV sales growth and reduce progress toward nationwide emissions targets. Moreover, executive initiatives have focused on rolling back EV incentives and easing regulations tied to gasoline-powered cars, generating uncertainty for both manufacturers and buyers. State-level incentives remain robust in some cases, including separate rebates, sales tax exemptions, and HOV lane privileges, which partially counterbalance any federal reductions but do not uniformly fill the gap.

Meanwhile, China applies a dual-credit policy that mandates a certain share of low-emission or zero-emission vehicles from automakers. Non-compliant manufacturers must purchase credits from those exceeding the targets, effectively rewarding higher EV output. Although monetary subsidies for battery electric vehicles (BEVs) have been scaled back, they continue to offset the "green premium," especially in big cities with stringent pollution control measures. Significant research and development funding also aims to reduce costs, enhance battery performance, and propel mass-scale manufacturing, firmly placing China at the forefront of global EV sales and technology.

In Europe, bloc-wide incentives coordinated by the European Commission involve purchase subsidies and targeted tax breaks that boost EV demand. This approach is designed to revitalize a lagging car market and curb emissions. Many EU countries add further supports such as registration tax breaks, point-of-sale rebates, and reduced annual vehicle taxes. Recognizing the strategic importance of battery manufacturing, the EU also provides research grants and favorable financing terms to stimulate local production. Standardized regulations and collaborative supply chain measures reinforce Europe's competitiveness in the EV arena.

Comparative observations reveal that the United States faces policy uncertainties, primarily because of potential federal rollbacks that risk undercutting past momentum. Nevertheless, localized initiatives at the state level stand strong. China's cohesive mix of stringent mandates, subsidies, and heavy technological investments maintains its upward trajectory in EV adoption. Europe's coordinated incentives and robust attention to battery manufacturing create a holistic support system for EV expansion.

Incentivizing EVs through tax breaks, rebates, and mandates crucially influences market momentum. Where policies remain proactive and consistent, as in China and Europe, EV adoption thrives. By contrast, looming debates over incentive phase-outs in the United States underscore how abrupt policy reversals can introduce uncertainty, discouraging consumer and industry investment. Moving forward, the most successful strategies seem poised to combine financial support with broader infrastructure commitments, fostering a resilient EV market that can endure policy shifts and economic cycles.

## 4.2 Financial and Macroeconomic Factors

Electric vehicles often present a favorable long-term cost proposition compared with internal combustion engine (ICE) counterparts, although upfront prices for certain models may be higher. Several elements help stack the economics in favor of EVs:

**Maintenance Efficiencies.** EVs generally have fewer moving parts, leading to reduced mechanical failures and less frequent servicing. Owners save on costs such as oil changes, resulting in up to 31% lower maintenance expenditures than traditional vehicles.

**Charging vs. Fuel Costs.** Electricity as a fuel source typically costs less per mile than gasoline or diesel. Annual charging expenses for many EV owners amount to hundreds of dollars, whereas comparable ICE vehicles often require thousands per year in fuel.

**Government Incentives.** Various regions provide tax credits, rebates, or reduced reg-

istration fees that narrow the initial cost gap. Some also offer perks like HOV lane access and reduced tolls, enhancing daily cost savings.

**Residual Value Trends.** While EV insurance can be slightly higher due to specialized components, premium EV models often show strong resale value. Over approximately 15 years, cumulatively lower maintenance and fuel costs, coupled with promising resale prices, can generate net savings in the range of US$7,000 to US$11,000 compared to ICE equivalents.

Fuel-price dynamics highlight that electricity rates, though subject to moderate inflation and influenced by external factors like natural gas prices, tend to exhibit more stability over time. By contrast, gasoline and diesel costs remain vulnerable to spikes arising from oil supply constraints or geopolitical events. This relative stability in electricity pricing serves as a compelling advantage for EV adoption, especially when conventional fuel costs surge abruptly.

Broader economic considerations, such as job creation and sectoral shifts, reveal that emerging EV manufacturing segments (battery assembly, software development, charging infrastructure deployment) can bolster overall employment. Demand for electricity will grow, prompting necessary grid upgrades, additional battery storage solutions, and expanded renewable power capacities—developments that can offer a wide range of new economic opportunities. Crucially, the zero tailpipe emissions associated with EVs also deliver societal benefits, improving air quality, cutting healthcare costs, and advancing decarbonization targets. Over the long term, supportive policy frameworks and technology investments can reduce dependence on oil markets and guide a future that emphasizes innovative, low-emission mobility solutions.

## 4.3 Consumer Behavior and Market Demand

Consumer behavior toward electric vehicles is influenced by multiple interrelated factors, including reputational influences, range anxiety, and rapidly shifting attitudes about sustainability.

**Reputational Influences.** Numerous buyers view EVs as status symbols, associating them with advanced technology, environmental consciousness, and forward-thinking lifestyles. This dimension of prestige can be harnessed by manufacturers who promote cutting-edge design and branding strategies that spotlight innovation and exclusivity.

**Range Anxiety.** Many would-be EV owners remain apprehensive about battery range, fearing the inconvenience of running out of power. Addressing these concerns involves expanding fast-charging infrastructures, clarifying real-world range capabilities, and steadily advancing battery technologies that shrink the perceived gap in convenience vs. conventional vehicles.

**Evolving Views on Sustainability.** Growing recognition of climate change's implications and the imperative to reduce carbon footprints encourage more individuals to consider EVs. Hands-on experiences, outreach campaigns explaining the full environmental impact (from battery production to recycling), and supportive word-of-mouth from early adopters

can accelerate broader acceptance.

As these factors converge, strategies designed to integrate reputational appeal, mitigate range worries, and emphasize sustainability frequently produce the most substantive gains in adoption. Wider EV uptake involves collaboration among policymakers, automakers, and infrastructure planners, ensuring that new buyers see both the practical and societal merits of electrified transport.

# 5 Addressing the Infrastructure Requirements and Progress for EVs

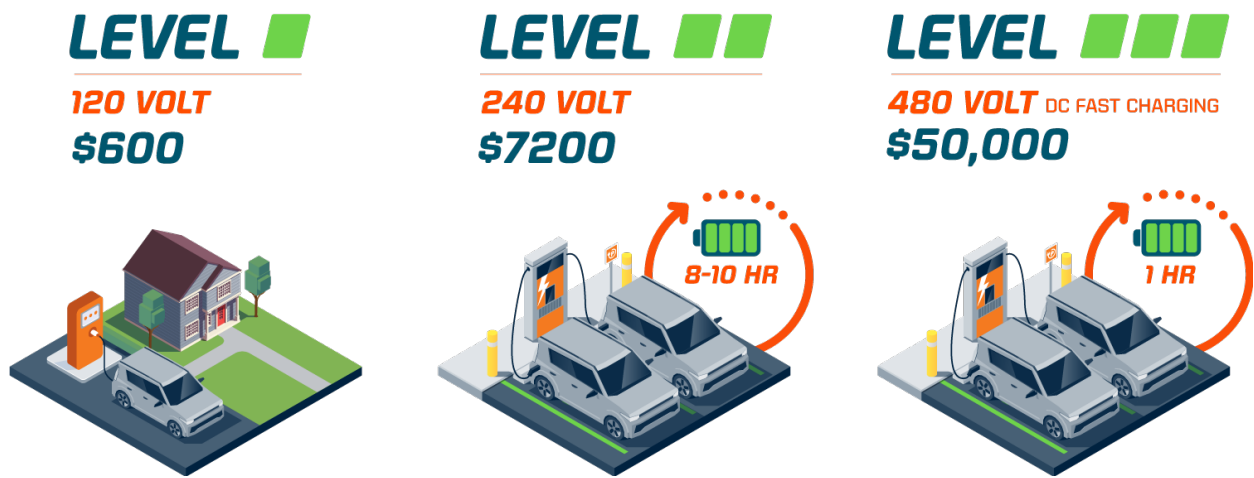

Figure 9: An overview of EV charging infrastructure developments.

## 5.1 Charging Solutions: Public and Residential

Electric vehicle (EV) charging infrastructure spans from basic, low-power chargers used in residences to high-power stations that facilitate rapid charging along major highways. Several charger levels have emerged, each offering distinct speed, cost, and functionality attributes.

- **Level 1 Chargers (120V):** Provide around 1.3 kW to 2.4 kW of AC power, requiring over 40 hours to fully charge most EVs. While these chargers are convenient as a basic or emergency option, the slow charging rate typically renders them impractical for long commutes or trips.

- **Level 2 Chargers (240V):** Operate at roughly 3 kW to 20 kW, allowing drivers to add 20–30 miles of range per hour. Designed for both home and commercial settings, Level 2 stations often represent the most common mid-range solution for balancing cost and charging convenience.

- **Level 3 Chargers (DC Fast Chargers):** Provide anywhere from 50 kW to 350 kW DC, cutting charge times significantly (30–60 minutes to reach near-full capacity).

These chargers play an essential role along travel corridors, supporting both local fleets and private EV owners who need rapid turnarounds.

Beyond hardware, implementation strategies focus on maximizing availability. Governments worldwide fund station installations and mandate technical standards for interoperability. Automobile manufacturers and energy companies also invest in fast-charging networks, catering to the growing population of long-distance EV drivers. Partnerships between public agencies, utilities, and private enterprises further streamline station deployment, often with on-site renewable energy integration.

Recent trends underline notable growth in the number of new public chargers and the implementation of advanced features like "Plug and Charge," which simplifies the authentication process. Wireless (inductive) charging, though still in development, may eventually offer an alternative solution in areas such as public transit hubs or specialized commercial use cases.

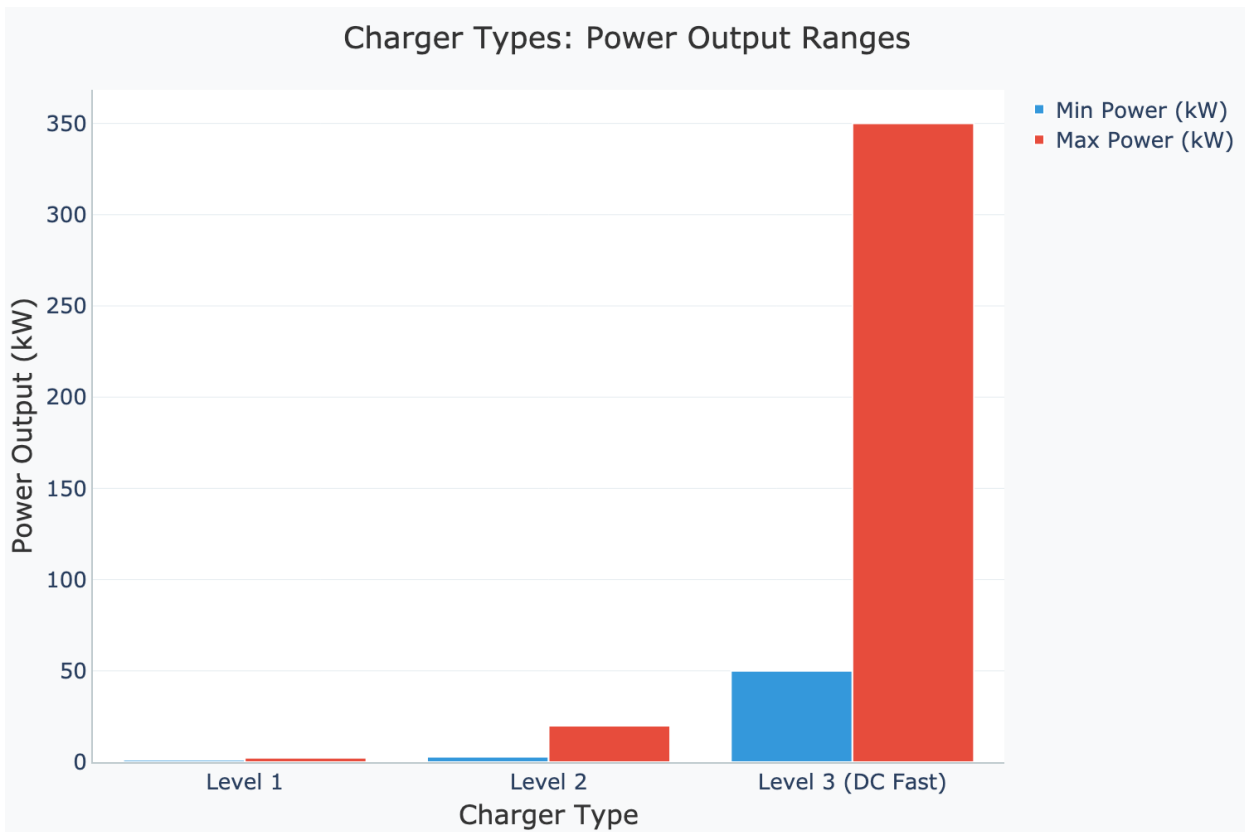

*Figure 10: Power output ranges (kW) across different charger types.*

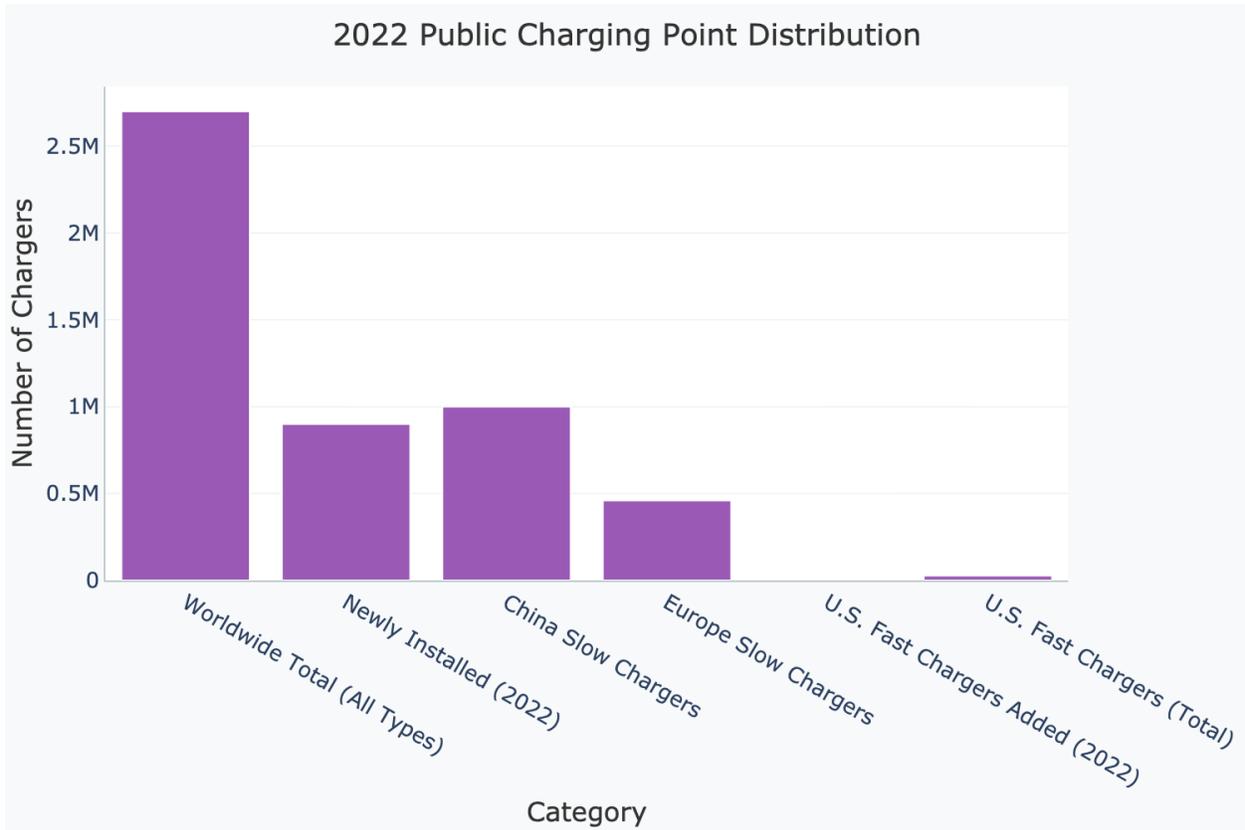

Figure 11: Global distribution of public charging points in 2022, highlighting regional differences.

## 5.2 Grid Capacity and Adaptation

Power utilities are making significant changes to accommodate a surge in electricity demand, particularly from EVs and other electrified sectors such as data centers and new manufacturing facilities. These adaptations include:

- **Multifaceted Investment Approaches:** Many utilities strategically invest in both renewable and conventional generation to balance policy requirements with the need for reliable baseload power. This diversification stabilizes supply amid rapid load growth.

- **Smart Grid Technology Adoption:** Advanced sensor networks, real-time analytics, and automated controls help utilities track and adjust supply-demand balances quickly. Demand-response programs encourage customers to shift usage away from peak periods, easing strain on distribution systems.

- **Renewable Energy Integration:** Widespread adoption of wind and solar necessitates battery storage strategies and updated network planning protocols. Reducing intermittent supply fluctuations is essential to ensure consistent power availability as renewable penetration climbs.

- **Demand Response and Grid Flexibility:** With the rise of electric vehicle charging, utilities deploy demand-response tools alongside dynamic pricing models to motivate EV owners to charge during off-peak hours. These measures preserve grid stability and minimize the need for major capacity expansions.

- **Policy and Manufacturing Drivers:** Government incentives often offset the costs of modernizing substations, transformers, and high-voltage lines, accelerating the rollout of advanced grid components. An upward trend in domestic manufacturing—in machinery, technology, and battery production—adds to the load but also justifies further grid investment.

- **Case Studies and Outlook:** Data centers, with significant electricity needs, benefit from time-dependent tariffs. Advanced metering infrastructure captures real-time usage data, helping operators plan. Simultaneously, pilot initiatives explore vehicle-to-grid (V2G) features, aiming to harness EV batteries as on-demand energy reserves for critical grid support.

## 5.3 Technological Innovations in Infrastructure

Ongoing advances in EV infrastructure technology are redefining charging speeds and energy distribution:

- **Ultra-Fast Charging:** Modern stations can deliver over 300 kW, enabling EVs to recharge significantly within 15–20 minutes. Achieving and sustaining these rates requires robust hardware (e.g., high-capacity transformers) paired with intelligent software that manages load and temperature constraints.

- **Battery Technology Advancements:** While lithium-ion dominates the current EV market, solid-state batteries are gaining traction for higher energy density and potentially faster charge times. There is rising attention on end-of-life management, including recycling programs and second-life applications, to enhance long-term sustainability.

- **Vehicle-to-Grid (V2G) Integration:** By enabling EVs to feed power back into the grid at peak times, V2G offers an alternative model of demand response and distributed storage. Though regulatory frameworks remain in flux, the concept promises to increase renewable energy capacity on the grid and provide financial incentives for EV owners.

- **Outlook and Implications:** Combining ultra-fast charging, cutting-edge battery systems, and V2G potential will likely accelerate EV usage even further. These improvements reduce charging downtime, augment grid resilience, and increase the practicality of electric transport. Collaboration among automakers, utilities, governments, and research bodies remains a primary driver in realizing these innovations.

# 6 Explore the Challenges Facing EV Infrastructure

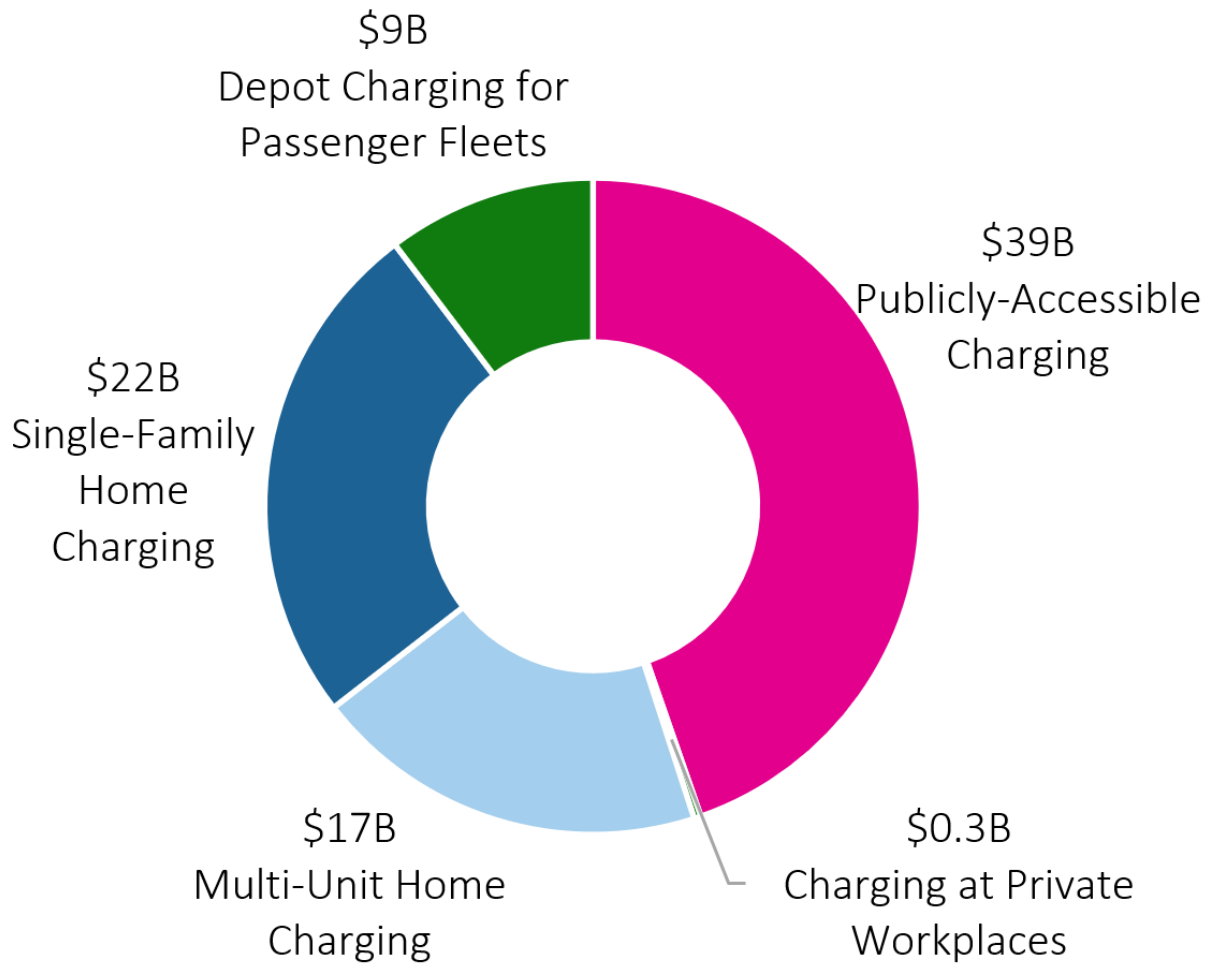

Figure 12: A conceptual view of an EV charging station network.

## 6.1 Costs and Investment Models

**Introduction**

The development and expansion of electric vehicle (EV) charging infrastructure involve substantial costs that fall on both the public and private sectors. Governments worldwide invest in policy incentives, grants, and direct funding, while private companies shoulder operational, installation, and maintenance responsibilities. Understanding how these costs are distributed offers crucial insights into the future trajectory of EV adoption and effective infrastructure planning.

**Public Sector Responsibilities and Investments**

**Government Funding and Incentives**

- Several nations have allocated significant funds to spur the growth of charging networks. In the United States, for example, a notable program dedicates considerable resources each year to develop highway-based charging corridors.

- In Europe, large-scale capital has been deployed through investment banks and public-private partnerships to strengthen fast-charging infrastructure.

- Such programs specifically target underdeveloped or lower-revenue areas, aiming to reduce risk for private investors and ensure broad network coverage.

**Public Charging as Crucial Infrastructure**

- Public investments generally focus on high-impact locations, such as major transit routes and city centers, to address "range anxiety" for prospective EV buyers.

- Governments often fund advancements in fast-charging or alternative technologies, fostering innovation that private entities can adopt at broader scales.

**Private Sector Contributions and Strategies**
**Installation and Operation Costs**

- Automotive manufacturers and dedicated charging providers often bear the bulk of equipment procurement, site preparation, and ongoing operational expenses.

- Ongoing expenses include routine hardware servicing, software updates for charger management systems, and electricity costs. These can be partially recouped via charging fees.

**Public-Private Partnerships**

- A stronger collaboration model has emerged in many regions, blending governmental support (e.g., tax incentives or grants) with private investment to share financial burdens.

- This approach can streamline regulatory approvals and site development while ensuring both parties bear proportionate risk and benefit from long-term gains.

**Regional Variations in Burden and Distribution**
**Developed Markets vs. Emerging Markets**

- In higher EV adoption regions (e.g., parts of Europe and China), significant capital is allocated to augment existing networks and reach more remote areas, while private investments expand coverage in profitable urban corridors.

- Emerging markets often rely on initial government financing to establish baseline charging capacity. Private sector engagement typically follows once demand rises.

**Influence on Adoption Rates**

- Jurisdictions that maintain robust funding programs and encourage private sector participation often see faster growth of charging networks, leading to higher EV uptake rates.

- Regions with limited or erratic support may lag in infrastructure deployment, highlighting the importance of balanced and stable strategies.

**Key Considerations for Ongoing and Future Investments**
**Scaling with Demand**

- As EV sales surge annually, both public and private entities face mounting pressure to address growing demand, necessitating expansions in charger capacity and technological upgrades.

- Governments may adapt or increase long-term incentives to ensure that infrastructure can accommodate future EV fleets without congestion or service shortfalls.

**Balancing Profitability and Accessibility**

- Private operators aim to recover costs by achieving sufficient utilization rates and optimizing operational efficiencies. Governments, meanwhile, must balance near-term spending with sustainability and environmental targets.

- Novel technologies like vehicle-to-grid (V2G) or wireless charging could alter the financial equation by providing additional revenue streams and more efficient resource use.

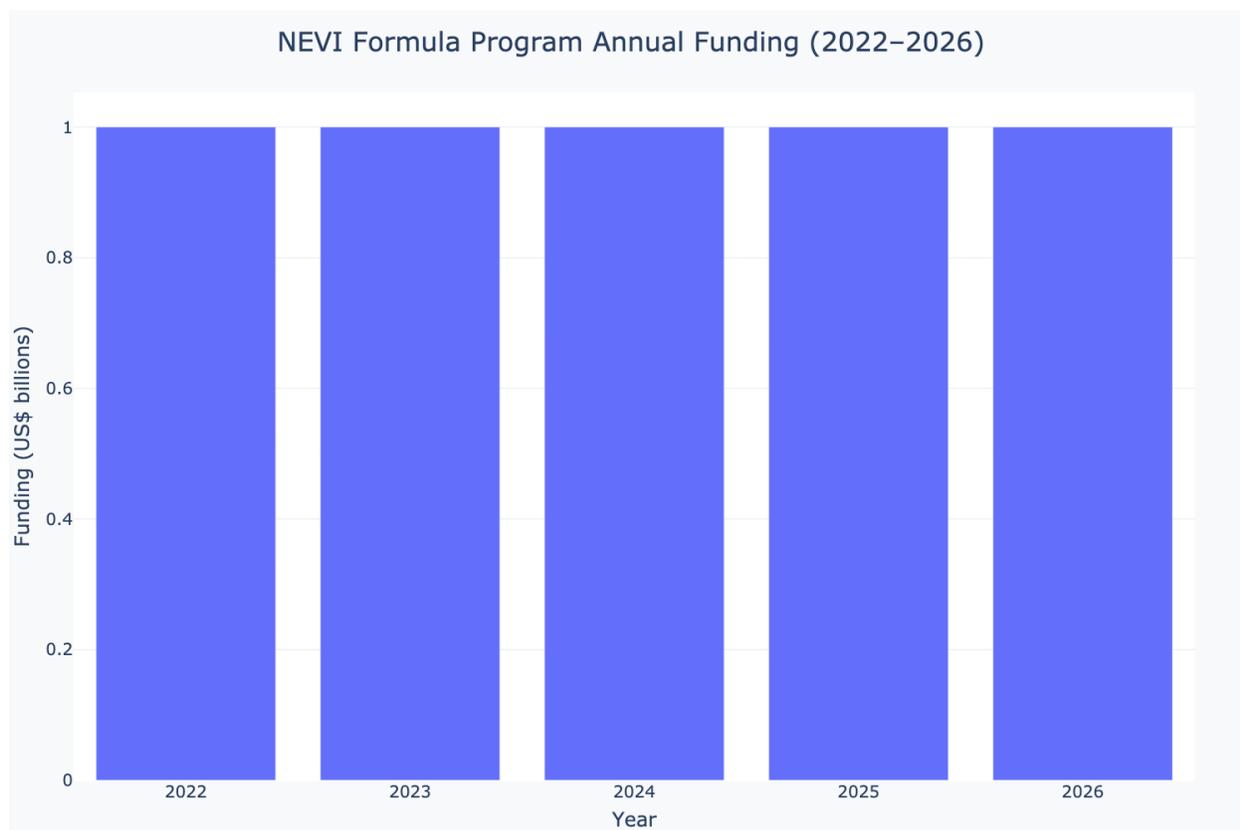

*Figure 13: Annual NEVI Formula Program Funding Allocations (2022–2026).*

The bar chart above illustrates the annual funding allocations for one significant U.S. program, which commits US$1 billion per year from 2022 to 2026. This funding is intended for states to develop charging corridors, with a set-aside to support areas requiring additional assistance.

*Table 2: Financial Investments in MDHD EV Charging Infrastructure (U.S.)*

| Country/Region | Type | Investment Source | Amount (Billion USD) | Notes |
|---|---|---|---|---|
| United States | Public | Federal Government | 21 | Driven by the 2022 federal Inflation Reduction Act (IRA). |
| United States | Public | California State | 2.4 | California Energy Commission funding. |
| United States | Private | Private Sector (U.S. Overall) | 4.2 | Significant increase from previous years. |
| United States | Private | Private Sector (Outside CA) | 0.322 | Tripled from 2021 investment levels. |

As shown above, both the public and private sectors play critical roles in financing medium- and heavy-duty (MDHD) EV charging infrastructure. Federal, state, and industry contributions are all integral factors in building out a robust, nationwide charging ecosystem.

## 6.2 Standardization and Hardware Compatibility

The growth of electric vehicles (EVs) depends not only on advancements in battery technology and consumer incentives but also on the infrastructure that supports them. A key aspect of this support is having unified standards for charging connectors and communication protocols. When manufacturers and charging providers align on standardized hardware and software, EVs can charge reliably and efficiently across different networks.

**Why Unified Standards Matter**

- **Streamlined User Experience:** EV owners benefit from the confidence that any charging station following common connector and protocol specifications will work seamlessly with their vehicle.

- **Reduced Complexity for Stakeholders:** Automakers, charging network operators, utilities, and policymakers can dedicate resources more efficiently when there is consistency across components.

- **Enabling Advanced Capabilities:** Many next-generation features, such as automated billing or vehicle-to-grid integration, rely on standardized data exchange. Without common protocols, these functionalities would remain fragmented and difficult to scale.

**Key Standards and Protocols**
**Connector and Cable Standards**

- IEC 62196: Defines AC and DC charging connectors, addressing higher power levels for evolving EV battery capacities.

- Combined Charging System (CCS): Integrates AC and DC charging in one port, reducing connector complexity.

- Tesla's North American Charging Standard (NACS): Expanding in use and supporting both AC and DC charging.

**Communication Protocol Standards**

- ISO 15118: Bi-directional communication for features such as Plug and Charge and vehicle-to-grid.

- SAE J1772, SAE J2847/2, SAE J2931: Provide definitions for physical and digital interactions between vehicles and chargers.

- OCPP and OCPI: Open protocols for data exchange between charging stations and backend systems, promoting network interoperability.

**Impact on Compatibility**

By adhering to recognized standards—from CCS or NACS for connectors to OCPP for network communications—vehicles charge reliably in different regions and at stations owned by different operators. This eliminates the need for adaptors or specialized apps. For charging network operators, standardization simplifies station management, as diagnostics, updates, and data collection use consistent approaches.

**Effect on Efficiency**

**Physical Infrastructure Efficiency**

Standard connectors and cables reduce production complexity and overall costs, streamlining assembly processes and maintenance across the industry.

**Network Coordination Efficiency**

Open communication protocols allow station operators to monitor performance in real time, deploy uniform software updates, and integrate dynamic load management. Advanced features like vehicle-to-grid rely on consistent, well-defined data formats to balance energy demands effectively.

**Future Directions**

Organizations such as ISO, IEC, and SAE continue to refine charging standards as EV technologies evolve toward faster charging speeds, higher battery capacities, and more integrated grid functionalities. Additionally, security within these communications is increasingly critical, ensuring the charging network remains resilient. Ongoing collaboration among automakers, utilities, and governments aims to create an environment where a single charging solution works for nearly all EVs, thus enhancing consumer confidence and fueling broader adoption.

## 6.3 Regulatory and Policy Obstacles

**Zoning Laws and Building Codes**

Zoning regulations and local building codes frequently limit the pace at which charging stations can be installed. In many jurisdictions, authorities have not yet formally classified EV chargers within existing zoning structures, resulting in ambiguity over permissible locations and requirements. While some localities have introduced "EV-ready" zones, others still rely on outdated ordinances that complicate large-scale deployment.

**Complex Permitting Processes**

Permitting remains one of the primary sources of delay for new installations. Each locale commonly applies unique rules for site assessments, electrical capacity, and structural modifications. Even in regions that mandate fast-tracking of EV charging projects, multiple agencies often must sign off, prolonging the timeline. Current systems often treat EV charger applications like typical construction endeavors, adding extra steps that could be avoided with more specialized guidelines.

**Fragmented Regulatory Framework**

Across federal, state, and municipal levels, inconsistent definitions create a patchwork of overlapping or contradictory requirements. Charging station operators may be classified similarly to utilities in some states, affecting permissible fee structures and obligations. Meanwhile, varied data privacy regulations complicate the transfer and protection of personal information, while liability concerns around hardware upkeep can dissuade businesses from rapid deployment.

**Misalignments Across Sectors**

Energy and transportation policies do not always move in lockstep. Public agencies may adopt electric buses, yet discover that utility budgeting processes cannot immediately support necessary grid upgrades, leading to unanticipated costs. Similarly, private property owners can find it difficult to recoup expenses for charger installations if local rules do not provide clear mechanisms for reimbursement or usage fees. Bridging these sectoral gaps requires coordination among stakeholders who have historically followed different regulatory assumptions.

**Consequences for Infrastructure Expansion**

Taken together, these legal and policy challenges create unpredictable project timelines and ultimately hamper infrastructure rollouts. Even well-funded charging initiatives may stall during regulatory interpretations or court actions. In turn, constrained charging availability slows consumer uptake of electric vehicles. Ensuring legal and policy alignment is therefore paramount to accelerating the timely scale-up of charging infrastructure.

**Potential Policy Reforms**

- **Streamlined Permitting:** Adopting uniform standards and digital review systems specific to EV charging.

- **Coordinated Zoning and Planning:** Updating local zoning codes to reflect modern EV-ready requirements.

- **Regulatory Harmonization:** Encouraging stronger cooperation among transportation agencies, utilities, and building authorities for coherent strategies.

- **Liability and Data Sharing Guidelines:** Standardizing rules for data privacy and legal responsibility to support modern, connected charging solutions.

# 7 Projecting the Future of EV Infrastructure

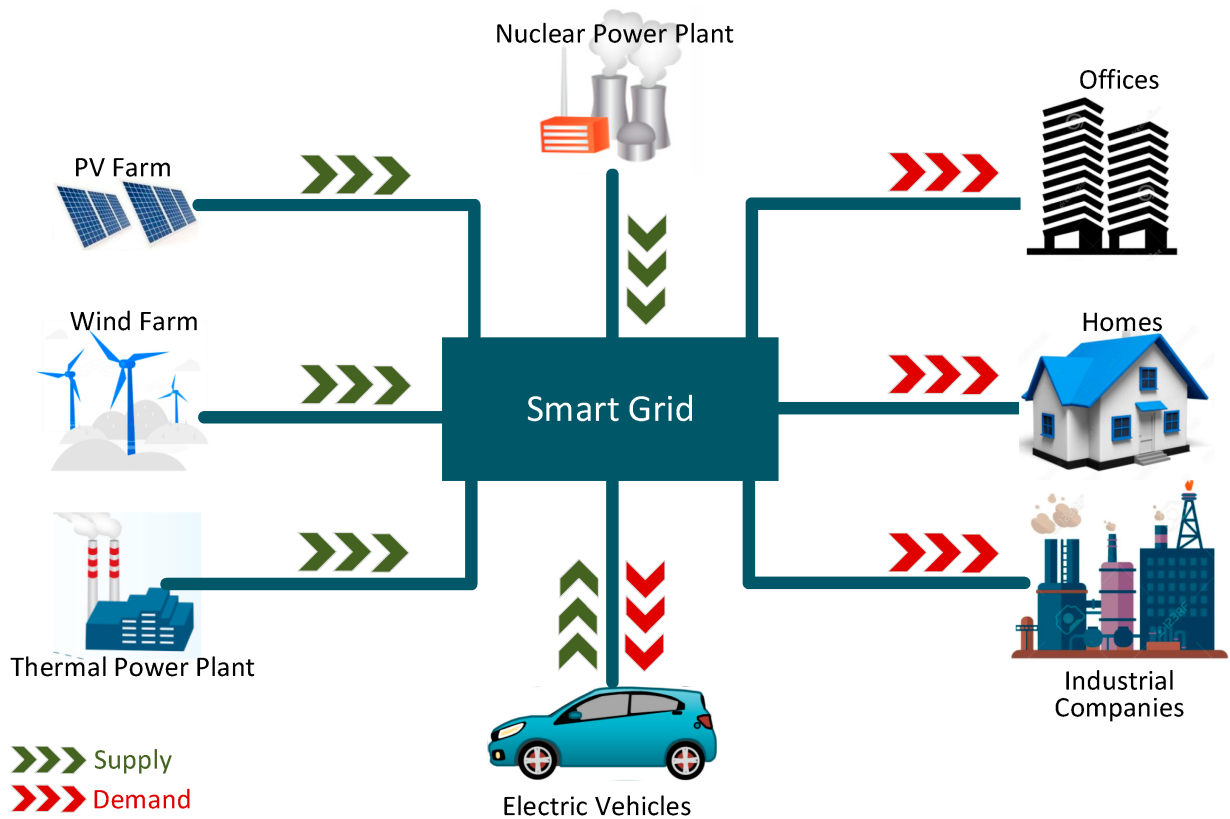

Figure 14: Forecasting next-generation EV infrastructure developments.

## 7.1 Integration with Renewable Energy Sources

Integration of solar, wind, and other sustainable energy sources into electric vehicle (EV) charging networks has become a powerful way to enhance the environmental and economic benefits of e-mobility. Below is a concise yet in-depth overview of how these renewables are being incorporated and the impact they are having:

1. **Introduction** The shift to electric mobility gains additional momentum when the electricity used for charging is generated from clean sources. By capturing solar, wind, and other renewable energies, EV charging networks can reduce their carbon footprints, stabilize local power grids, and offer cost advantages for operators and consumers alike.

2. **Key Technologies and Integration Methods a. AI-Driven Charging Optimization**

- Advanced algorithms assess real-time demand, predict charging station load, and match it with fluctuations in wind or sunlight. This helps ensure continuous availability of renewable power while minimizing strain on the grid.

b. **On-Site Renewable Generation and Storage**

- Charging facilities commonly install solar panels or small-scale wind turbines, paired with battery storage units to balance and store excess energy.

- Second-life batteries, repurposed from EVs, often serve as stationary storage solutions, reducing overall costs and adding resilience to the network.

c. **Vehicle-to-Grid (V2G) and Demand Response**

- Through V2G, EVs can feed stored energy back to the grid during peak consumption or emergencies. This model allows renewable resources, once integrated, to be deployed more flexibly—improving grid stability and reducing the need for additional fossil-fueled generation.

**3. Policy and Regulatory Support a. Federal and National Programs**

- Initiatives in the United States, such as the Inflation Reduction Act (IRA) and the National Electric Vehicle Infrastructure (NEVI) Formula Program, provide tax credits and funding that encourage stations to adopt renewable generation.

b. **State and Local Incentives**

- Programs like New York's Charge Ready NY and California's CalCAP Zero-Emission incentives promote on-site solar installations, streamlined permitting, and financial support aimed at both public and private charging operators.

c. **Private Sector and Utility Engagement**

- Collaboration among charger manufacturers, energy companies, and utilities focuses on standardizing equipment, optimizing load management, and advancing renewable integration.

**4. Cost-Effectiveness and Broader Advantages a. Lower Operational Costs**

- Installing solar panels or wind turbines at charging sites can reduce electricity expenses over the long term, especially when combined with storage systems to offset demand charges or high-peak utility rates.

b. **Enhanced Grid Resilience**

- In areas prone to interruptions or extreme weather, local renewable generation and battery storage reduce reliance on traditional power plants and diversify the energy supply.

c. **Environmental and Social Benefits**

- By powering EV charging stations predominantly with renewable sources, overall greenhouse gas emissions decline, improving local air quality and aligning with decarbonization goals.

**5. Challenges and Future Outlook a. Intermittent Nature of Renewables**

- Wind and solar power can vary throughout the day or season, necessitating robust battery storage, flexible charging strategies, or backup conventional power sources.

**b. Widespread Deployment and Scalability**

- As more sites look to adopt on-site renewables, systematic planning among government entities, utilities, and private operators remains essential for successful scaling.

**c. Ongoing Technological Evolution**

- Many of the most promising innovations—such as advanced V2G protocols and AI-based grid balancing—are still evolving. Overcoming cost and technical complexities is key to broader adoption.

**6. Conclusion** Integrating solar, wind, and other renewable energy options into EV charging networks is fast becoming a cornerstone of sustainable transportation. By merging intelligent software, battery storage, and policy incentives, these systems not only reduce emissions but also increase reliability and can lower operating expenses over time. Although challenges remain—particularly with intermittency and the need for further technological standardization—the overall trend points to growing use of renewables in the EV charging landscape, delivering significant environmental and economic advantages for the future of mobility.

## 7.2 Advancements in Battery Technology and Charging Speed

**Next-Generation Batteries, Faster Chargers, and Their Potential to Reshape EV Adoption**

**1. Introduction** The electric vehicle (EV) landscape is poised for significant change as pioneering battery chemistries and faster charging systems address longstanding hurdles like high costs, limited range, and protracted charging times. While lithium-ion batteries once dominated the EV market, developments in solid-state, lithium-sulfur, and other emerging chemistries have started to shift industry perceptions. Meanwhile, new charging infrastructure, including ultra-fast and wireless options, is taking shape to meet the growing demand for more convenient, efficient refueling solutions.

**2. Advances in Battery Technology a. Solid-State and Lithium-Sulfur Developments** Among the most anticipated advances are solid-state batteries, which replace traditional liquid electrolytes with solid alternatives to deliver higher energy density and superior safety. Companies such as Toyota and Solid Power Inc. are refining these batteries

by integrating sulfide superionic conductors, potentially enabling quicker charge–discharge cycles. On another front, lithium-sulfur batteries leverage sulfur's abundance to reduce costs and improve energy density. Recent prototypes at research institutions like the Korea Electrotechnology Research Institute (KERI) demonstrate hope that this promising technology can mitigate issues like the polysulfide shuttle effect, which affects a cell's long-term stability.

**b. Emerging Chemistries and Ongoing Challenges** Beyond solid-state and lithium-sulfur, other chemistries—such as graphene-based batteries, zinc-air, or aluminum-air—also seek to deliver faster charging, longer range, and improved safety. However, these designs often confront scale-up issues, from manufacturing limitations to raw material logistics. Despite notable test cases and pilot programs, mass-market commercialization remains constrained by technical complexities, production inefficiencies, and cost considerations. Consequently, experts expect incremental improvements to existing lithium-ion technology to remain a near-term focus, supplemented by demonstration projects for cutting-edge chemistries.

**3. Progress in Ultra-Fast Charging a. Higher-Power DC Stations** Charging speed is crucial to bridging the convenience gap with traditional refueling. Ultra-fast chargers, generally rated at 350 kW or above, can shrink charging times substantially. Leading the way is BYD, which has introduced 1,000 kW chargers capable of delivering hundreds of kilometers of range in less than 10 minutes. These milestones, combined with broader infrastructural investments and government support, help build consumer confidence by reducing range anxiety and ensuring a robust network of high-power stations for widespread EV use.

**b. Wireless and Bidirectional Solutions** In parallel, wireless charging is under active development to streamline the charging process further. Through specialized pads in parking areas, vehicles can charge automatically—either while parked or, in some pilot programs, when driving over embedded coils on designated roads. Though existing systems must overcome efficiency and cost barriers, the possibility of hands-free charging remains enticing. Additionally, progress in bidirectional charging—where vehicles both draw power from and return energy to the grid—may reshape local energy management and permit EVs to serve as distributed energy resources for homes and businesses.

**4. Potential Impact on Adoption** Continual improvements in battery capacity, longevity, and charging rates are expected to reduce the total cost of ownership for EVs significantly, boosting consumer interest and adoption. Analysts predict a continued annual growth rate exceeding 30% in EV sales over the coming decade, with a potential market valuation surpassing USD 4 trillion by the 2030s. Key factors influencing this growth include:

- Reduced charging times that approach the convenience of traditional fueling.

- Lower battery costs, leading to more competitive EV sticker prices.

- Greater energy density and overall driving range, alleviating anxiety about running out of charge.

- Expansion of charging infrastructure, further normalizing EV usage.

Policy direction remains a substantial variable; robust government incentives, stronger emissions regulations, and infrastructure investments usually amplify consumer willingness to embrace EVs. In places with supportive regulatory frameworks, businesses and consumers are more likely to adopt the latest generation of EVs. Conversely, inconsistent policy environments or funding stalls can slow the build-out of essential charging networks and hinder technology deployment.

5. **Concluding Remarks** Next-generation battery research and the rollout of ultra-fast chargers together signal a pivotal shift in how society powers transportation. While commercial-scale adoption of some emerging chemistries is still on the horizon, even incremental improvements to existing lithium-ion cells boost performance and affordability in the near term. In parallel, ultra-fast, wireless, and bidirectional charging innovations are steadily breaking down barriers to widespread EV uptake. These developments, supported by coordinated policy, private investment, and an evolving consumer mindset, stand to position EVs as a practical, mainstream choice capable of reshaping global mobility.

## 7.3 Smart Charging and V2G Outlook

Smart charging and vehicle-to-grid (V2G) technologies are poised to become integral elements of the electric vehicle (EV) ecosystem. By dynamically managing how and when EVs draw power from—or send electricity back to—the grid, these solutions promise more efficient energy allocation, reduced strain at peak times, and the potential for lower operating costs for both consumers and utilities.

1. **Balancing Loads and Peak Shaving**

- **Demand Shifting:** Smart charging systems can tailor EV charging rates to align with off-peak periods, preventing large surges in electricity demand when overall grid usage is high. This adjustment not only offers cost savings to drivers via time-of-use rate structures but also helps utilities smooth out daily demand fluctuations.

- **Renewable Integration:** By adapting charging patterns to periods of abundant wind or solar generation, smart charging can make better use of intermittent renewables and bolster energy sustainability.

- **Deferred Infrastructure Upgrades:** When properly coordinated, these peak-shaving strategies can delay or reduce the need for costly expansions in transmission and distribution networks.

2. **Enhancing Grid-to-Vehicle Synergy**

- **Bidirectional Flow:** V2G technology lets EV owners discharge a portion of their battery energy back into the grid. During peak load or grid stress events, this reserve power can support system stability, much like a distributed network of batteries.

- **Real-Time Coordination:** Advanced software and Internet of Things (IoT) platforms monitor electricity prices, grid conditions, and battery states to ensure that EVs provide power or charge up at optimal times.

- **Resilience Benefits:** Aggregating thousands of vehicles effectively creates a mobile energy reserve. In regions prone to grid disruptions—be they weather-related or demand-driven—the ability to leverage EV batteries can strengthen the entire electricity ecosystem.

3. **Technological Evolution and Future Outlook**

- **Rapid Market Growth:** Analyst forecasts consistently project strong growth in V2G solutions, with some estimates predicting market valuations increasing more than sevenfold over the coming decade. Similarly, the market for smart EV chargers is expanding rapidly, propelled by consumer demand for home energy management, private-sector fleet electrification, and large-scale utility partnerships.

- **Innovative Hardware and Software:** Ongoing developments in power electronics, artificial intelligence, and secure transactional technologies (e.g., blockchain) refine grid-vehicle interactions. These innovations improve charging efficiency, limit wear on EV batteries, and streamline the user experience.

- **Policy and Incentives:** A combination of stricter emissions limits, broader renewable energy mandates, and financial benefits—like tax credits and rebates—encourages EV owners and fleet operators to adopt charging hardware capable of grid interactions. As these policies gain momentum, further scaling of both smart charging and V2G is expected.

- **Implementation Challenges:** Key technical concerns include battery degradation from increased cycling, the establishment of universal communication protocols, and the need for robust cyber-physical security across charging networks. Addressing these aspects ensures consistent performance and system reliability.

4. **Implications for EV Infrastructure**

- **Station Design and Placement:** Anticipated load swings and two-way power flows may drive station designs with greater onsite energy storage—potentially paired with solar canopies or wind systems. These sites can buffer power demands and supply, offering local resiliency.

- **Enhanced Software Integration:** "Intelligent" management systems will communicate with vehicles, grid assets, and renewable generators in real time, turning each charging point into a potential micro-hub for balancing local energy needs.

- **Community and Urban Planning:** New developments, such as residential complexes or commercial districts, increasingly incorporate advanced EV charging capabilities. This proactive approach can prevent future grid congestion and secure affordable power supplies.

In summary, smart charging and V2G technologies are set to transform how vehicles draw and deliver energy, serving a dual role in supporting the grid while optimizing charging costs for EV owners. Although factors like battery longevity, policy support, and interoperability standards require continued attention, these technologies stand at the forefront of a more flexible, resilient, and sustainable energy future.

# 8 Examine Regional Variations in EV Adoption and Infrastructure

This section explores how different regions approach electric vehicle (EV) adoption and infrastructure development, highlighting the diverse policy measures, successful large-scale rollouts, and lessons that can inform broader international strategies.

## 8.1 Divergent Policy Frameworks

Electric vehicles have evolved from niche products to viable transportation solutions, thanks in large part to policy frameworks designed to incentivize consumer uptake and guide manufacturers. Governments around the world have introduced or strengthened mandates, emissions regulations, financial incentives, and infrastructure initiatives.

**United States.** Federal strategies encourage EV adoption through consumer tax credits, which can reach up to US$7,500 for new EVs. Additional credits also apply to certain pre-owned EVs. The Inflation Reduction Act expands coverage to commercial and fleet vehicles, reflecting a broader electrification plan. In parallel, stricter emissions rules under the current administration target lowering carbon output, with projections showing a significant increase in EV share of new passenger vehicle sales.

**China.** China's Dual-Credit Mechanism employs a cap-and-trade-style approach, tying fuel-efficiency standards to the production of New Energy Vehicles (NEVs). This system powerfully motivates automakers to emphasize EV models in their portfolios. Production and sales targets under the NEV Industry Development Plan ensure continued growth, reinforced by roadmaps that address both technology and infrastructure needs.

**Europe.** Stringent emissions mandates aim to reduce $CO_2$ emissions by 55% for new cars (and by 50% for vans) by 2030—stepping up to a requirement that new passenger vehicles reach zero-emission status no later than 2035. Member states promote extensive charging infrastructure and e-mobility innovation, partly by providing funding for advanced battery and hydrogen fuel cell technologies.

**Observations and Future Outlook.** From the United States' consumer-centric approach to China's high-volume NEV strategy and Europe's robust emissions mandates, it

is clear that every region's policy framework is tailored to its market and industrial ecosystem. Nevertheless, the common thread is a commitment to furthering EV adoption and supporting the infrastructure required for cleaner, more sustainable transport.

*Table 3: Select EV Policies, Emissions Mandates, and Infrastructure Initiatives*

| Region | Key Policy / Incentives | Emissions & Mandates | Infrastructure & Support | Additional Notes |
|---|---|---|---|---|
| United States | Consumer Tax Credits: up to US$7,500 for new EVs; extra credits for eligible used EVs Inflation Reduction Act expands credits to commercial fleets | Strengthened federal emissions rules for lowering carbon output | Emphasis on expanding charging for both consumers and fleets Broader push to electrify medium- and heavy-duty vehicles | potential portion passen by ear |
| China | Dual-Credit Mechanism incentivizes NEV production NEV Industry Development Plan (2021–2035) sets ambitious EV targets | Cap-and-trade-style system ties fuel-efficiency standards to EV output | Demonstrates large-scale EV production Roadmaps guide technology advances and widespread market penetration | extend tery plug-in and vehicle |
| Europe | Rigorous mandates: 55% CO₂ cuts for cars by 2030; zero-emission requirement by 2035 | Strict $CO_2$ targets for automakers, pushing zero-emission technologies | EU funds e-mobility innovation, including advanced battery production Member states prioritize wide charging coverage | to pres compe for autom |

## 8.2 Case Studies of Effective Implementation

Large-scale EV adoption transcends borders, emerging in contexts as varied as dense urban environments and resource-constrained settings. Each success story offers a distinctive approach:

**Korea.** Korea's strategy revolves around expanding high-speed chargers in core urban areas, easing range anxiety and driving market growth. Collaborative government and industry efforts align mandates with pragmatic infrastructure rollouts. These actions showcase how a well-planned network from the outset can accelerate uptake.

**Norway.** Norway boasts one of the highest EV penetration rates. Generous financial incentives, such as tax exemptions, combine with extensive home and public charging opportunities. This confluence of policies has embedded EVs deeply in everyday life, ensuring that vehicle electrification remains viable across the entire country.

**United Kingdom.** A focus on "smart charging," supported by progressive building codes requiring charging points in new developments, plays a substantial role. Public-private collaboration bridges land ownership complexities, supplementing broad programs that encourage off-peak charging and modernization of charging stations.

**United States and Canada.** While both countries must contend with extensive geographies and older grid systems, incentives at various governmental levels stimulate EV purchasing and infrastructure upgrades. Utilities and local governments plan for rising charging demand, modernizing grids through strategic investment in transmission and distribution.

**India and Regions of Africa.** Battery-swapping programs address longer charging times and uneven electricity supply. Collaborations often pair governmental backing with private-sector efficiency and innovation. India's pilot swapping initiatives and similar programs in parts of Africa illustrate the creativity needed to overcome infrastructure hurdles.

**Collective Observations.** Across every region, public-private partnerships anchor EV rollouts, offering a balanced formula of incentives, regulatory support, and commercial ingenuity. Such models unify efforts around funding, technology integration, and the modernization of grid systems, leading to faster and more reliable EV adoption.

## 8.3 Lessons for Wider Application

From financial incentives to infrastructure innovations, practitioners worldwide have converged on a few fundamental strategies to advance EV uptake.

**Government-Led Measures and Emissions Mandates.** Streamlined permitting and tax benefits remain key to motivating early adoption. Regions that tie overall emissions goals and production quotas to EV deployment (e.g., emission credit systems, zero-emission targets) often experience steady growth in EV manufacturing and supportive infrastructure rollouts.

**Battery Swapping and High-Speed Charging.** Battery swapping offers an alternative to fast-charging in segments such as public transport or last-mile delivery. At the same time, the deployment of ultra-fast chargers in more conventional consumer markets mitigates range anxiety. These methods each serve specific mobility patterns and societal contexts.

**Public-Private Partnerships.** Whether through enormous capital outlays for charging networks or through domestic manufacturing mandates, most sustainable EV programs blend government-led initiatives with market-driven solutions. Such partnerships leverage complementary resources, from financial incentives and policy support to technical expertise and operational efficiency.

**Consumer Engagement and Education.** Outreach programs can remove common misconceptions around charging times and total ownership costs. Parallel incentives—tax benefits, reduced electricity rates, or free parking—reinforce consumer confidence in electric mobility.

**Smart Charging and Renewable Integration.** Avoiding grid stress calls for advanced load management, aligning EV charging with off-peak hours or renewable energy peaks. Coupling solar or wind energy production with battery storage at charging stations also fosters a complete transition to low-carbon transportation.

**Flexibility and Scale-Up.** Every locale features unique cultural, economic, and infrastructural realities. Successful programs share a foundation in government action, industry alignment, and education, but customization remains crucial. Battery-swapping in emerging markets, for instance, suits short, frequent trips, whereas high-power charging models take hold in densely populated areas seeking convenience.

To facilitate further comparison of multiple regions' progress, policy goals, and market

shares, the following figure highlights how EV adoption rates differ across China, Europe, and the United States:

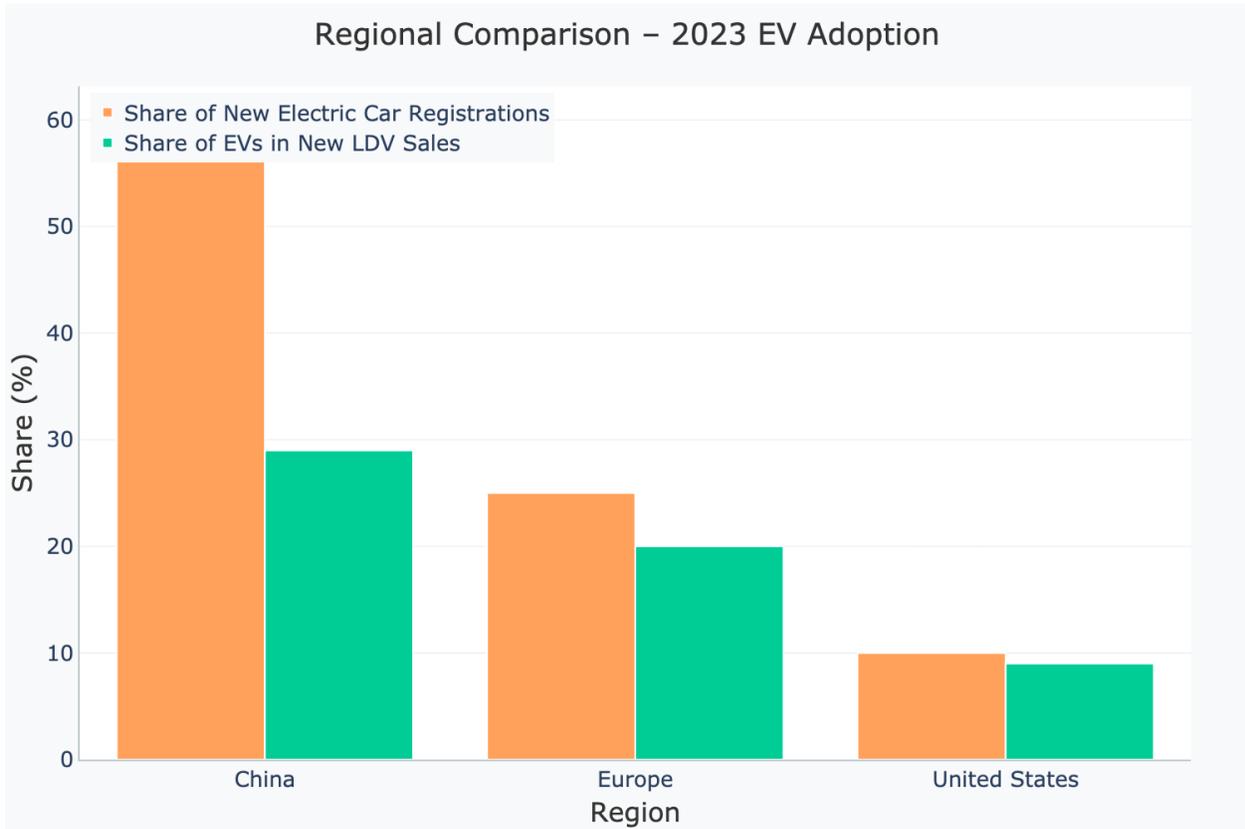

Figure 15: Regional Comparison – 2023 EV Adoption. Share of new electric car registrations vs. share of EVs in new LDV sales.

# 9 Forward-Looking Approaches to EV Adoption

## 9.1 Ongoing Research and Development

Cutting-edge developments in EV infrastructure revolve around three main areas: the evolution of solid-state batteries, the introduction of high-power wireless charging, and advancements in charging solutions designed to enhance access, speed, and sustainability.

1. Solid-State Batteries

Research into solid-state batteries has accelerated as scientists and automotive manufacturers strive for innovations beyond traditional lithium-ion cells. In particular, leading university teams are exploring anode-free solid-state designs to significantly increase energy density. These efforts aim to:

- Enable EVs to travel 500 miles or more on a single charge.

- Improve safety by reducing the risk of thermal runaway associated with liquid electrolytes.

- Lower overall weight while increasing the battery's lifetime.

Researchers do note, however, that practical hurdles remain in scaling to commercial production. Issues include high manufacturing costs, materials sourcing, and ensuring consistent quality at volume. Even so, ongoing breakthroughs in chemistry and materials science suggest that solid-state technology could begin making its way into premium or specialized EVs in the not-too-distant future.

2. Wireless Charging Innovations

Wireless charging is evolving beyond small-scale demos to real-world commercial and transit applications. High-power inductive systems—some offering up to 450 kW—are already being integrated into electric bus platforms in select regions. These installations:

- Provide hands-free, automated charging without cables.

- Reduce downtime for commercial fleets by streamlining the charging process.

- Allow for safer and more flexible charge points in congested or space-constrained environments.

Although personal passenger-car adoption of wireless charging is still in its early stages, this technology's rapid scalability and improving performance point to broader deployments for individual consumers over the next several years.

3. Other Pioneering Topics in EV Infrastructure

**(a) Ultra-Fast Charging and Interoperability**

The industry is witnessing a marked increase in ultra-fast chargers capable of delivering power that can replenish 80% of a battery in around 20 minutes. Beyond sheer speed, interoperability standards—often facilitated by open communication protocols—are making it easier for EV owners to use various public charging networks. This user-friendly approach reduces barriers to adoption and promotes a more consistent charging experience.

**(b) Expanded Charging Networks and Grid Integration**

Charging services are broadening their geographic reach, moving beyond city centers to provide coverage for rural and suburban areas. These efforts, often supported by governmental incentives or public-private collaborations, seek to guarantee reliable charging access for all communities. Complementing the physical rollout, grid integration practices such as dynamic load balancing and proactive renewable energy use help ensure charging can expand without compromising electrical reliability.

**(c) Cost Reductions and Sustainability**

Sustained innovation and production efficiencies across the EV industry continue driving down costs—both for vehicles and charging infrastructure. As these expenses diminish, they lower the economic barriers to EV adoption. Taken together with ongoing policy efforts promoting cleaner fleets, this cost trajectory aligns with broader aims of cutting carbon emissions and increasing air quality in densely populated areas.

Overall, pioneering work on solid-state battery chemistry, faster and more convenient charging methods, and forward-thinking infrastructure deployment is charting a path toward a future where electric vehicles are both easier to own and far more sustainable. While challenges remain—especially in large-scale production and the establishment of universal standards—the progress underway consistently addresses some of the most significant hurdles faced by the EV sector.

## 9.2 Anticipated Market Shifts

The electric vehicle (EV) sector is witnessing significant market shifts characterized by evolving consumer trends, proactive automaker strategies, and innovative business models. Consumer sentiment toward EVs is strengthening, with around 23% of consumers "very likely" to choose an EV for their next vehicle purchase. The retail share of EVs has surpassed 10%, moving beyond a niche to a mainstream choice. Brand reputation is crucial, with Tesla's market share slightly declining due to increased competition from Chevrolet and other automakers. EVs maintain an edge in total cost of ownership, with lower fuel and maintenance costs offsetting higher initial prices. Technological advancements, such as solid-state and aluminum-ion batteries, are reshaping buyer expectations by addressing range and charging time concerns.

Automakers are aligning product lines with EV demand, focusing on technology-centric approaches, financial accessibility, and competitive positioning. New business models include battery swapping networks, consumer-friendly financing, and integrated energy solutions, which are redefining the EV landscape. The market is projected to surpass US$4 trillion by 2033, with growth influenced by government incentives, battery cost reductions, and user-friendly charging solutions. The sector's success will depend on consumer acceptance of new technologies and business models.

## 9.3 Cross-Sector Collaboration

Inter-industry alliances have become a pivotal force in accelerating EV infrastructure growth. These collaborations unite automakers, energy corporations, technology enterprises, and government bodies to address the complex requirements of EV charging networks. By pooling expertise in automotive engineering, renewable energy, and digital technology, these partnerships more effectively scale charging infrastructure and enable seamless integration with the power grid.

1. Recent Partnerships and Initiatives

**(a) Pilot Company, General Motors, and EVgo**

One prominent example is the collaboration between Pilot Company, General Motors, and EVgo. Together, they have installed more than 130 fast-charging sites in over 25 U.S. states. By strategically positioning chargers along major travel corridors, these partners help alleviate range anxiety and boost consumer confidence in electric vehicle ownership.

**(b) NaaS Technology and Xiaomi Auto**

Another illustrative partnership is that between NaaS Technology and Xiaomi Auto. Their joint venture centers on creating a smart EV charging ecosystem that leverages both technology-driven innovations and robust automotive expertise. By focusing on interoperability and digital integration, they aim to make the charging process more user-friendly and adaptable to evolving standards.

2. Policy Frameworks and Government Facilitation

A key driver behind many of these alliances is the evolving policy environment, which places a premium on clean transportation solutions. For instance, the U.S. Department of Energy's "Strategy for Achieving a Beneficial Vehicle Grid Integration Future" underscores the importance of synchronizing transportation electrification with grid reliability. Such government-supported initiatives provide both regulatory clarity and financial incentives, encouraging collaboration among stakeholders.

3. Impact on Infrastructure Expansion

By combining resources across multiple sectors, these alliances can:

- Rapidly expand the geographic coverage of charging stations, making EV ownership more seamless for a broader audience.

- Integrate solutions that leverage advanced grid management, such as vehicle-to-grid (V2G), reducing stress on the power system and aiding overall grid stability.

- Pool research and development efforts, facilitating faster innovation in battery technology, charging speeds, and the reliability of charging networks.

4. Lessons and Future Directions

- Successful alliances typically benefit from clear role definitions, reliable funding, and strong technology-based synergies.

- Collaborations aligned with government initiatives often progress more smoothly, as policy frameworks help coordinate infrastructure development by offering both guidance and incentive structures.

- The trend points toward even deeper cross-industry coordination, given the growing demand for EV-friendly infrastructure and the universal push for energy sustainability.

Taken together, these partnerships highlight how collectively leveraging expertise from the automotive, energy, and technology sectors can effectively overcome many of the challenges still facing EV infrastructure rollout. Looking to the future, such alliances are poised to advance both clean transportation and power resilience on a global scale.

## 9.4 Anticipated Market Shifts

The electric vehicle (EV) sector is witnessing significant market shifts characterized by evolving consumer trends, proactive automaker strategies, and innovative business models. Consumer sentiment toward EVs is strengthening, with around 23% of consumers "very likely" to choose an EV for their next vehicle purchase. The retail share of EVs has surpassed 10%, moving beyond a niche to a mainstream choice. Brand reputation is crucial, with Tesla's market share slightly declining due to increased competition from Chevrolet and other automakers. EVs maintain an edge in total cost of ownership, with lower fuel and maintenance costs offsetting higher initial prices. Technological advancements, such as solid-state and aluminum-ion batteries, are reshaping buyer expectations by addressing range and charging time concerns.

Automakers are aligning product lines with EV demand, focusing on technology-centric approaches, financial accessibility, and competitive positioning. New business models include battery swapping networks, consumer-friendly financing, and integrated energy solutions, which are redefining the EV landscape. The market is projected to surpass US$4 trillion by 2033, with growth influenced by government incentives, battery cost reductions, and user-friendly charging solutions. The sector's success will depend on consumer acceptance of new technologies and business models.

## 9.5 Cross-Sector Collaboration

Inter-industry alliances have become a pivotal force in accelerating EV infrastructure growth. These collaborations unite automakers, energy corporations, technology enterprises, and government bodies to address the complex requirements of EV charging networks. By pooling expertise in automotive engineering, renewable energy, and digital technology, these partnerships more effectively scale charging infrastructure and enable seamless integration with the power grid.

1. Recent Partnerships and Initiatives

**(a) Pilot Company, General Motors, and EVgo**

One prominent example is the collaboration between Pilot Company, General Motors, and EVgo. Together, they have installed more than 130 fast-charging sites in over 25 U.S. states. By strategically positioning chargers along major travel corridors, these partners help alleviate range anxiety and boost consumer confidence in electric vehicle ownership.

**(b) NaaS Technology and Xiaomi Auto**

Another illustrative partnership is that between NaaS Technology and Xiaomi Auto. Their joint venture centers on creating a smart EV charging ecosystem that leverages both technology-driven innovations and robust automotive expertise. By focusing on interoperability and digital integration, they aim to make the charging process more user-friendly and adaptable to evolving standards.

2. Policy Frameworks and Government Facilitation

A key driver behind many of these alliances is the evolving policy environment, which places a premium on clean transportation solutions. For instance, the U.S. Department of Energy's "Strategy for Achieving a Beneficial Vehicle Grid Integration Future" underscores the importance of synchronizing transportation electrification with grid reliability. Such government-supported initiatives provide both regulatory clarity and financial incentives, encouraging collaboration among stakeholders.

3. Impact on Infrastructure Expansion

By combining resources across multiple sectors, these alliances can:

- Rapidly expand the geographic coverage of charging stations, making EV ownership more seamless for a broader audience.

- Integrate solutions that leverage advanced grid management, such as vehicle-to-grid (V2G), reducing stress on the power system and aiding overall grid stability.

- Pool research and development efforts, facilitating faster innovation in battery technology, charging speeds, and the reliability of charging networks.

4. Lessons and Future Directions

- Successful alliances typically benefit from clear role definitions, reliable funding, and strong technology-based synergies.

- Collaborations aligned with government initiatives often progress more smoothly, as policy frameworks help coordinate infrastructure development by offering both guidance and incentive structures.

- The trend points toward even deeper cross-industry coordination, given the growing demand for EV-friendly infrastructure and the universal push for energy sustainability.

Taken together, these partnerships highlight how collectively leveraging expertise from the automotive, energy, and technology sectors can effectively overcome many of the challenges still facing EV infrastructure rollout. Looking to the future, such alliances are poised to advance both clean transportation and power resilience on a global scale.

# 10 Opportunities for Stakeholders in the EV Market

## 10.1 Strategic Partnerships and Investments

Cooperative ventures involving government entities, automakers, and energy providers play a pivotal role in accelerating the expansion of electric vehicle (EV) infrastructure and boosting overall adoption rates. By leveraging the expertise and resources of each stakeholder group—policy direction and funding from governments, technological know-how and consumer outreach from automakers, and grid management plus power distribution from energy

firms—these collaborations help address critical challenges in building a robust, future-ready transportation ecosystem.

1. Large-Scale Charging Networks One of the most notable examples is the formation of high-powered charging networks through joint ventures that bring together leading automakers and supportive government frameworks. By pooling investments and streamlining approvals for charger installations, such alliances allow for swifter deployment of infrastructure along highways and in urban hubs. This broad coverage not only enhances consumer confidence in EVs but also ensures that charging solutions meet stringent standards set by various stakeholders.

2. Highway Corridor Initiatives In North America, partnerships such as those involving major travel center operators, automakers, and specialized charging providers place fast-charging stations along critical highway routes. State and federal agencies commonly assist by offering grants or tax credits, simplifying land-use permits, and providing guidelines to ensure new charging sites integrate with local grid requirements. These efforts make long-distance EV travel far more viable for drivers while easing concerns about range anxiety.

3. Fleet Electrification Programs Electrified fleets represent a high-impact opportunity to lower greenhouse gas emissions, especially in densely populated areas. Automakers, logistics-focused technology firms, and public utilities often coordinate with municipal governments to encourage large-scale fleet conversions. By working together on streamlined grid connections, specialized charging setups, and subsidized acquisition costs, these cross-sector endeavors guide high-mileage commercial vehicles toward cleaner powertrains and faster adoption.

4. International and Emerging Market Collaborations Some of the most dynamic growth arises in markets where governmental bodies see EV infrastructure as a strategic opportunity for economic development. Joint ventures among automakers, start-ups offering services like battery swapping, and established energy providers are often encouraged through supportive regulatory targets and subsidized partnerships. In many emerging markets, these frameworks aim to tackle air quality issues, reduce dependence on imported fossil fuels, and foster new, sustainable industries.

5. Grid and Energy Management Innovations As EV adoption continues expanding, cooperative ventures that align charging with grid capacity are also gaining traction. Pilot projects, often backed by government funding, use smart scheduling to shift vehicle charging to off-peak hours or coordinate with renewable energy availability. Automakers contribute technical interfaces that enable two-way communication, and power providers focus on upgrading distribution grids to handle the new electrical loads. This integrated approach helps prevent strain on existing infrastructure and promotes clean energy usage.

Ultimately, these multi-stakeholder partnerships exemplify how coordinated efforts across the public and private sectors can stimulate EV adoption, enhance charging accessibility, and support broader clean-energy goals. By combining government-led policy incentives, automaker technology and consumer reach, and energy providers' operational expertise, cooperative ventures stand at the forefront of accelerating our transition to a more sustainable transportation landscape.

## 10.2 Socioeconomic and Environmental Benefits

Electric vehicle (EV) adoption offers considerable socioeconomic and environmental advantages by driving job creation, improving public health, and reducing harmful emissions:

1. Job Creation EV adoption stimulates economic growth by creating new opportunities in both established and emerging industries.

- In manufacturing, demand for advanced batteries, electric motors, and specialized vehicle parts fosters employment at battery production facilities, automotive plants, and logistics hubs.

- Infrastructure development, particularly the deployment of charging stations, provides employment for engineers, electricians, construction workers, and maintenance personnel.

- Government policies and incentives accelerate EV-related job growth by encouraging R&D and deployment, leading to a diverse range of positions in fields like software, data analytics, utilities, and renewable energy storage.

2. Public Health Electric vehicles offer notable public health benefits by significantly reducing pollutants and noise:

- Replacing internal combustion engines with EVs lowers emissions of particulate matter and nitrogen oxides, mitigating respiratory and cardiovascular risks.

- EV motors generate far less noise than combustion engines, which in turn can improve sleep quality and reduce stress in urban settings.

3. Emissions Reductions From a climate perspective, EV adoption drives significant reductions in greenhouse gas outputs:

- Even accounting for electricity generation, EVs generally exhibit a smaller carbon footprint than gasoline vehicles, especially in regions with higher shares of renewable energy.

- Widespread electrification could cut over 2 gigatonnes of CO emissions worldwide by the mid-2030s, further amplified by ongoing advances in battery chemistry, recycling, and clean-grid integration.

Taken together, these benefits highlight how electric mobility not only propels job growth and technological innovation but also contributes to healthier communities and a more sustainable environment.

## 10.3 Recommendations for Accelerating EV Adoption

Below are targeted, actionable steps designed to guide policymakers, industry leaders, and local communities in advancing EV deployment and infrastructure:

- **Strengthen and Harmonize Policy Frameworks** Policymakers can enforce stricter emissions standards, provide financial incentives, streamline permitting processes, and invest in R&D for advanced battery technologies. For example, several mandates in the EU and the U.S. Inflation Reduction Act encourage cleaner transport by setting CO fleet targets, purchase subsidies, and rebates. Specific actions include:
  - Enforce stricter emissions standards and gradually tighten CO fleet targets.
  - Provide comprehensive financial incentives—purchase subsidies, tax breaks, and rebates.
  - Streamline permitting and regulatory processes for charging infrastructure.
  - Invest in R&D programs focused on advanced battery innovations.

- **Engage Industry and Automakers in Collaborative Deployments** Industry leaders can form strategic partnerships, develop standardized platforms, promote fleet electrification, and strengthen consumer marketing. Some initiatives in North America and Europe involve collaborations between automakers, utilities, and charging providers. Key measures include:
  - Coordinate on charging station deployments.
  - Develop standardized connectors and network management platforms.
  - Encourage large-scale fleet electrification with clear cost-benefit messaging.
  - Strengthen consumer marketing to address range anxiety and highlight cost advantages.

- **Empower Local Communities and Grassroots Campaigns** Municipalities and community groups play a central role in fostering neighborhood-level charging, educational outreach, and zoning changes. Progressive cities like Oslo have instituted successful programs. Recommended efforts include:
  - Launch local charging initiatives tailored to neighborhood needs.
  - Collaborate with municipal authorities on EV-friendly planning and zoning.
  - Host community events, public test drives, and hands-on workshops to demystify EV usage.
  - Introduce EV-preferential codes for new building developments.

- **Foster Cross-Sector Collaboration and Data Sharing** Policymakers, industry leaders, utilities, and local stakeholders should coordinate on charging usage, grid capacity, and consumer adoption. Public-private alliances sometimes integrate renewable energy solutions alongside EV infrastructure. Suggested approaches include:

    - Exchange real-time data on charging patterns through common platforms or aggregated databases.
    - Identify high-demand areas for targeted infrastructure investments.
    - Integrate renewables and adopt strategies that reinforce grid resilience.

- **Emphasize Long-Term Sustainability and Equity** Ensuring broad societal gains entails extending incentives to underserved communities, supporting eco-friendly supply chains, and adopting clear monitoring frameworks. This might include transparent guidelines for responsible battery mineral sourcing. Recommended steps:

    - Provide targeted incentives for lower-income neighborhoods and rural regions.
    - Support environmentally responsible battery production and recycling.
    - Employ consistent monitoring and reporting standards for EV program impacts.

# 11 Conclusion

Electric vehicle (EV) adoption has undergone a profound transformation in recent years, fueled by technological improvements, supportive policy frameworks, and increasingly favorable consumer perceptions. The emergence of advanced battery chemistries and more efficient propulsion designs has addressed concerns about range, maintenance, and performance. Charged by expanded incentives and stronger emissions mandates, governments worldwide have played a pivotal role in accelerating the shift toward cleaner automotive solutions. In parallel, automakers have invested heavily in research and development to broaden their EV model offerings, improve affordability, and cultivate greater mainstream appeal.

Enabling infrastructure—particularly a widespread, reliable charging network—remains critical to sustaining the EV movement. Public and private stakeholders benefit from collaborative arrangements that fund station deployment, modernize grid systems, and adopt unifying standards. Through dynamic load management, renewable energy integration, and emerging technologies such as vehicle-to-grid (V2G), the compatibility of EVs with overall energy ecosystems continues to evolve. Such solutions address not only individual driver needs but also broader societal objectives, including reduced emissions, enhanced public health, and energy resilience.

Although new challenges will likely emerge, many of the barriers once stalling EV adoption have receded. Cost declines in battery manufacturing, growth in high-power DC charging stations, and policy-driven incentives have built confidence in electric mobility. Regular

collaboration among automakers, utilities, policymakers, local communities, and technology innovators is quickly reshaping a transportation landscape once dominated by internal combustion engines. Looking ahead, EVs are on track to become an integral part of global strategies for sustainable development, generating significant economic opportunities and ecological benefits in the process.

# 12 References

## 12.1 Introduction to Electric Vehicles (EVs) and the Conditions Shaping Modern Transportation

## 12.2 Foundations of EV Technology

51

- News Channel Nebraska: "Automotive Emission Control System Market: 2025."

- SAE International: "Advances in Emission Regulations and Control Technologies."

- Sparkion (2024): "Top EV Technology Trends Shaping the Future in 2025."

- Greenlancer (2024): "EV Market Trends."

- WiTricity: "EV vs ICE: Surprising Differences in Efficiency, Cost, and Impact."

- International Energy Agency (IEA): "Electric Vehicles" Overview.

- IEA: "Global EV Outlook 2023."

- U.S. Environmental Protection Agency (EPA): "Highlights of the Automotive Trends Report."

- Grand View Research: "Internal Combustion Engine Market Size, Share Report, 2030."

- American Automobile Association (AAA) Data on Vehicle Operating Costs.

- Electric Power Research Institute (EPRI) Data on EV Maintenance and Repair.

## 12.3 Overview of Electric Vehicle Adoption

- International Energy Agency (IEA) – Global EV Outlook 2024: Descriptive Link

- EV Volumes – The Electric Vehicle World Sales Database: Descriptive Link

- MIT Technology Review on China's EV Dominance: Descriptive Link

- U.S. Department of Energy – AFDC: Descriptive Link

- European Environment Agency – Electric Vehicles Analysis: Descriptive Link

- ICCT (International Council on Clean Transportation) EV Policy Developments: Descriptive Link

- AAA – Annual Electric Vehicle Sentiment Survey: Descriptive Link

- Pew Research Center – Americans' Views of Electric Vehicles: Descriptive Link

## 12.4 Policy and Economic Drivers of EV Adoption

## 12.5 Addressing the Infrastructure Requirements and Progress for EVs

## 12.6 Explore the Challenges Facing EV Infrastructure

## 12.7 Projecting the Future of EV Infrastructure

- Driivz (2025). "Top EV Charging Trends for 2025." Retrieved from: https://driivz.com/blog/top-ev-charging-trends-2025-predictions/

- ScienceDirect (2025). "Renewable Energy Integration with Electric Vehicle Technology." Available at: Descriptive Link

- Business Council for Sustainable Energy & BloombergNEF (2025). "2025 Sustainable Energy in America Factbook." Retrieved from: https://bcse.org/wp-content/uploads/2025/02/2025-Sustainable-Energy-in-America-Factbook.pdf

- Database of State Incentives for Renewables & Efficiency (DSIRE). Accessed at: https://www.dsireusa.org

- U.S. Department of Energy. "EV Charging Incentives" Database. Accessed at: https://www.energy.gov/charging-incentives

- Tech Xplore. (2025). Next-generation lithium–sulfur batteries: Scientists develop large-area, high-capacity prototypes. Retrieved from https://techxplore.com/news/2025-01-generation-lithiumsulfur-batteries-scientists-large.html

- Electrek. (2025). BYD's new 1000 kW EV fast chargers are ready, could hit China as early as next week. Retrieved from https://electrek.co/2025/03/26/byds-new-1000-kw-ev-fast-chargers-are-ready-and-could-hit-china-as-early-as-next-week/

- ScienceDirect. (2025). "What's next for batteries? A radical rethink of battery technology development for transportation and grid applications." Retrieved from Descriptive Link

- GlobeNewswire. (2025). Electric Vehicle Market Size, Trends, Top Companies & Regional Insights by 2033. Retrieved from https://www.globenewswire.com/news-release/2025/03/06/3038303/0/en/Electric-Vehicle-Market-Size-Trends-Top-Companies-Regional-Insights-by-2033-Straits-Research.html

- A comprehensive review of vehicle-to-grid integration in electric vehicles: Powering the future. ScienceDirect. Available at: Descriptive Link

- Role of smart charging of electric vehicles and vehicle-to-grid in integrated renewables-based energy systems on country level. ScienceDirect.

- Vehicle-to-grid [V2G] Market Size, Share & Forecast, 2032. Fortune Business Insights. Descriptive Link

- Smart Ev Charger Market Size, Growth, Trends, Report 2034. Market Research Future. Descriptive Link

- WO2023250213A1 - Electric vehicle smart charging algorithms and hardware. Google Patents. Descriptive Link

- U.S. Department of Energy, National Renewable Energy Laboratory (NREL) – EV Charging Infrastructure Trends Report (2023)

- EVgo Corporate Announcements, 2024

- ScienceDirect – "Renewable Energy Integration with Electric Vehicle Technology," 2025

## 12.8  Examine Regional Variations in EV Adoption and Infrastructure

1. IRS Credits for New Clean Vehicles Purchased in 2023 or After: https://www.irs.gov/credits-deductions/credits-for-new-clean-vehicles-purchased-in-2023-or-after

2. DieselNet – China NEV Policy: https://dieselnet.com/standards/cn/nev.php

3. European Environment Agency – Electric Vehicles: https://www.eea.europa.eu/en/topics/in-depth/electric-vehicles

4. Inflation Reduction Act, U.S. Congress: Descriptive Link

5. International Energy Agency (2024). Global EV Outlook 2024 – Trends in Electric Vehicle Charging: Descriptive Link

6. Web Search Summary (2025). Information on large-scale EV rollout progress in Korea, Norway, UK, US, Canada, India, and Africa (Mar 27, 2025).

7. U Power & SUSCO Joint Venture (EV Magazine): Descriptive Link

8. UAE to Expand EV Infrastructure (Travel and Tour World): Descriptive Link

9. Saudi Arabia's Automotive Ambitions (The Business Year): Descriptive Link

10. Supplemental findings on EV adoption strategies in South America, Southeast Asia, and the Middle East (updated Mar 2025).

11. ICCT Major Markets EV Briefing (2024).

## 12.9  Forward-Looking Approaches to EV Adoption

- Princeton University's Anode-Free Solid-State Battery Research

- ENC and InductEV High-Power Wireless Charging Collaboration

- Universal EV Chargers – Top 5 Trends Shaping EV Charging in 2025

- [EverCharge – 2025 Industry Trends in EV Charging](#)
- [J.D. Power – GlobalData Forecast (March 2025)](#)
- [SkillFusion EV Market Insights (March 2025)](#)
- [CleanTechnica: NIO & CATL Battery Swapping (March 2025)](#)
- [Electrek EV Financing Article (March 2025)](#)
- [Pilot, GM, and EVgo Collaboration](#)
- [NaaS Technology and Xiaomi Auto Partnership](#)
- [U.S. Department of Energy VGI Strategy (2025)](#)

## 12.10  Opportunities for Stakeholders in the EV Market

- Electric Vehicle Charging & Infrastructure (2025). Private sector initiatives & innovations (including PPPs). Retrieved from: [Descriptive Link](#)
- Pilot, GM, and EVgo Collaboration Press Release (2024). Multiple fast-charging locations across US states.
- U Power and SUSCO Joint Venture (2025). Joint statement on battery swapping and Thai EV strategy.
- Dutch "Grid-Aware" Charging Pilot Report (2025). Government-sponsored pilot reducing grid congestion.
- North American Charging Network (2025). Seven automakers unite to challenge Tesla's charging dominance.
- The Atmospheric Fund (TAF) – [Descriptive Link](#)
- U.S. Bureau of Labor Statistics – [Descriptive Link](#)
- International Energy Agency (IEA), Global EV Outlook – [Descriptive Link](#)
- U.S. Department of Energy's Alternative Fuels Data Center – [Descriptive Link](#)
- U.S. Environmental Protection Agency (EPA), Electric Vehicle Myths – [Descriptive Link](#)
- National Institutes of Health (NIH), Public Health & Vehicle Electrification – [Descriptive Link](#)
- Infrastructure Investment and Jobs Act (US): [Descriptive Link](#)

- Inflation Reduction Act (US): Descriptive Link
- IEA Global EV Outlook 2021: Descriptive Link
- ScienceDirect Study on EV Adoption Policies: Descriptive Link
- Climate Adaptation Platform Policy Insights: Descriptive Link
- International Energy Agency (IEA), "Global EV Outlook 2023"
- U.S. Federal Highway Administration – Various P3 and EV charging infrastructure resources (June 2024 report)
- U.S. Department of Energy's Federal Energy Management Program (FEMP) resources on fleet electrification

## 12.11 Conclusion



\end{document}